\documentclass[preprint,3p,times,twocolumn,authoryear]{elsarticle}

\usepackage{amssymb}
\usepackage{amsmath}
\usepackage{amssymb}
\usepackage{mathtools}
\usepackage{amsthm}
\usepackage{bbm}
\usepackage{bm}
\usepackage{graphicx}
\usepackage{tikz} 
\usepackage{subfig}

\newtheorem{theorem}{Theorem}

\setcitestyle{square}
\usetikzlibrary{arrows, arrows.meta}
\tikzset{unit_1/.style={circle,minimum size=3mm,fill=red,draw,line width=0.3mm}}
\tikzset{unit_2/.style={circle,minimum size=3mm,fill=white,draw,line width=0.3mm}}
\tikzset{arr/.style={line width=0.15mm}}
\usetikzlibrary{positioning}
\usetikzlibrary{calc}

\usepackage[capitalize,noabbrev]{cleveref}

\journal{Neural Networks}

\begin{document}

\begin{frontmatter}



\title{Learning Sequence Attractors in Recurrent Networks  with Hidden Neurons}

\author[a1]{Yao Lu}
\author[a1]{Si Wu}

\affiliation[a1]{organization={School of Psychological and Cognitive Sciences,  IDG/McGovern Institute for Brain Research, Bejing Key Laboratory of Behavior and Mental Health, Peking-Tsinghua Center for Life Sciences, Center of Quantitative Biology, Academy for Advanced Interdisciplinary Studies, \\ Peking University},country={China}}

\begin{abstract}
The brain is targeted for processing temporal sequence information. It remains largely unclear how the brain learns to store and retrieve sequence memories. Here, we study how recurrent networks of binary neurons learn sequence attractors to store predefined pattern sequences and retrieve them robustly. We show that to store arbitrary pattern sequences, it is necessary for the network to include hidden neurons even though their role in displaying sequence memories is indirect. We develop a local learning algorithm to learn sequence attractors in the networks with hidden neurons. The algorithm is proven to converge and lead to sequence attractors. We demonstrate that the network model can store and retrieve sequences robustly on synthetic and real-world datasets.  We hope that this study provides new insights in understanding sequence memory and temporal information processing in the brain.
\end{abstract}



\begin{keyword}
Recurrent networks \sep attractor networks \sep sequence memory \sep sequence attractors
\end{keyword}

\end{frontmatter}


\section{Introduction}
The brain is targeted for processing temporal sequence information. Taking visual recognition for example, the conventional setting of static image processing never happens in the brain. Starting from retina, visual inputs of an image arrive in the form of optical flow, which are transformed into spike trains of retinal ganglia cells, and then transmitted through LGN, V1, V2, V4 and higher cortical regions until the image is recognized. Along the whole pathway, the computations performed by the brain are in the form of temporal sequential processing, rather than being static. For another example, when we recall an episodic memory, a sequence of events represented by neuronal responses flows into our mind, and these events do not come in disorder or isolation, but are unfolded in time, as we experience ``mental time travel'' \citep{tulving2002episodic}.
The hippocampus has been revealed to be essential for sequence memories by physiological and behavioral studies. In animals, sequences of neural activity patterns are observed in the hippocampus for memory replay and memory related tasks \citep{nadasdy1999replay,lee2002memory,foster2006reverse,pastalkova2008internally,davidson2009hippocampal,pfeiffer2013hippocampal,pfeiffer2015autoassociative}. The discovery of time cells in hippocampus shows that the brain has specialized neurons encoding the temporal structure of events \citep{eichenbaum2014time}. 
Overall, the processing of temporal sequences is critical to the brain, but 
computational modeling study on this issue lags far behind that on static information processing.

Attractor neural networks are promising computational models for elucidating the mechanisms of the brain representing, memorizing and processing information~\citep{amari1972learning,hopfield1982neural,amit1989modeling}. An attractor network is a type of recurrent networks, in which information is stored as stable states (attractors) of the network. Once stored, a memory can be retrieved robustly under the evolving of the network dynamics given noisy or incomplete cues. The experimental evidences have indicated that the brain employs attractor networks for memory related tasks \citep{khona2022attractor}. By considering simplified neuron model and threshold dynamics, the classical Hopfield networks have successfully elucidated how recurrent networks learn to store static memory patterns \citep{hopfield1982neural}. Recurrent networks of binary neurons can also generate pattern sequences by employing asymmetric weight connections \citep{amari1972learning,hopfield1982neural,kleinfeld1986sequential,sompolinsky1986temporal,bressloff1992perceptron}, to explain the sequential neural activities widely observed in the brain (e.g., in memory retrieval in the hippocampus). In this paper, we follow and extend the standard form of the recurrent networks of binary neurons and threshold dynamics, as this enables us to pursue theoretical analysis, and we investigate how the networks learn to store sequence attractors. By sequence attractor, it means the state of a recurrent network evolves in the order of the stored pattern sequence and being robust to noise.

We first show that, to store arbitrary pattern sequences, the recurrent networks which contains only visible binary neurons \citep{amari1972learning,hopfield1982neural,kleinfeld1986sequential,sompolinsky1986temporal,bressloff1992perceptron} is inadequate in Section 3. Then we argue that it is necessary for the networks to include hidden neurons in Section 4. These neurons are not directly involved in expressing pattern sequences, but they are indispensable for the networks to store and retrieve arbitrary pattern sequences.
We further develop a local learning algorithm to learn sequence attractors in the networks with hidden neurons in Section 5. The algorithm is proven to converge and lead to sequence attractors. We demonstrate that our network model can learn to store and retrieve pattern sequences robustly on synthetic and real-world datasets in Section 6. 

\section{Related Work and Our Contributions}
Learning temporal sequences in recurrent networks has been studied previously in the field of computational neuroscience. These works employ different forms of recurrent networks and have different focuses of investigation. Specifically, \citep{amari1972learning,hopfield1982neural,kleinfeld1986sequential,sompolinsky1986temporal,bressloff1992perceptron,fiete2010spike} investigated recurrent networks of binary neurons and simple threshold dynamics. This approach takes advantages of simplified models that capture the essential features of neural dynamics and allows us to pursue theoretical analysis. \citep{brea2013matching,tully2016spike} investigated recurrent networks of spiking neurons which are more biologically realistic but hard to analyze theoretically. \citep{laje2013robust,rajan2016recurrent,gillett2020characteristics,rajakumar2021stimulus} investigated recurrent networks of firing-rate neurons (e.g., sigmoid and linear-threshold), whose complexity is a trade-off between binary neurons and spiking neurons. More recently, \citep{karuvally2023general,chaudhry2023long} employ modern Hopfield networks \citep{krotov2016dense} and \citep{tang2023sequential} employ predictive coding networks for modeling sequence memory. 


In this paper, we study and extend the classical recurrent network model \citep{amari1972learning,hopfield1982neural,kleinfeld1986sequential,sompolinsky1986temporal}, with the focus of theoretical analysis. Below summarizes the main contributions of our work in comparison to related work.  

\begin{itemize}
    \item We highlight the importance of hidden neurons in the recurrent networks of binary neurons for learning arbitrary pattern sequences. \citep{amari1972learning,hopfield1982neural,sompolinsky1986temporal,bressloff1992perceptron,fiete2010spike} considered only visible neurons and hence the pattern sequences they can generate are limited. Although \citep{laje2013robust,brea2013matching,rajakumar2021stimulus,chaudhry2023long,tang2023sequential} also employed hidden neurons in sequence learning, they are based on different network architectures or neuron models.
    
    \item We have clear theoretical characterization of sequences that can be generated by our networks (Theorem 1), a result which is lacking in all other related work. Despite that this conclusion comes from the analysis of the simple model we use, it lays foundation for future work to test it in biologically more realistic networks. Although \citep{chaudhry2023long} also provided theoretical characterization of the network capacity, it is based on Rademacher (random and uniformly distributed) sequence patterns  and several approximations.

    \item Our learning algorithm is proven to converge and lead to sequence attractors, while most previous work only provided  empirical evidences on the effectiveness of their learning algorithms \citep{laje2013robust,brea2013matching,rajan2016recurrent,rajakumar2021stimulus}. Although \citep{amari1972learning,bressloff1992perceptron} gave provable results on sequence attractors, they did not include hidden neurons and hence the results only hold for a restricted class of sequences.

    \item  Our learning algorithm only requires local information between neurons, which is believed to be biologically plausible. \citep{rajakumar2021stimulus} used backpropagation which is often criticized for its biologically implausibility as it requires gradient computation and has the weight transport problem \citep{lillicrap2020backpropagation}. 
\end{itemize}

\section{Limitation of Networks without Hidden Neurons}

We first consider recurrent networks of $N$ visible binary neurons \citep{amari1972learning,hopfield1982neural,sompolinsky1986temporal,bressloff1992perceptron}. All the neurons are bidirectionally connected and their weight matrix is $\mathbf{W}$ of which $W_{ij}$ denotes the synaptic weight from the $j$-th neuron to the $i$-th neuron. Let $\bm{\xi}(t) = (\xi_1(t),...,\xi_N(t))^\top \in \{-1,1\}^N$ be the states of the neurons at time $t$. These states are synchronously updated according to the threshold dynamics, for $i=1,...,N$,
\begin{align}
    \xi_i(t+1) = \text{sign}\Big(\sum_{j=1}^N W_{ij} \xi_j(t) \Big),
\end{align}
where $\text{sign}(x)=1$ if $x\geq 0$ and $\text{sign}(x)=-1$ otherwise. The bias parameters are omitted here as they can be absorbed into the equation. Given a pair of successive network states $\bm{\xi}(t)$ and $\bm{\xi}(t+1)$, the dynamics of the network can be unfolded in time and viewed as a feedforward network, in which each output neuron is a perceptron of the inputs.

Given a sequence in the form of $\mathbf{x}(1),...,\mathbf{x}(T)\in\{-1,1\}^N$, one can use a learning algorithm to adjust $\mathbf{W}$ such that the evolution of the network state matches the pattern sequence.
Although the networks can generate some sequences of maximal length $2^N$ \citep{muscinelli2017exponentially}, they are fundamentally limited in the class of sequences that can be generated.  Since each neuron can be regarded as a perceptron, the condition that sequence $\mathbf{x}(1),...,\mathbf{x}(T)$ can be generated by the network is, for each $i$, the dataset $\{(\mathbf{x}(t),x_i(t+1))\}_{t=1}^{T-1}$ is linearly separable \citep{le1986learning,bressloff1992perceptron,brea2013matching,muscinelli2017exponentially}.

For a simple example of sequences which cannot be generated by the networks, consider the sequence 
\begin{align*}
    \begin{pmatrix}
        1 \\
        1
    \end{pmatrix}, 
    \begin{pmatrix}
        \ \ \ 1 \\
        -1
    \end{pmatrix},    
    \begin{pmatrix}
        -1 \\
        \ \ \ 1
    \end{pmatrix},
    \begin{pmatrix}
        -1 \\
        -1
    \end{pmatrix},
    \begin{pmatrix}
        1 \\
        1
    \end{pmatrix}
\end{align*}
with $N=2$ and $T=5$. To generate this sequence, the first neuron of the network needs to map $(1,1)^\top$ to $1$, $(1,-1)^\top$ to $-1$, $(-1,1)^\top$ to $-1$ and $(-1,-1)^\top$ to $1$. This mapping is essentially the XOR operation which cannot be performed by a perceptron \citep{marvin1969perceptrons}.

\begin{figure}[h!]
    \centering
    \includegraphics[width=0.4\textwidth]{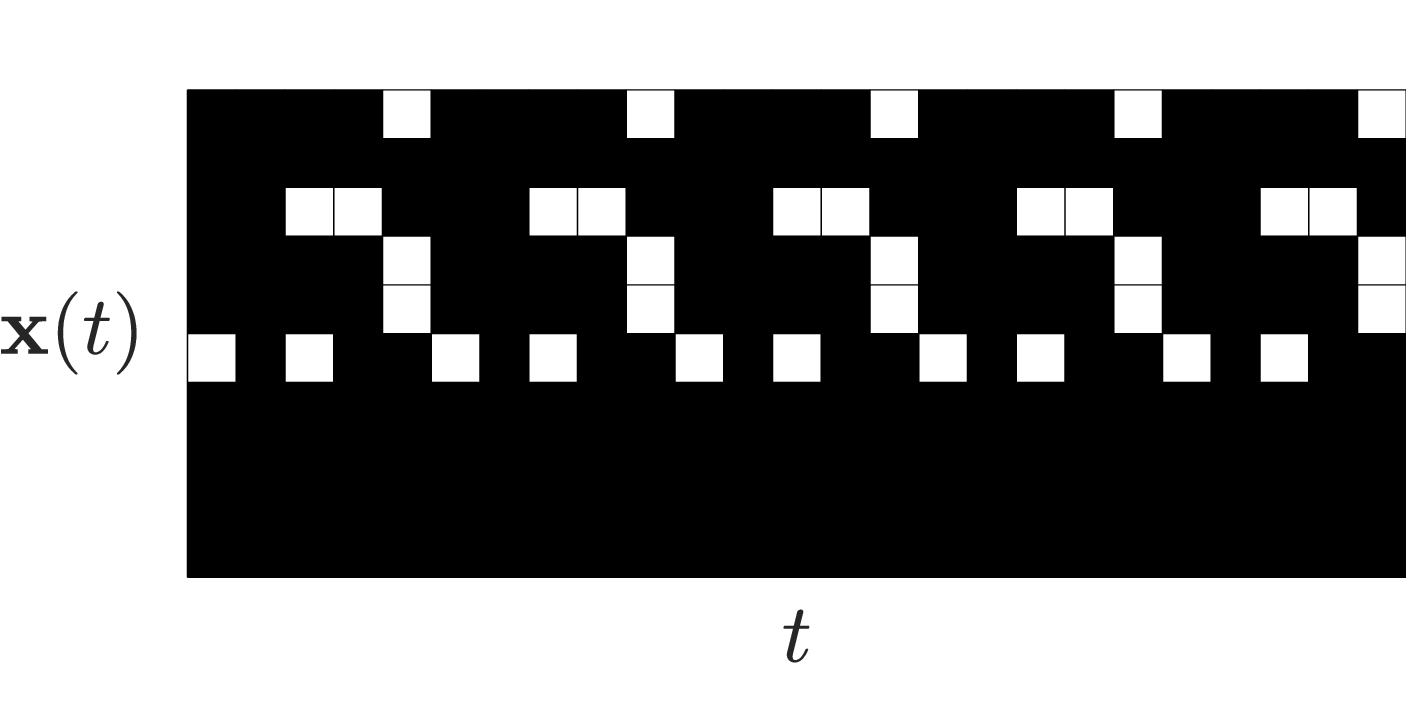} 
    \hspace{0.5cm}
    \includegraphics[width=0.4\textwidth]{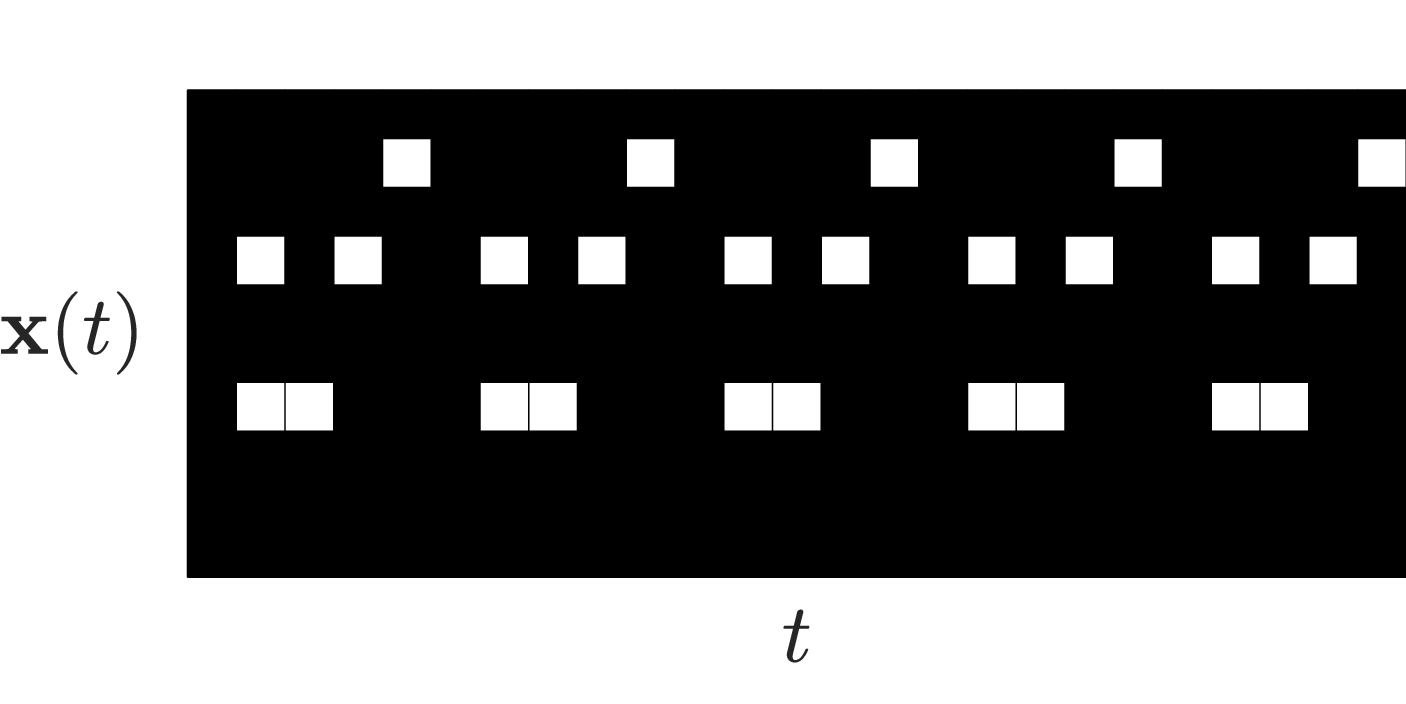} 
    \caption{Two example sequences which cannot be generated by networks without hidden neurons. White squares denote positive ones and black squares denote negative ones.}
    \label{fig:toy_one_layer}
\end{figure}

In Figure \ref{fig:toy_one_layer}, we show additional examples of sequences that cannot be generated by the network. The sequences are synthetically constructed. We then test if the perceptron learning algorithm can learn the sequences. Since the algorithm converges if the linear separability condition is met \citep{marvin1969perceptrons}, the divergence of the algorithm implies that the sequences cannot be generated by the networks.

\section{Networks with Hidden Neurons}
\begin{figure*}[h!]
\centering
\begin{tikzpicture}
\node[unit_1] (1) at (0,-.55) {};
\node[unit_1] (2) at (0.6,-.55) {};
\node[unit_1] (3) at (1.2,-.55) {};
\node[unit_1] (4) at (-0.6,-.55) {};
\node[unit_1] (5) at (-1.2,-.55) {};
\node[unit_2] (6) at (-0.75,.55) {};
\node[unit_2] (7) at (-0.25,.55) {};
\node[unit_2] (8) at (0.25,0.55) {};
\node[unit_2] (9) at (0.75,0.55) {};
\path[>={Latex[length=1mm,width=1mm]},<->]
(1) edge[arr] node {} (6)
(1) edge[arr] node {} (7)
(1) edge[arr] node {} (8)
(1) edge[arr] node {} (9)
(2) edge[arr] node {} (6)
(2) edge[arr] node {} (7)
(2) edge[arr] node {} (8)
(2) edge[arr] node {} (9)
(3) edge[arr] node {} (6)
(3) edge[arr] node {} (7)
(3) edge[arr] node {} (8)
(3) edge[arr] node {} (9)
(4) edge[arr] node {} (6)
(4) edge[arr] node {} (7)
(4) edge[arr] node {} (8)
(4) edge[arr] node {} (9)
(5) edge[arr] node {} (6)
(5) edge[arr] node {} (7)
(5) edge[arr] node {} (8)
(5) edge[arr] node {} (9);
\node[] (n1) at (2,0.) {};
\node[] (n2) at (5,0.) {};
\path[>={Latex[length=2mm,width=2mm]},->]
(n1) edge[line width=0.8mm] node[above] {\small{Unfolding in time}} (n2);

\path[draw=none] (2,0) -- (-2,0);
\node[unit_1] (1) at (0+7,-1.1) {};
\node[unit_1] (2) at (-0.6+7,-1.1) {};
\node[unit_1] (3) at (0.6+7,-1.1) {};
\node[unit_1] (4) at (1.2+7,-1.1) {};
\node[unit_1] (5) at (-1.2+7,-1.1) {};
\node[unit_1] (6) at (0+7,1.1) {};
\node[unit_1] (7) at (-0.6+7,1.1) {};
\node[unit_1] (8) at (0.6+7,1.1) {};
\node[unit_1] (9) at (1.2+7,1.1) {};
\node[unit_1] (10) at (-1.2+7,1.1) {};
\node[unit_2] (11) at (-0.75+7,0) {};
\node[unit_2] (12) at (-0.25+7,0) {};
\node[unit_2] (13) at (.25+7,0) {};
\node[unit_2] (14) at (0.75+7,0) {};

\node[align=left] (15) at (2.1+7,-1.1) {$\bm{\xi}(t)$};
\node[align=left] (16) at (2.1+7,1.1) {$\bm{\xi}(t+1)$};
\node[align=left] (17) at (2.1+7,0) {$\bm{\zeta}(t)$};
\node[] (18) at (1.3+7,0.55-1.1) {$\mathbf{U}$};
\node[] (19) at (1.3+7,1.65-1.1) {$\mathbf{V}$};
\node[] (20) at (-2.9+7,1-1.1) {};
\path[>={Latex[length=1mm,width=1mm]},->]
(1) edge[arr] node {} (11)
(1) edge[arr] node {} (12)
(1) edge[arr] node {} (13)
(1) edge[arr] node {} (14)
(2) edge[arr] node {} (11)
(2) edge[arr] node {} (12)
(2) edge[arr] node {} (13)
(2) edge[arr] node {} (14)
(3) edge[arr] node {} (11)
(3) edge[arr] node {} (12)
(3) edge[arr] node {} (13)
(3) edge[arr] node {} (14)
(4) edge[arr] node {} (11)
(4) edge[arr] node {} (12)
(4) edge[arr] node {} (13)
(4) edge[arr] node {} (14)
(5) edge[arr] node {} (11)
(5) edge[arr] node {} (12)
(5) edge[arr] node {} (13)
(5) edge[arr] node {} (14)
(11) edge[arr] node {} (6)
(11) edge[arr] node {} (7)
(11) edge[arr] node {} (8)
(11) edge[arr] node {} (9)
(11) edge[arr] node {} (10)
(12) edge[arr] node {} (6)
(12) edge[arr] node {} (7)
(12) edge[arr] node {} (8)
(12) edge[arr] node {} (9)
(12) edge[arr] node {} (10)
(13) edge[arr] node {} (6)
(13) edge[arr] node {} (7)
(13) edge[arr] node {} (8)
(13) edge[arr] node {} (9)
(13) edge[arr] node {} (10)
(14) edge[arr] node {} (6)
(14) edge[arr] node {} (7)
(14) edge[arr] node {} (8)
(14) edge[arr] node {} (9)
(14) edge[arr] node {} (10);
\path[draw=none] (-2.5,0) -- (3,0);
\end{tikzpicture}
\caption{Recurrent network with hidden neurons. The red circles denote visible neurons and the white circles denote hidden neurons.}
\label{fig:model_h}
\end{figure*}
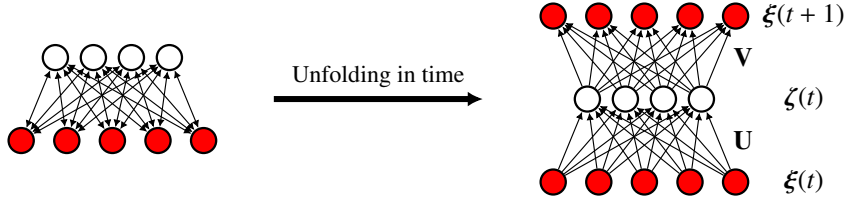

To overcome the limitation of networks with only visible neurons, we consider including a group of hidden neurons in the network.
The visible and hidden neurons are bidirectionally connected, and there is no intra-connection within visible neurons or hidden neurons. Let $\mathbf{U}$ be the weight matrix from visible neurons to hidden neurons, of which $U_{ij}$ denotes the synaptic weight from the $j$-th visible neuron to the $i$-th hidden neuron, and $\mathbf{V}$ be the weight matrix from hidden neurons to visible neurons, of which $V_{ji}$ denotes the synaptic weight from the $i$-th hidden neuron to the $j$-th visible neuron. Let $\bm{\xi}(t) = (\xi_1(t),...,\xi_N(t))^\top\in\{-1,1\}^N$ be the states of visible neurons and $\bm{\zeta}(t)=(\zeta_1(t),...,\zeta_M(t))^\top\in\{-1,1\}^M$ be the states of hidden neurons at time $t$. 
These states are synchronously updated according to, for $i=1,...,M$ and $j=1,...,N$,
\begin{align}
    \zeta_i(t) &= \text{sign}\Big(\sum_{k=1}^N U_{ik} \xi_k(t)\Big),
    \label{model:1}\\
    \xi_j(t+1) &= \text{sign}\Big(\sum_{k=1}^M V_{jk}\zeta_k(t) \Big),
    \label{model:2}
\end{align}
where we omit the bias parameters as they can be absorbed into the equations. As illustrated in Figure \ref{fig:model_h}, given a pair of successive network states $\bm{\xi}(t)$ and $\bm{\xi}(t+1)$, the dynamics of the network can be unfolded in time and viewed as a feedforward network with a hidden layer of neurons.

The networks of $M$ hidden neurons can generate arbitrary sequences with Markov property and of length at least $M$, as stated in Theorem 1. We provide a constructive proof based on an one-hot encoding by the hidden neurons in the Appendix.

\begin{theorem}
Let $\mathbf{x}(1),...,\mathbf{x}(T)\in\{-1,1\}^N$ such that $\mathbf{x}(i)\neq \mathbf{x}(j)$ for $i\neq j$ except that $\mathbf{x}(1)=\mathbf{x}(T)$. Then $\mathbf{x}(1),...,\mathbf{x}(T)$ can be generated by the network defined in (\ref{model:1})(\ref{model:2}) for $M=T-1$.
\end{theorem}

\section{Learning}

To learn the weight matrices, one can first unfold the recurrent network with hidden neurons in time such that it becomes a feedforward network with a hidden layer, and the pairs of successive patterns in the sequence constitute the training examples, as illustrated in Figure \ref{fig:model_h}. 
However, learning in the unfolded feedforward network is difficult since the backpropagation algorithm cannot be applied as the neurons are not differentiable. 

We propose a new learning algorithm to learn the weight matrices in the unfolded feedforward networks, which draws inspirations from three ideas: feedback alignment \citep{lillicrap2016random}, target propagation \citep{yann1987modeles,bengio2014auto,litwin2017optimal} and three-factor rules \citep{fremaux2016neuromodulated,kusmierz2017learning}. As in feedback alignment, it requires a random matrix $\mathbf{P}$, which is fixed during the learning process, to backpropagate signals. As in target propagation, it does not propagate errors but targets to create surrogate targets for the hidden neurons. Each weight parameter is updated by a three-factor rule, in which the presynaptic activation, the postsynaptic activation and an error term as neuromodulation are multiplied. The three-factor rule is similar to the one in \citep{bressloff1992perceptron} and known as margin perceptron in the machine learning literature \citep{collobert2004links}.

The algorithm works as follows. Given a pair of successive patterns $\mathbf{x}(t)$ and $\mathbf{x}(t+1)$, for $i=1,...,M$ and $j=1,...,N$ in parallel,

\begin{enumerate}
\item Update $\mathbf{U}$ by
\begin{align}
z_i(t+1) &= \text{sign}\Big(\sum_{k=1}^N P_{ik}x_k(t+1)\Big),
\label{alg:U1}\\
\mu_i(t) &= H\Big(\kappa - z_i(t+1)\sum_{k=1}^N U_{ik} x_k(t)\Big), \label{alg:U2}\\
U_{ij} &\leftarrow U_{ij} + \eta \mu_i(t) z_i(t+1)x_j(t).
\label{alg:U3}
\end{align}
\item Update $\mathbf{V}$ by
\begin{align}
y_i(t) &= \text{sign}\Big(\sum_{k=1}^N U_{ik} x_k(t)\Big), \label{alg:V1}\\
\nu_j(t) &= H\Big(\kappa - x_j(t+1)\sum_{k=1}^M V_{jk} y_k(t)\Big), \label{alg:V2}\\
V_{ji} &\leftarrow V_{ji} + \eta \nu_j(t) x_j(t+1)y_i(t), \label{alg:V3}
\end{align}
\end{enumerate}
where $P_{ik}$ denotes the $(i,k)$ entry of the fixed random matrix $\mathbf{P}$, $H(\cdot)$ is the Heaviside function ($H(x)=1$ if $x\geq 0$ and $H(x)=0$ otherwise), $\kappa > 0$ is the robustness hyperparameter and $\eta > 0$ is the learning rate hyperparameter. $\mu_i(t)$ and $\nu_j(t)$ can be interpreted as the error terms for the hidden and the visible neurons, respectively. $z_i(t+1)$ can be interpreted as the synaptic input from an external neuron. The above procedure is then repeated for each $t$.

\subsection{Analysis}
In this section, we provide theoretical analysis of the algorithm. The proofs are left to the Appendix.
First, we provide convergence guarantee of the algorithm.
\begin{theorem}
Given the definitions in (4)(5)(7)(8), for all $i$, $j$ and $t$, if a solution exists such that $\mu_i(t)=0$ and $\nu_j(t)=0$ , then the algorithm (4)-(9) converges in finite steps given $U_{ij}$ and $V_{ji}$ are initialized to zero.
\end{theorem}

Next, we show the algorithm can reduce error $\mu_i(t)$ for a single step of updating $\mathbf{U}$. The theorem can be trivially extended for $\nu_j(t)$ and $\mathbf{V}$ by a similar proof.
\begin{theorem}
Given the definitions in (4)(5), let
\begin{align}
U_{ik}' &= U_{ik} + \eta \mu_i(t) z_i(t+1)x_k(t), \\
\mu_i'(t) &= H\Big(\kappa - z_i(t+1)\sum_{k=1}^N U_{ik}'  x_k(t)\Big).
\end{align}
Then $\mu_i'(t) = 0$ for sufficiently large $\eta > 0$.
\end{theorem}

To understand why reducing the errors $\mu_i(t)$ and $\nu_j(t)$ leads to sequence attractors, we present the following result. 
\begin{theorem}

Given the definitions in (7)(8), let $\hat{\mathbf{y}}(t)=(\hat{y}_1(t),...,\hat{y}_M(t))^\top\in \{-1,1\}^M$ such that $\sum_k |\hat{y}_k(t)-y_k(t)| < \epsilon$. If $\nu_j(t)=0$ and 
\begin{align}
\epsilon \cdot \max_k |V_{jk}| < \kappa,
\label{eq:kappa}
\end{align}
then
\begin{align}
     x_j(t+1) = \emph{\text{sign}}\Big(\sum_{k=1}^M V_{jk} \hat{y}_k(t)\Big).
\end{align}
\end{theorem}
Theorem 4 shows that when the errors are zero, given perturbed hidden neuron states $\hat{\mathbf{y}}(t)$, we have $\mathbf{x}(t+1) = \text{sign}(\mathbf{V}\hat{\mathbf{y}}(t))$. The result can be trivially extended to show that given perturbed visible neuron states $\hat{\mathbf{x}}(t)$ we have $\mathbf{y}(t) = \text{sign}(\mathbf{U}\hat{\mathbf{x}}(t))$ by a similar proof. Therefore, the network can generate sequence $\mathbf{x}(1),...,\mathbf{x}(T)$ as an attractor. From Theorem 4, we can also see that $\kappa$ acts as the robustness hyperparameter as it controls the level of perturbation $\epsilon$ for inequality (\ref{eq:kappa}) to hold.

\begin{figure}[h!]
    \centering
    \subfloat[][Gaussian]{
    \includegraphics[width=0.35\textwidth]{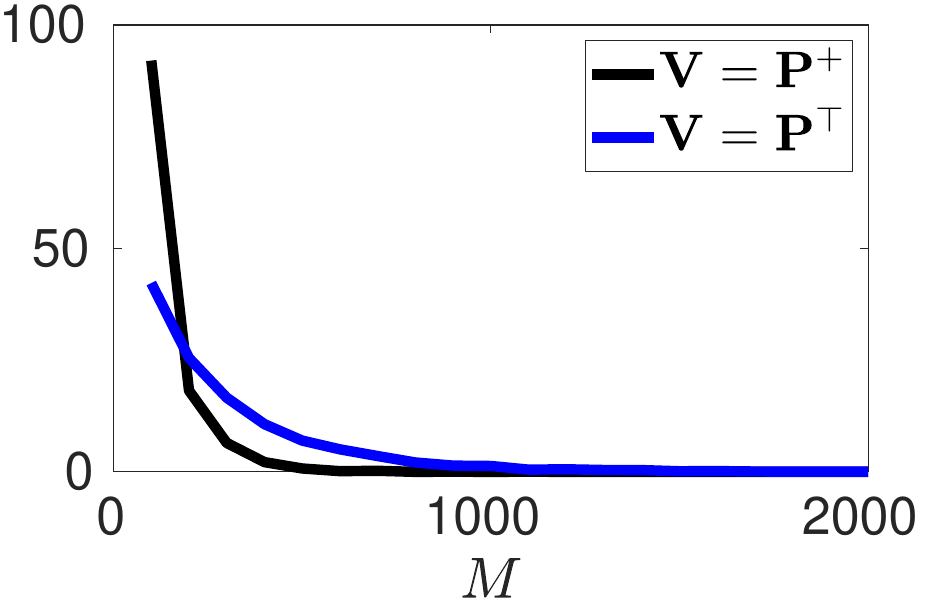}}
    \hspace{0.5cm}
    \subfloat[][Uniform]{
    \includegraphics[width=0.35\textwidth]{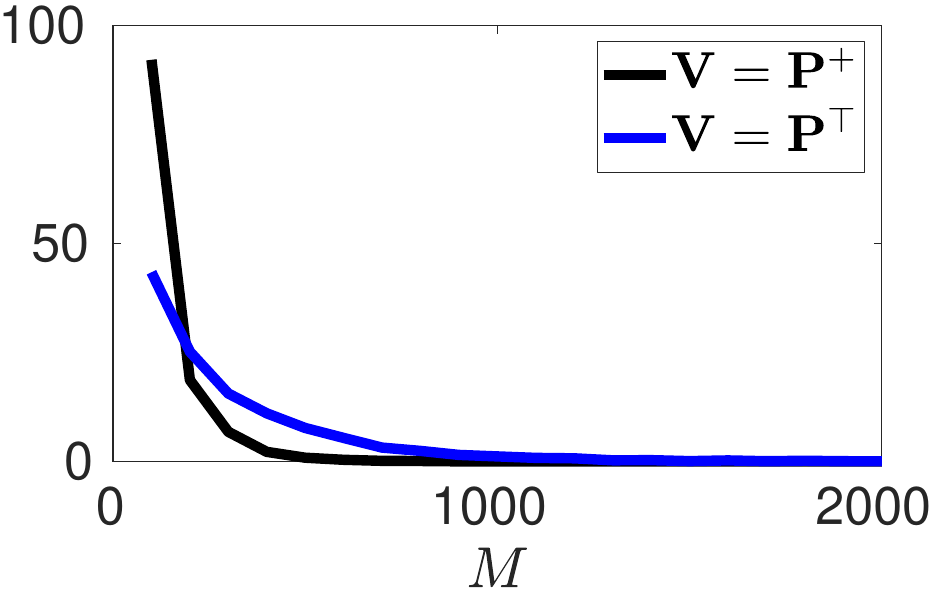}}
    \caption{Reconstruction error $\|\mathbf{x}- \text{sign}(\mathbf{V} \text{sign}(\mathbf{Px}))\|_1$ where $\mathbf{P}$ is a $M\times N$ random matrix and $\mathbf{x}$ is a random vector uniformly sampled from $\{-1,1\}^N$. $\mathbf{P}^+$ denotes the pseudo-inverse of $\mathbf{P}$. $\mathbf{P}^\top$ denotes the transpose of $\mathbf{P}$. (a) The entries of $\mathbf{P}$ are sampled i.i.d. from the standard Gaussian distribution. (b) The entries of $\mathbf{P}$ are sampled i.i.d. from the uniform distribution on ${[-1,1]}$. In (a) and (b), $N=100$ and the results are averaged over 100 trials. The results are similar in (a) and (b). The error bars are not displayed for visual clarity.}
    \label{fig:rand_perm}
\end{figure}

To understand why the algorithm works despite that $\mathbf{P}$ is a random matrix and fixed during learning, consider the following. If the update of $\mathbf{U}$ converges, then $\mu_i(t) = 0$ for all $i$. Therefore, 
\begin{align}
    y_i(t) 
    &= \text{sign}\Big(\sum_{k=1}^N U_{ik} x_k(t)\Big)  \\
    &= z_i(t+1) \\
    &= \text{sign}\Big(\sum_{k=1}^N P_{ik}x_k(t+1) \Big).
    \label{eq:y_i}
\end{align}
The update of $\mathbf{V}$ aims at making the condition 
$\text{sign}\Big(\sum_{k=1}^M V_{jk} y_k(t)\Big) = x_j(t+1)$ hold,
which is 
\begin{align}
\text{sign}(\mathbf{V} \text{sign}(\mathbf{Px}(t+1))) = \mathbf{x}(t+1),    
\end{align}
in matrix form when $y_k(t)$ is substituted by (\ref{eq:y_i}). For large $M$, a solution $\mathbf{V}$ exists, that is, the pseudo-inverse of $\mathbf{P}$ or the transpose of $\mathbf{P}$. The numerical result is shown in Figure \ref{fig:rand_perm}. The phenomenon might be explained by the high-dimensional probability theory \citep{vershynin2018high}.

\subsection{Robustness Hyperparameter}
Having a hyperparameter $\kappa$ in the algorithm is not problematic in practice. One can simply set $\kappa=1$ as we did for all the experiments in the next section and adjust the scale of initial weights and the learning rate. In margin perceptron, the margin learned is disproportional to the learning rate \citep{collobert2004links}. The margin is defined as the reciprocal of the weight magnitude, which  is related to the robustness hyperparameter, as shown in Theorem 4. Therefore, it can be interpreted that the robustness hyperparameter is automatically adjusted during learning.

\section{Experiments}\label{sec:exp}

We ran experiments on synthetic and real-world sequence datasets for the networks with hidden neurons by the algorithm proposed in the previous section to learn sequence attractors. All the experiments were carried out in MATLAB and PyTorch. In all the experiments, each weight parameter of $\mathbf{U}$, $\mathbf{V}$ and $\mathbf{P}$ was sampled i.i.d. from Gaussian distribution with mean zero and variance $1\times 10^{-6}$, learning rate $\eta = 1\times 10^{-3}$ and robustness $\kappa=1$. In each experiment, we ran the algorithm for $500$ epochs. In each epoch, the algorithm ran on $(\mathbf{x}(t),\mathbf{x}(t+1))$ from the start to the end of each sequence. No noise was added during learning. Noise was added only at retrieval. We also provide additional experiments in the Appendix.

\begin{figure}[t!]
    \centering
    \includegraphics[width=0.4\textwidth]{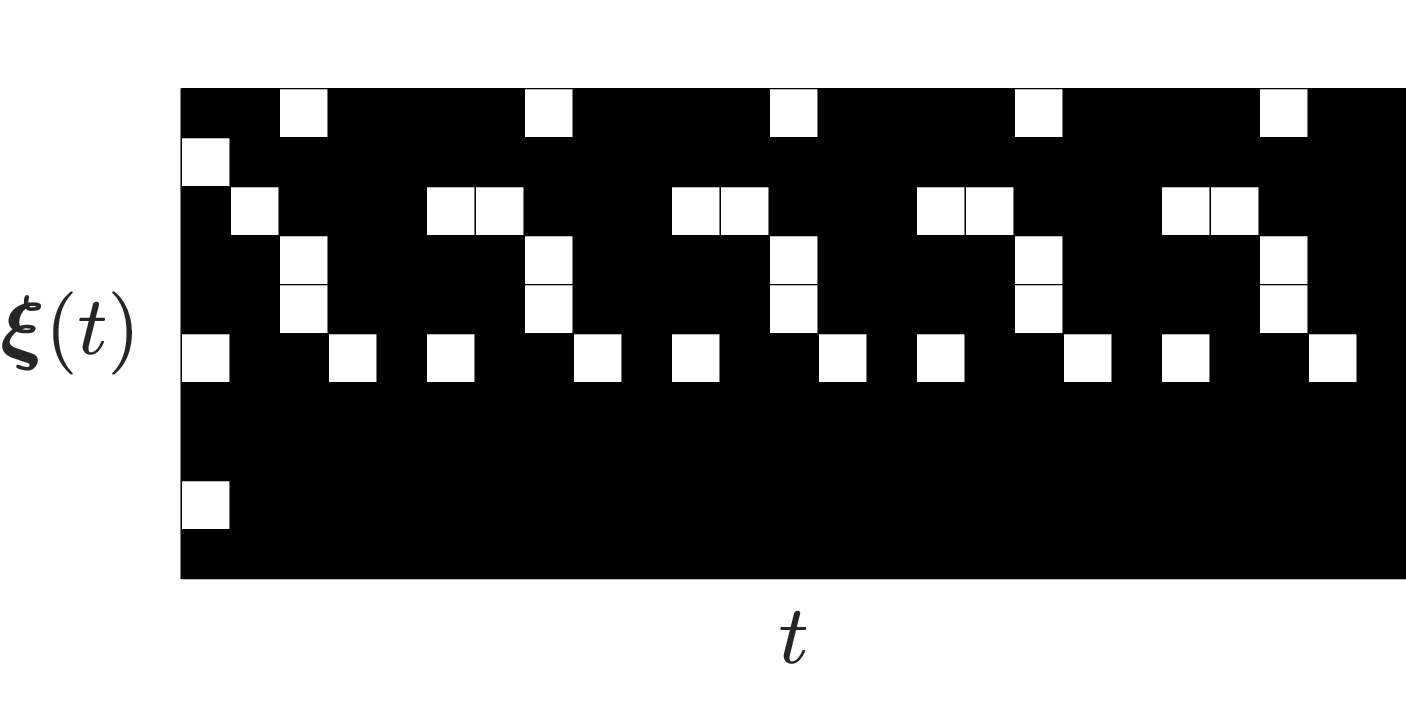}
    \hspace{0.5cm}
    \includegraphics[width=0.4\textwidth]{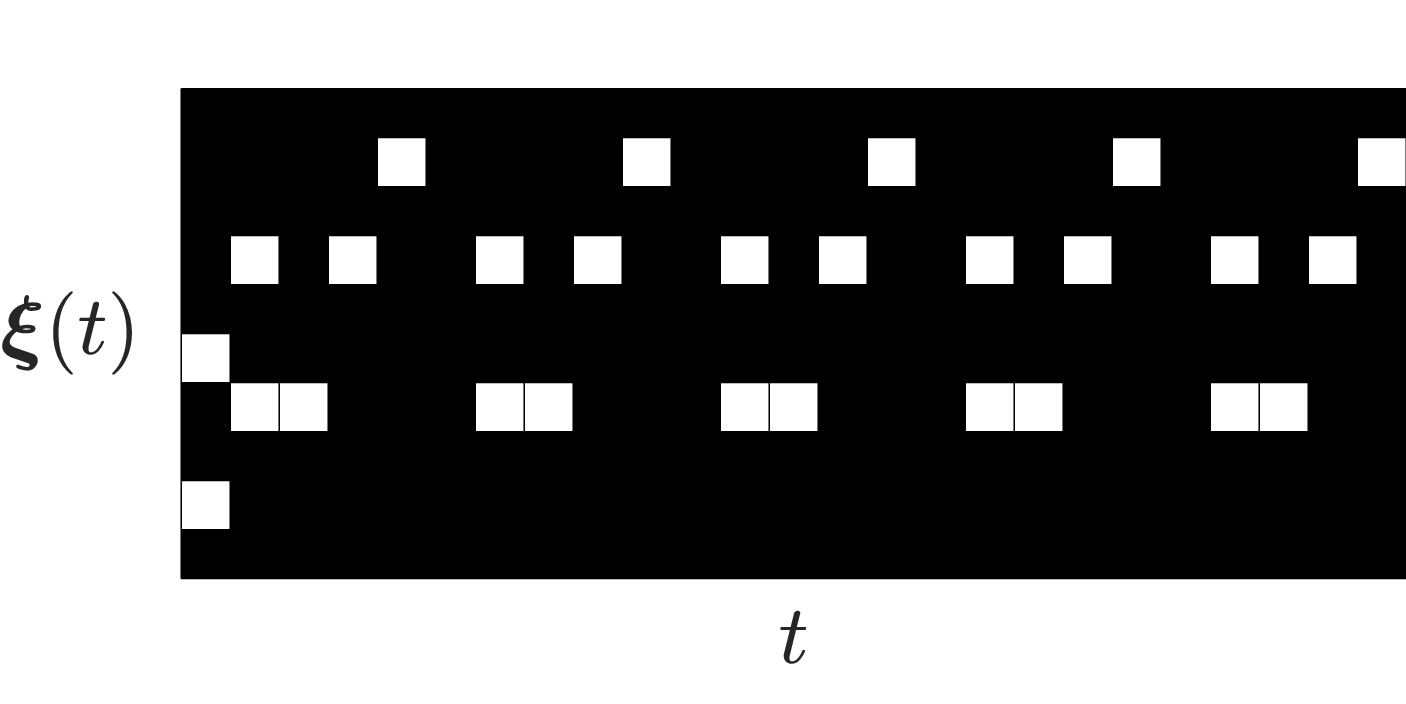}
    \caption{Recurrent networks with hidden neurons can generate the two sequences in Figure \ref{fig:toy_one_layer}, which cannot be generated without hidden neurons, despite noisy initial states. Note that in the first column of each diagram two salt-and-pepper noises are added to test the robustness of the retrieval.}
    \label{fig:toy_two_layer}
    
\end{figure}

\subsection{Toy Examples}

To show the recurrent networks with hidden neurons can overcome the limitation of the  networks without hidden neurons, we conducted experiments on the examples in Figure \ref{fig:toy_one_layer}. We constructed a network of visible neurons $N=10$ and hidden neurons $M=50$ for each example. After learning, we tested the robustness of the networks in retrieval by adding two salt-and-pepper noises (flipping the states of two out of ten neurons) to the first pattern of a sequence and set it to be the initial network state. 

The results are shown in Figure \ref{fig:toy_two_layer}, from which we can see that the networks with hidden neurons can generate sequences which cannot be generated by the networks without hidden neurons and retrieve them robustly under moderate level of noise.

\subsection{Random Sequences}

We generated periodic sequences of random patterns $\mathbf{x}(1),...,\mathbf{x}(T)\in \{-1,1\}^N$. In each sequence, $\mathbf{x}(i)\neq \mathbf{x}(j)$ for $i\neq j$ except that $\mathbf{x}(1)=\mathbf{x}(T)$ for the periodicity. We set $N=100$ and varied period length $T$. We sampled each $\mathbf{x}(t)$ independently from the uniform distribution of $\{-1,1\}^N$ for $t=1,...,T-1$ and then resampled it if it is identical to a previous pattern. Finally, we set $\mathbf{x}(T) = \mathbf{x}(1)$.

For each random sequence, we constructed a network with hidden neurons and applied the proposed  learning algorithm.  To evaluate the effectiveness of the learning algorithm, we compared learning only $\mathbf{V}$ (with $\mathbf{U}$ fixed during learning) and learning both $\mathbf{U}$ and $\mathbf{V}$.  Once the learning was done, we tested if the network can retrieve the sequence robustly given perturbed $\mathbf{x}(1)$ with $10$ salt-and-pepper noises as the initial network state $\bm{\xi}(1)$. We define that the retrieval is successful if $\bm{\xi}(\tau + t) = \mathbf{x}(t)$ for some $\tau$ and all $t=1,...,T$. We run $100$ trials for each $T$ or $M$ setting and count the successful retrievals.

\begin{figure}[h!]
    \centering
    \subfloat[Varying sequence length]{\includegraphics[width=0.45\textwidth]{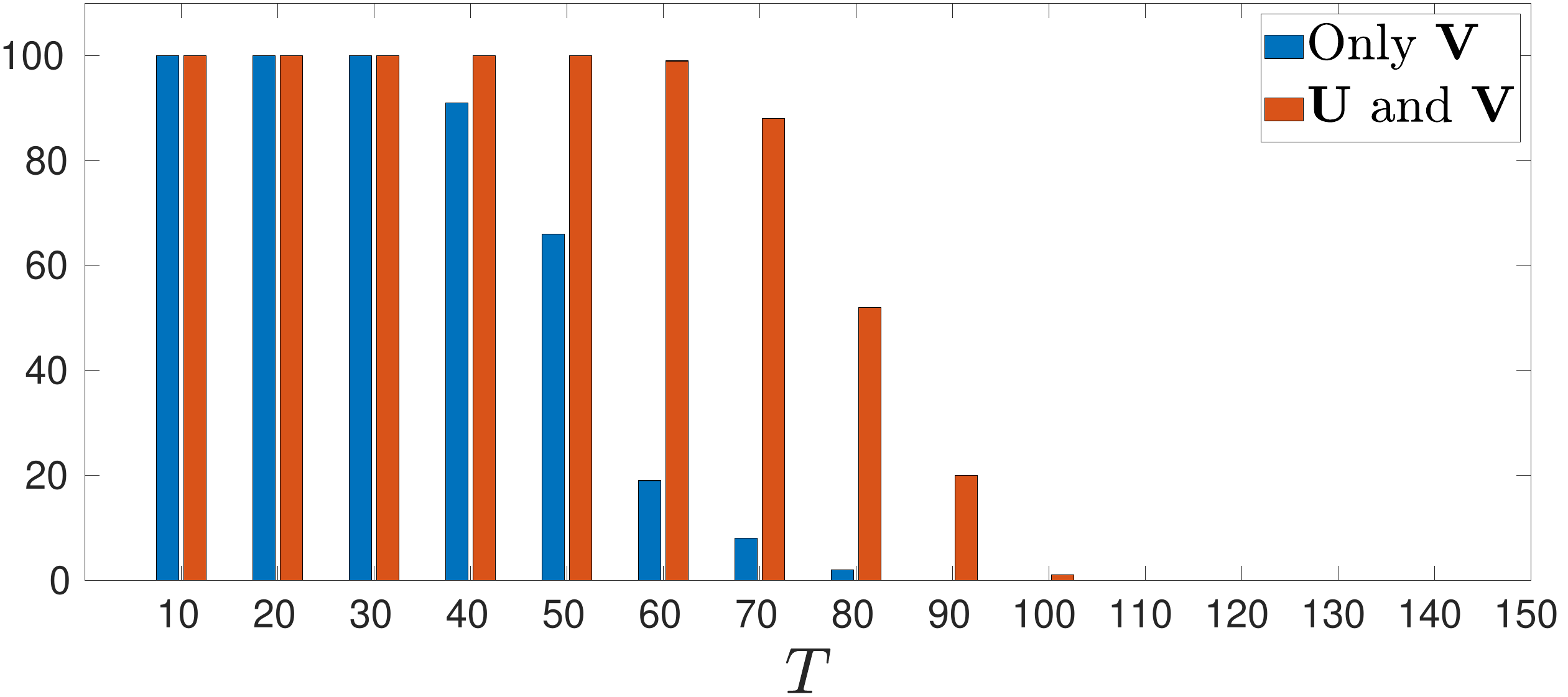}}
        
    \subfloat[Varying number of hidden neurons]{\includegraphics[width=0.45\textwidth]{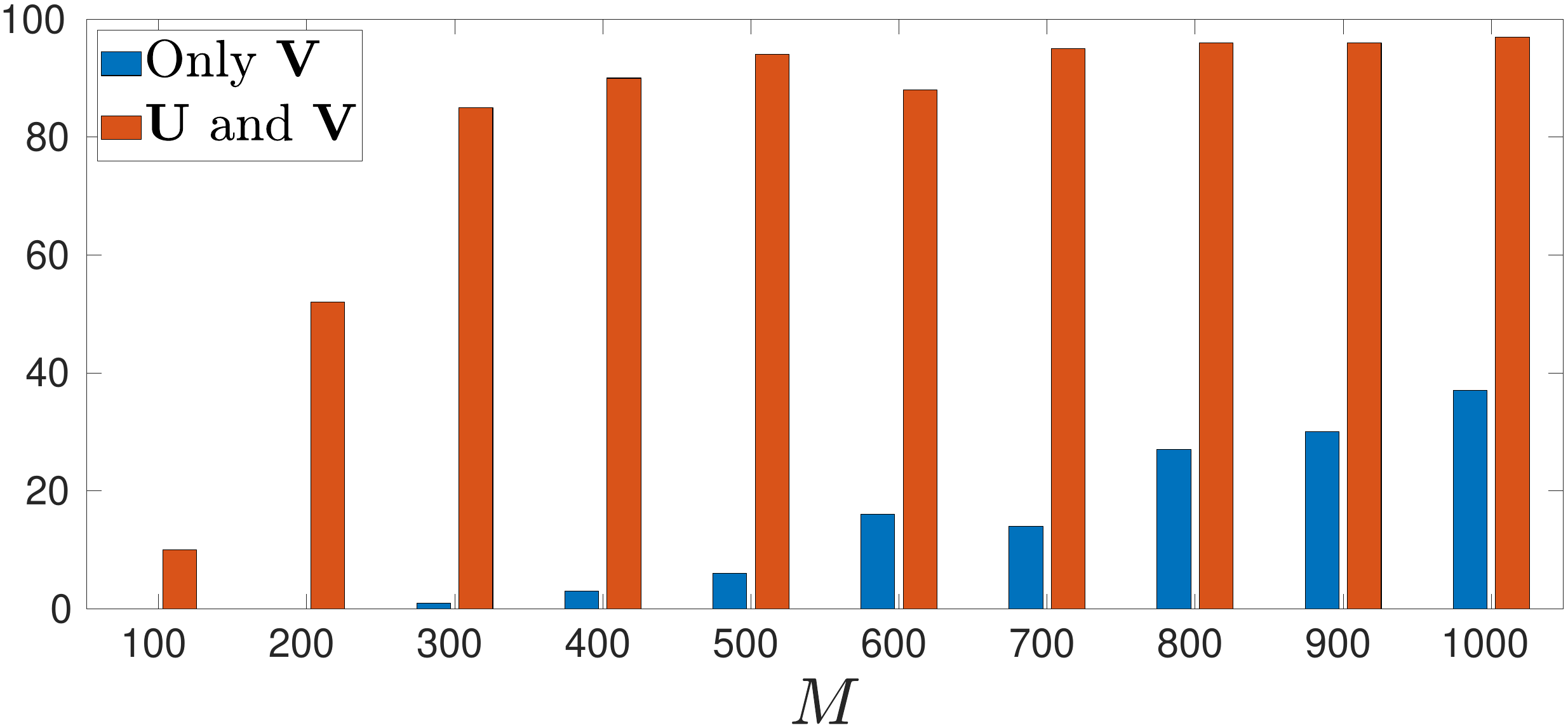}}
    \caption{Successful retrievals out of 100 trials under noise, comparing  learning only $\mathbf{V}$  and learning both $\mathbf{U}$ and $\mathbf{V}$.}
    \label{fig:vary}
\end{figure}

In Figure \ref{fig:vary} (a), we show the results with various period lengths $T$ for $M=500$. In Figure \ref{fig:vary} (b), we show the results with various numbers of hidden neurons $M$ for $T=70$. We can see learning both $\mathbf{U}$ and $\mathbf{V}$ is more effective than learning only $\mathbf{V}$. However, in both cases, the algorithm failed for large $T$, even if we increased the number of hidden neurons, which might be due to the suboptimality of the algorithm.

\subsection{Real-World Sequences}

We tested the networks with hidden neurons by our algorithm in learning real-world sequences on a silhouette sequence dataset (OU-ISIR gait database large population \citep{iwama2012isir}) and a handwriting sequence dataset (Moving MNIST \citep{srivastava2015unsupervised}). The patterns in the sequences are rather correlated since adjacent image frames are similar. To adopt the datasets for the networks to learn, we converted the image intensity values to $\pm 1$.

\begin{figure*}[h!]
    \centering
    \subfloat[][Ground truth $\mathbf{x}(t)$]{
    \includegraphics[width=0.09\textwidth]{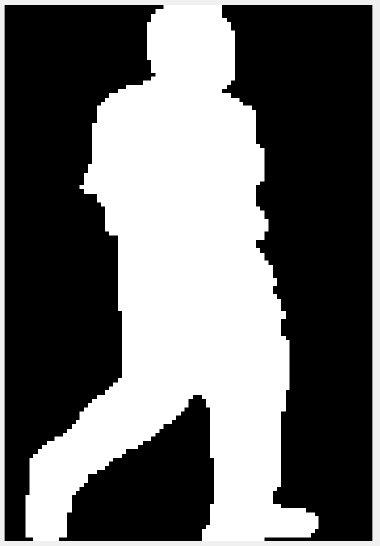}
    \includegraphics[width=0.09\textwidth]{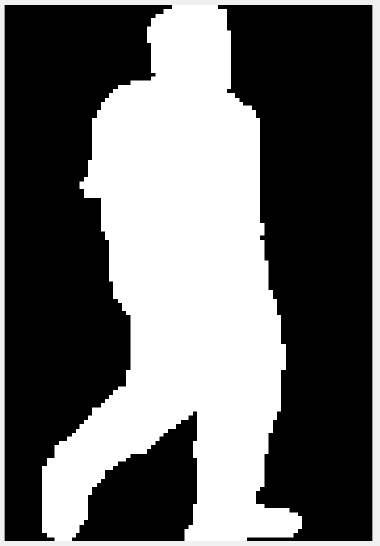}
    \includegraphics[width=0.09\textwidth]{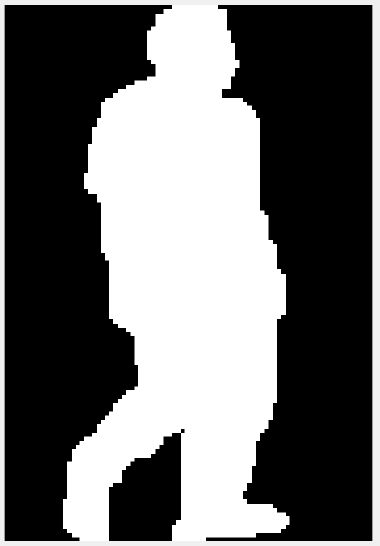}
    \includegraphics[width=0.09\textwidth]{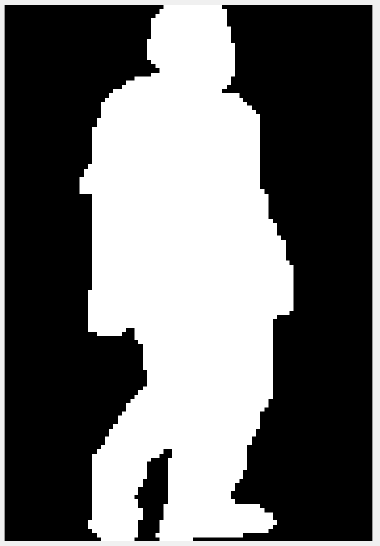}
    \includegraphics[width=0.09\textwidth]{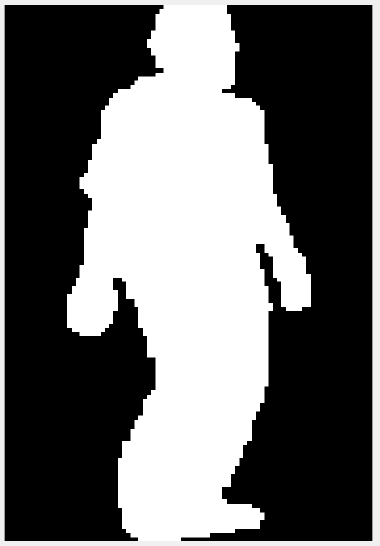}
    \includegraphics[width=0.09\textwidth]{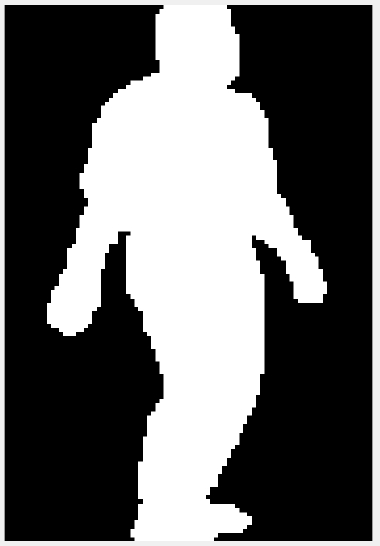}
    \includegraphics[width=0.09\textwidth]{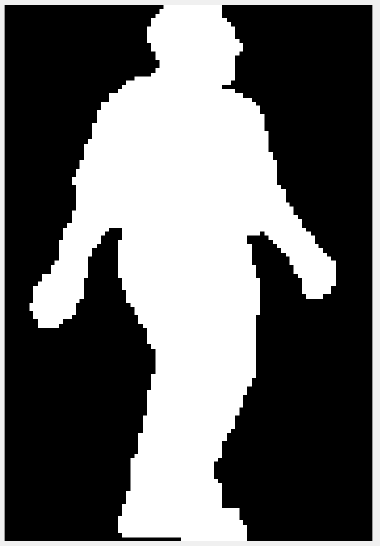}
    \includegraphics[width=0.09\textwidth]{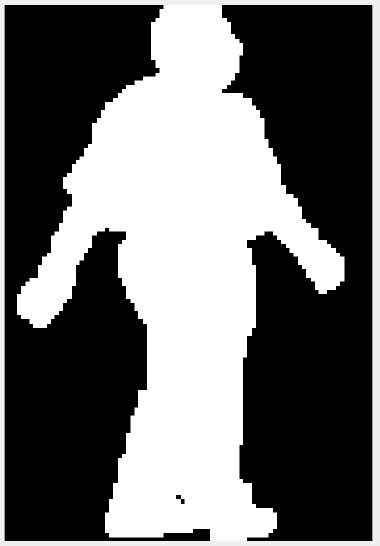}
    \includegraphics[width=0.09\textwidth]{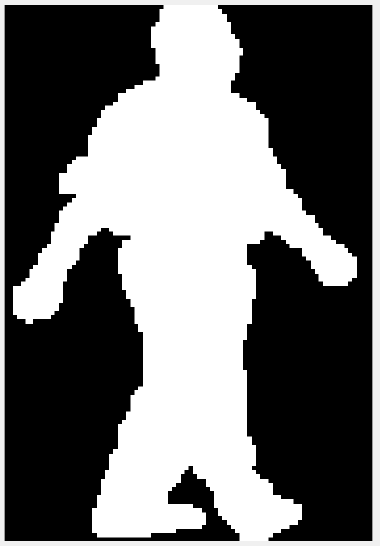}
    \includegraphics[width=0.09\textwidth]{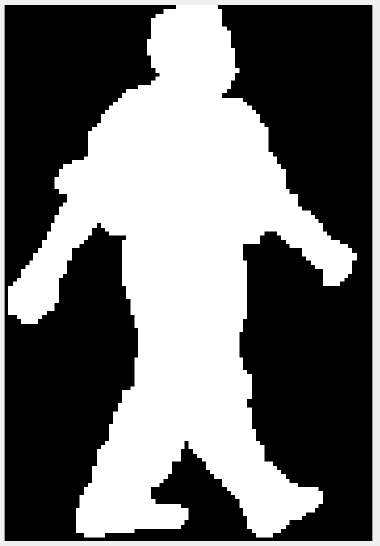}}

    \subfloat[][Network states $\bm{\xi}(t)$]{
    \includegraphics[width=0.09\textwidth]{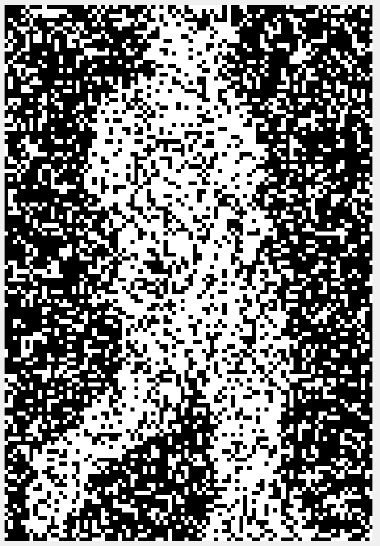}
    \includegraphics[width=0.09\textwidth]{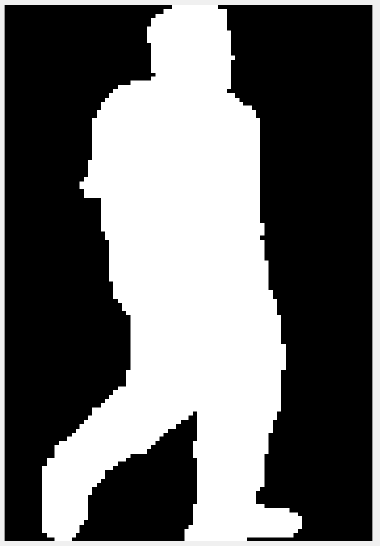}
    \includegraphics[width=0.09\textwidth]{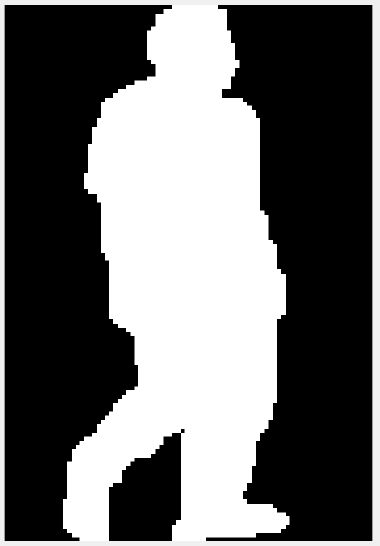}
    \includegraphics[width=0.09\textwidth]{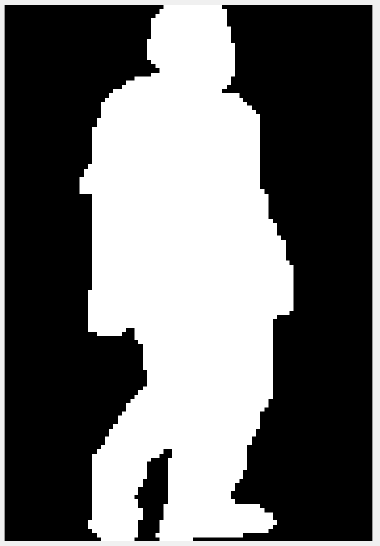}
    \includegraphics[width=0.09\textwidth]{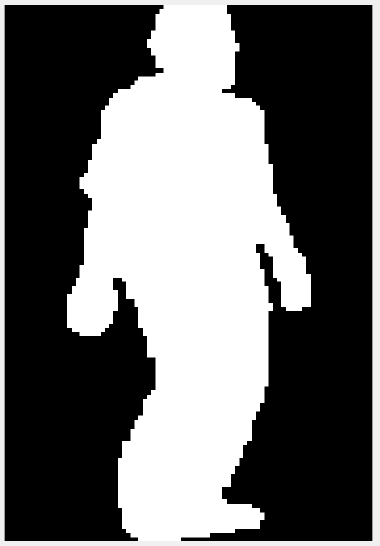}
    \includegraphics[width=0.09\textwidth]{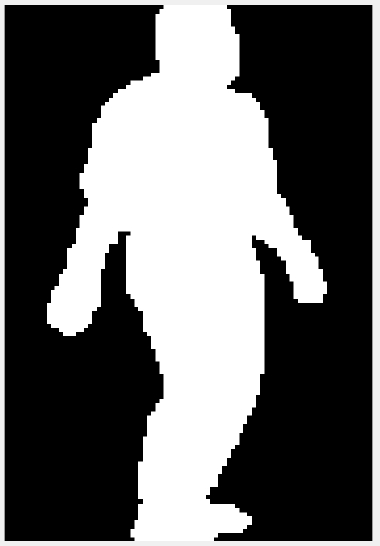}
    \includegraphics[width=0.09\textwidth]{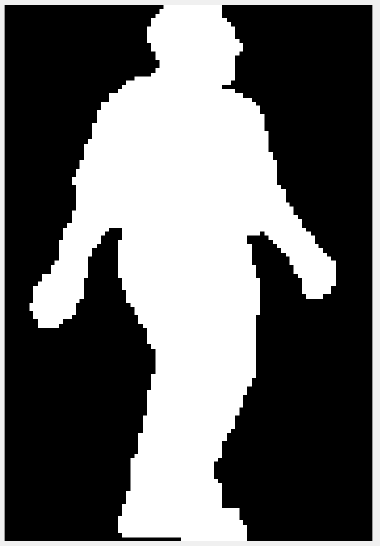}
    \includegraphics[width=0.09\textwidth]{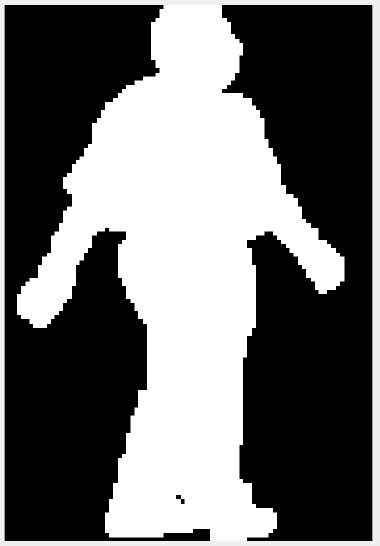}
    \includegraphics[width=0.09\textwidth]{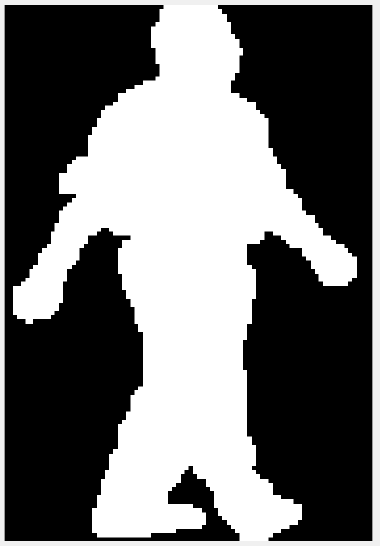}
    \includegraphics[width=0.09\textwidth]{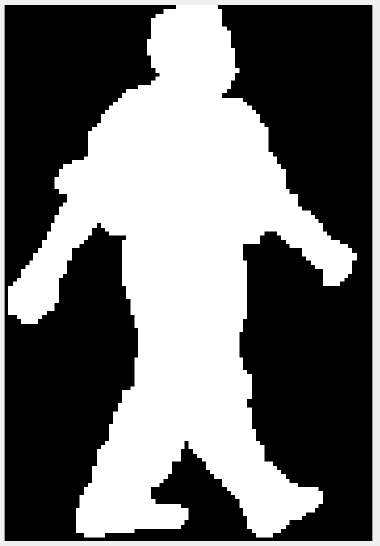}}
    
    \caption{Retrieval of sequences under noise on the OU-ISIR gait dataset. An image sequence of length $103$ is learned. Each image has size $88\times 128$. In (a) and (b), $\mathbf{x}(t)$ and $\bm{\xi}(t)$ are shown respectively for $t=1,...,10$. In (b), $2000$ salt-and-pepper noises are added to the first image. The corrupted image is set to be the initial state of the network.}
    \label{fig:silhouette}
\end{figure*}

\begin{figure*}[h!]
\vspace{0.5cm} 
    \centering
    \subfloat[][Ground truth $\mathbf{x}(t)$]{
    \includegraphics[width=0.115\textwidth]{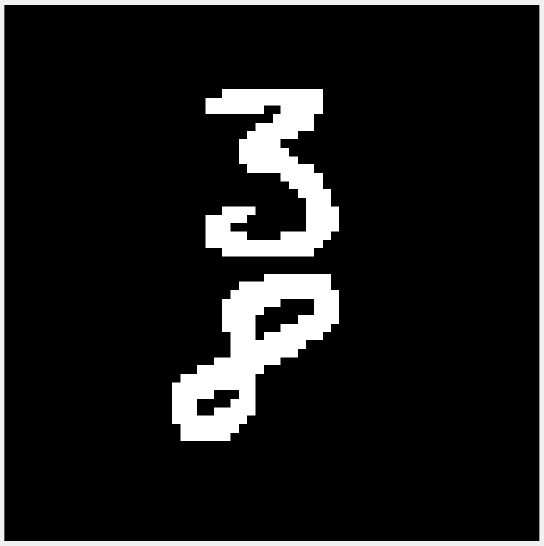}
    \includegraphics[width=0.115\textwidth]{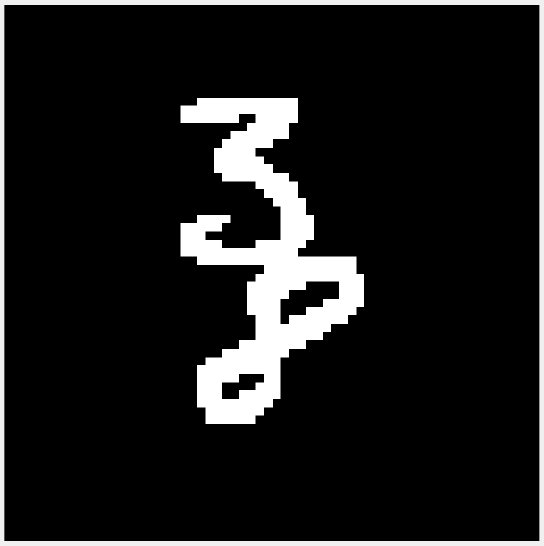}
    \includegraphics[width=0.115\textwidth]{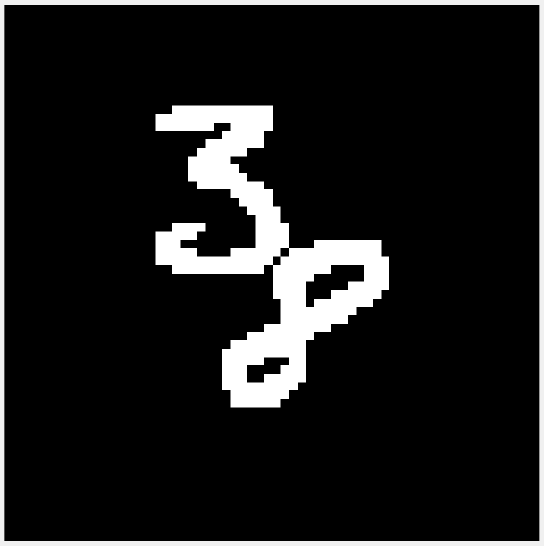}
    \includegraphics[width=0.115\textwidth]{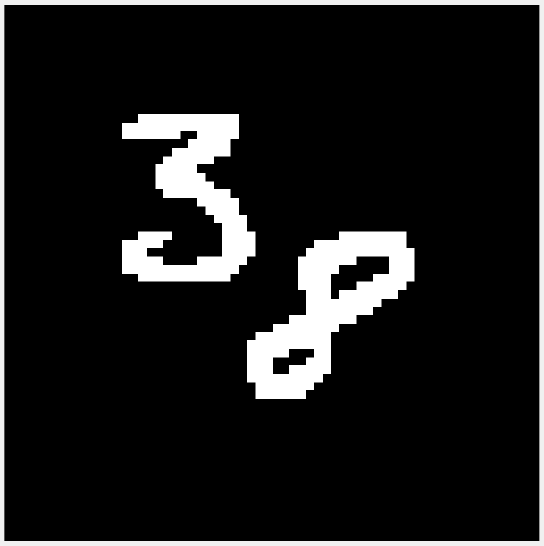}
    \includegraphics[width=0.115\textwidth]{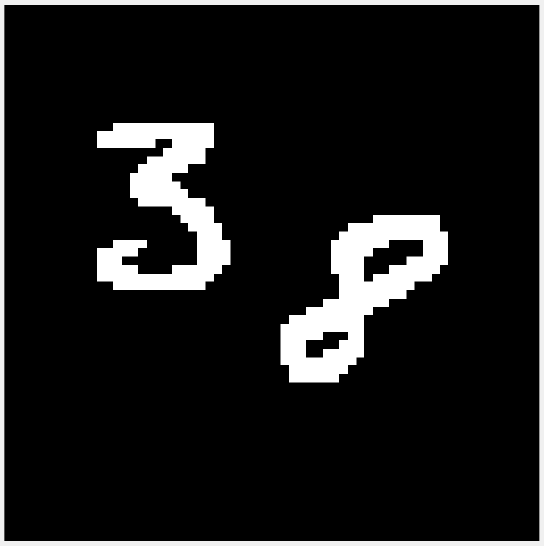}
    \includegraphics[width=0.115\textwidth]{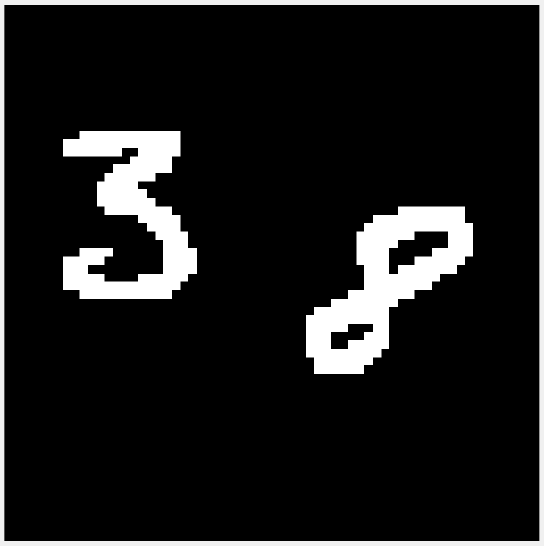}
    \includegraphics[width=0.115\textwidth]{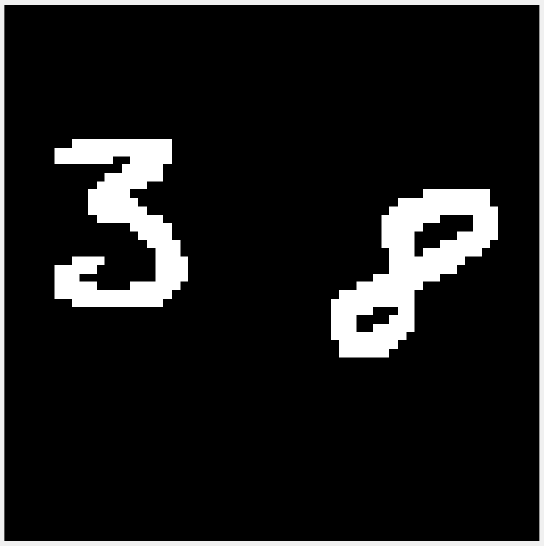}
    \includegraphics[width=0.115\textwidth]{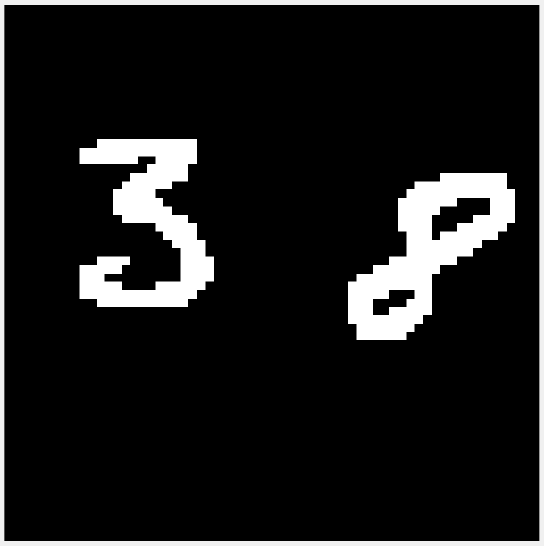}}

    \subfloat[][Network states $\bm{\xi}(t)$]{
    \includegraphics[width=0.115\textwidth]{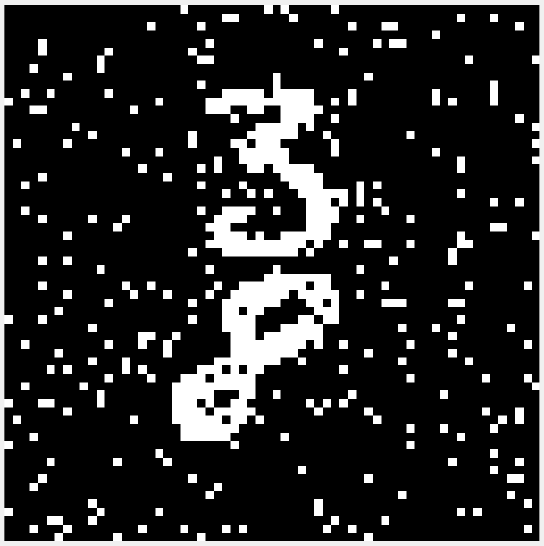}
    \includegraphics[width=0.115\textwidth]{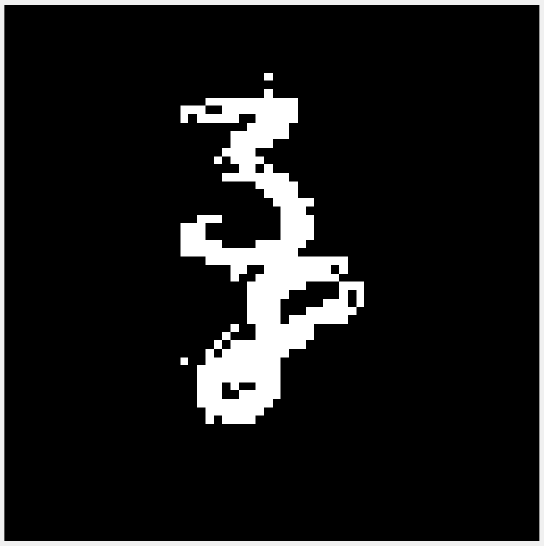}
    \includegraphics[width=0.115\textwidth]{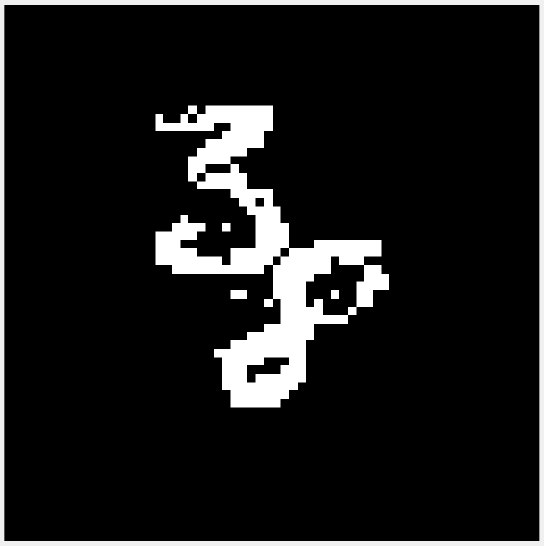}
    \includegraphics[width=0.115\textwidth]{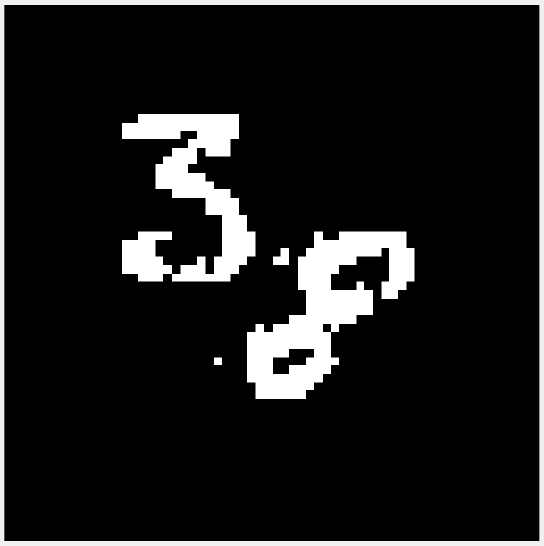}
    \includegraphics[width=0.115\textwidth]{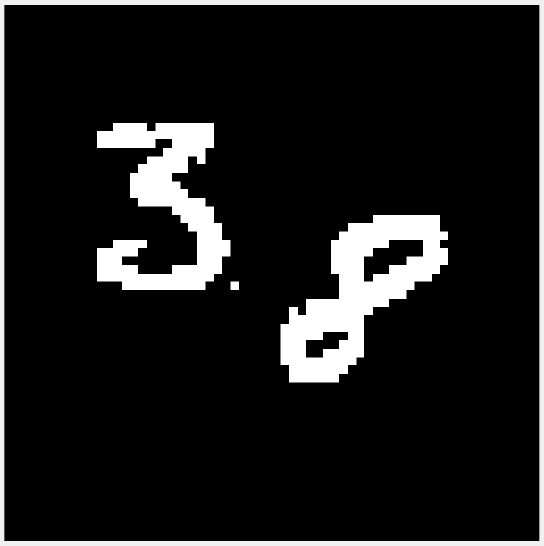}
    \includegraphics[width=0.115\textwidth]{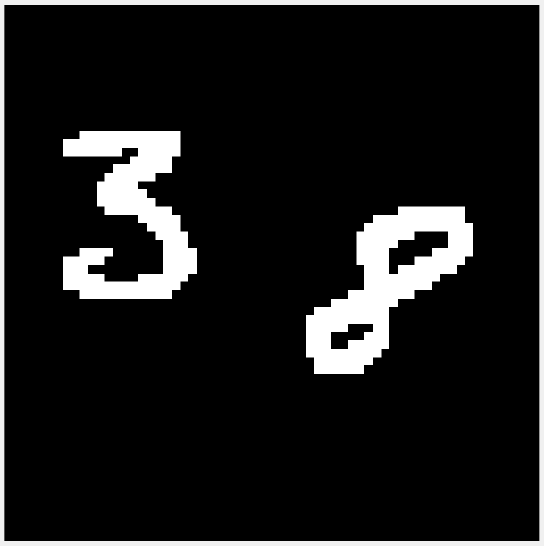}
    \includegraphics[width=0.115\textwidth]{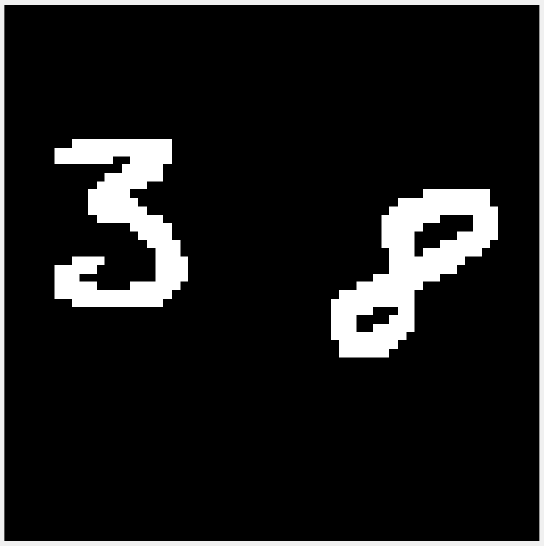}
    \includegraphics[width=0.115\textwidth]{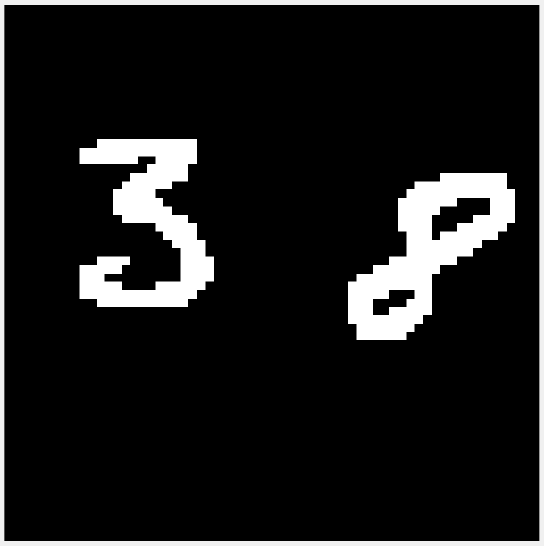}}
    
    \caption{Retrieval of sequences under noise on the Moving MNIST dataset. $20$ image sequences of length $20$ are learned. Due to space limitation, only one image sequence is displayed in here. Each image has size $64\times 64$. In (a) and (b), $\mathbf{x}(t)$ and $\bm{\xi}(t)$ are shown respectively for $t=1,...,8$. In (b), $300$ salt-and-pepper noises are added to the first image. The corrupted image is set to be the initial state of the network.}
    \label{fig:mnist}
\end{figure*}

For the OU-ISIR gait dataset, we used a network with hidden neuron number $M = 200$ to learn a single image sequence of length $103$, in which each image has size $88\times 128$. The images were flatten to vectors of size $88 \times 128 = 11264$.
For the Moving MNIST dataset, we used a network with hidden neuron number $M = 1000$ to learn $20$ image sequences of length $20$, in which each image has size $64\times 64$. The images were flatten to vectors of size $64 \times 64 = 4096$.
In Figure \ref{fig:silhouette} and \ref{fig:mnist}, we show the visualization results of the learned networks for robust retrieval, in which the first image of a sequence was corrupted and set to be the initial state of the network. 

\begin{figure*}[h!]

\vspace{0.5cm}
    \centering
    \subfloat[OU-ISIR gait]{
    \includegraphics[width=0.4\textwidth]{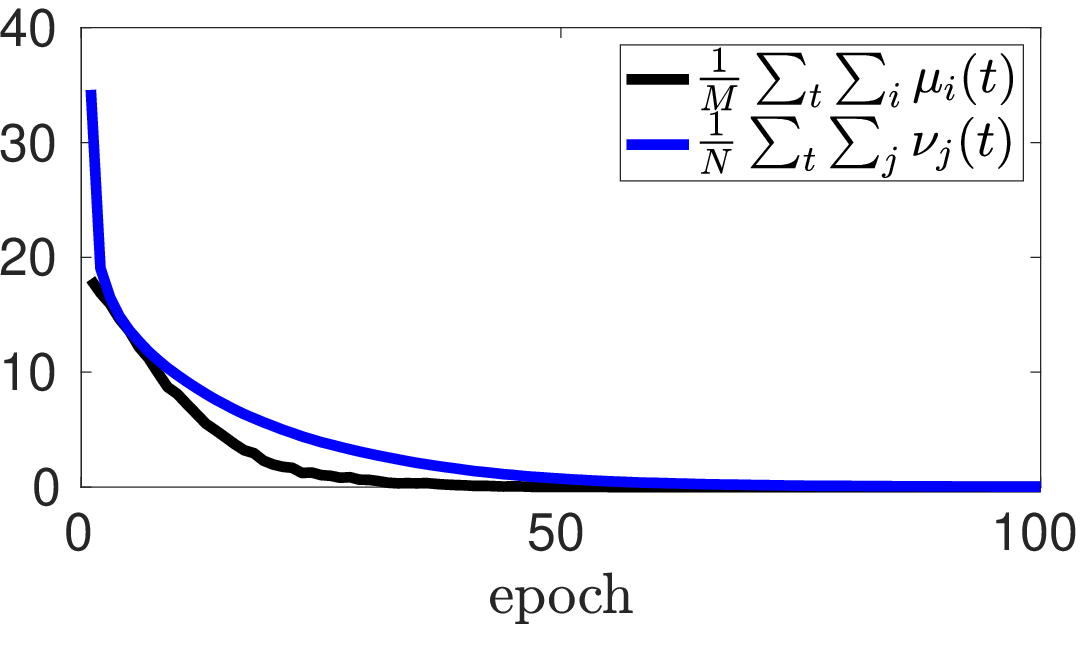}}
    \hspace{1cm}
    \subfloat[Moving MNIST]{
    \includegraphics[width=0.4\textwidth]{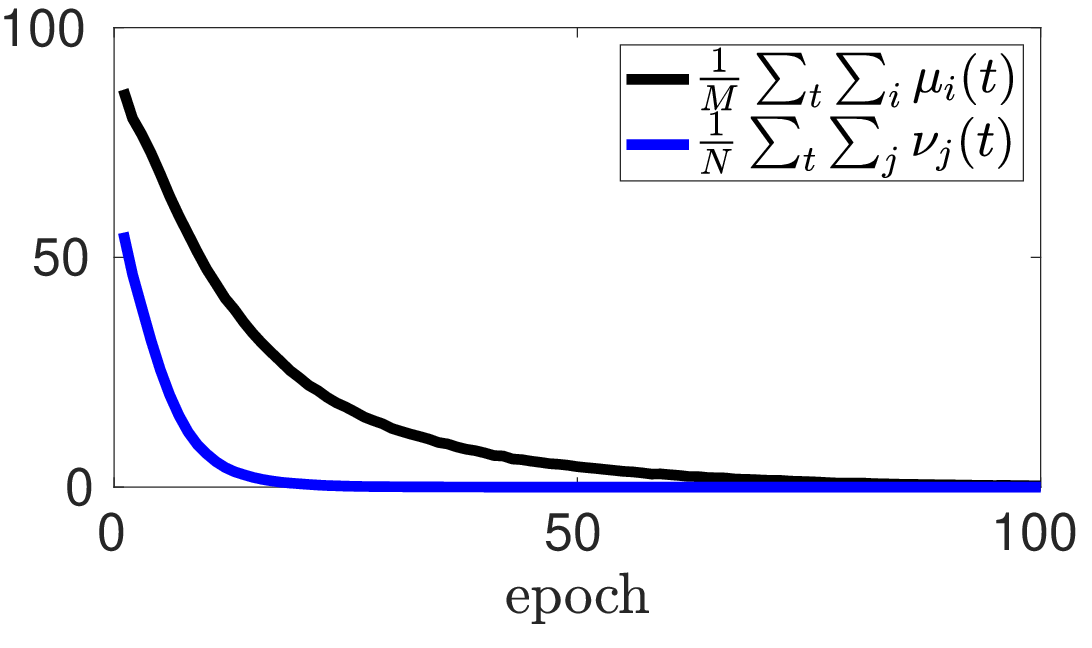}}
    \caption{Errors during learning. $\frac{1}{M} \sum_t\sum_i \mu_i(t)$ is the average error for the hidden neurons. $\frac{1}{N} \sum_t\sum_j \nu_j(t)$ is the average error for the visible neurons.}
    \label{fig:learning_errors}
\end{figure*}

In Figure \ref{fig:learning_errors}, we show the average errors $\frac{1}{M}\sum_t\sum_i \mu_i(t)$ and $\frac{1}{N}\sum_t\sum_j \nu_j(t)$ during the learning process, from which we can see that both errors reduce to zero smoothly. This result demonstrates that $\mathbf{U}$ and $\mathbf{V}$ can be cooperatively learned instead of conflicting with each other during learning.

\section{Discussion and Conclusion}


In this paper, we have investigated how recurrent networks of binary neurons learn sequence attractors to represent temporal sequence information. 
We showed that to store arbitrary sequence patterns, it is necessary for the networks to include hidden neurons. We developed a local learning algorithm and demonstrated that our model works well on synthetic and real-world datasets.  Thus, our work provides a possible biologically plausible mechanism in elucidating sequence memory in the brain.
In our model, hidden neurons are not directly involved in expressing pattern sequences. Instead, their contribution is on facilitating the storing and retrieving of pattern sequences. The indirect but indispensable role of hidden neurons may have a far-reaching implication to neural information processing.

Modern Hopfield networks also employ hidden neurons for sequence learning  \citep{chaudhry2023long}. In comparison, our approach is different in several aspects. First, our network model requires only the threshold activation function, while modern Hopfield networks require a polynomial or exponential activaton function for the hidden neurons. Second, the weights in our network model are learned from scratch in an online manner  while modern Hopfield networks require to store the sequence patterns explicitly as the weights in a predefined manner.
Third, our learning algorithm requires only local information between neurons, while modern Hopfield networks require the pseudo-inverse computation. Thus, our approach provides an alternative mechanism of sequence memory to modern Hopfield networks.

To pursue theoretical analysis, we have employed a very simple network model with binary neurons and threshold dynamics. From this simple model, we can get some insights into the neural mechanisms of sequence processing in the brain (as the classical Hopfield networks to static memories), but this simplification also incurs limitations. To fully validate our results, further researches with biologically more plausible models are needed, which include, for instances, biologically more realistic neuron models, synapses models, connection structures, learning rules and form of pattern sequences. Additionally, we look forward to seeing improvement of our learning algorithm for larger network capacity and robustness.

\section*{Acknowledgements}
This work was supported by the Science and Technology Innovation 2030-Brain Science and Brain-inspired Intelligence Project (No. 2021ZD0200204).

\clearpage

\bibliographystyle{apalike}
\bibliography{seq_attractor.bib}

\appendix
\onecolumn

\clearpage

\section{Proof of Theorem 1}

We construct a network such that, given $\bm{\xi}(t)=\mathbf{x}(i)$ for $i=1,...,T-1$, the hidden neurons provide an one-hot encoding of the successive pattern $\mathbf{x}(i+1)$, which is then decoded to be $\bm{\xi}(t+1)$.

To store $\mathbf{x}(1),...,\mathbf{x}(T) \in \{-1,1\}^N$ in (\ref{model:1})(\ref{model:2}), assuming  $\mathbf{x}(i)\neq\mathbf{x}(j)$ for $i\neq j$ except that $\mathbf{x}(1)=\mathbf{x}(T)$, let $M=T-1$ and construct  weight matrix $\mathbf{U}$ as
\begin{align}
    \mathbf{U} = (\mathbf{x}(1), \mathbf{x}(2),...,\mathbf{x}(T-1))^\top    
\end{align}
and hidden neurons $\bm{\zeta}(t)=(\zeta_1(t),...,\zeta_M(t))^\top$ as 
\begin{align}
    \zeta_i(t) &= \text{sign}\Big(\sum_{k=1}^N U_{ik} \xi_k(t) - N\Big) \\
    &=\text{sign}\big( \mathbf{x}(i)^\top \bm{\xi}(t) - N \big)
\end{align}
such that given $\bm{\xi}(t) = \mathbf{x}(i)$ for $i=1,...,T-1$, we have 
\begin{align}
    \zeta_j(t) = \begin{cases}
        \ \ \ 1, &\text{if }j=i, \\
        -1, &\text{otherwise}.
    \end{cases}
\end{align}
Next, we construct the weight matrix $\mathbf{V}$ as
\begin{align}
    \mathbf{V} = (\mathbf{x}(2), \mathbf{x}(3),...,\mathbf{x}(T) )  
\end{align}
and visible neurons $\bm{\xi}(t+1) =(\xi_1(t+1),...,\xi_N(t+1))^\top$ as
\begin{align}
    \bm{\xi}(t+1) = \text{sign}(\mathbf{V}\bm{\zeta}(t) + \bm{\theta})
\end{align}
where $\bm{\theta} = \sum_{j=2}^T \mathbf{x}(j)$ such that given the one-hot vector $\bm{\zeta}(t)$ we have 
\begin{align}
\bm{\xi}(t+1) &= \text{sign}\Big(\mathbf{x}(i+1) - \sum_{j\neq i+1}\mathbf{x}(j) + \sum_{j=2}^T \mathbf{x}(j)\Big) \\
&= \text{sign}(2\cdot\mathbf{x}(i+1) ) \\
&= \mathbf{x}(i+1)
\end{align}


\section{Proof of Theorem 2}

Note that the update of $\mathbf{U}$ (4)(5)(6) in Section 5 does not depend on $\mathbf{V}$. Therefore, we first prove the convergence of updating $\mathbf{U}$ for  $\eta > 0$ and $\kappa > 0$. The proof follows from \citep{gardner1988space}. Assume $\mathbf{U}^*$ exists such that, for all $t$ and $i$,
\begin{align}
z_i(t+1)\sum_k U^*_{ik} x_k(t) \geq \kappa.\label{eq:kappa2}
\end{align}
Define the $p$-th update of $\mathbf{U}$ with $\mu_i(t_p)=1$ by
\begin{align}
U_{ij}^{(p+1)} = U_{ij}^{(p)} + \eta z_i(t_p+1)x_j(t_p)
\end{align}
for some $t_p \in \{1,...,T-1\}$ and all $j$ in parallel. We assume zero-initialization, that is, $U_{ij}^{(1)} = 0$ for simplicity but the result holds if $|U_{ij}^{(1)}|$ is sufficiently small. Let 
\begin{align}
X_i^{(p+1)} = \frac{\sum_j U_{ij}^{(p+1)}U_{ij}^{*}}{\sqrt{\sum_j\big(U_{ij}^{(p+1)}\big)^2} \sqrt{\sum_j \big(U_{ij}^{*}\big)^2}}.
\end{align}
The Cauchy-Schwarz inequality, we have $X_i^{(p+1)} \leq 1$. Now we prove the convergence of updating $\mathbf{U}$ by contradiction. Assuming the update of $\mathbf{U}$ does not converge, we will show that $X_i^{(p+1)} > 1$ as $p\to \infty$. First, we have
\begin{align}
    \sum_j U_{ij}^{(p+1)}U_{ij}^{*} - \sum_j U_{ij}^{(p)}U_{ij}^{*} = \eta \sum_j z_i(t_p+1)U_{ij}^{*}x_j(t_p) \geq \eta\kappa 
\end{align}
due to (\ref{eq:kappa2}) and therefore
\begin{align}
    \sum_j U_{ij}^{(p+1)}U_{ij}^{*} &= \sum_j U_{ij}^{(p+1)}U_{ij}^{*} - \sum_j U_{ij}^{(p)}U_{ij}^{*} +...+ \sum_j U_{ij}^{(2)}U_{ij}^{*} - \sum_j U_{ij}^{(1)}U_{ij}^{*} + \sum_j U_{ij}^{(1)}U_{ij}^{*} \\
    &\geq \eta\kappa p 
\end{align}
since we assumed $U_{ij}^{(1)} = 0$. Next, we have 
\begin{align}
\sum_j \big(U_{ij}^{(p+1)}\big)^2 - \sum_j \big(U_{ij}^{(p)}\big)^2 &= \sum_j \big(U_{ij}^{(p)} + \eta z_i(t_p+1)x_j(t_p)\big)^2 - \sum_j \big(U_{ij}^{(p)}\big)^2 \\
&= 2\eta \sum_j  U_{ij}^{(p)} z_i(t_p+1)x_j(t_p)
+N\eta^2 \\
&= 2\eta z_i(t_p+1)\sum_j  U_{ij}^{(p)} x_j(t_p)
+N\eta^2 \\
&< 2\eta \kappa + N\eta^2
\end{align}
since we assumed $\mu_i(t_p)=1$ and therefore $z_i(t_p+1)\sum_j  U_{ij}^{(p)} x_j(t_p) < \kappa$.
Then, we have
\begin{align}
&\sqrt{\sum_j (U_{ij}^{(p+1)})^2} - \sqrt{\sum_j (U_{ij}^{(p)})^2}\\ 
= & \Big(\sum_j \big(U_{ij}^{(p+1)}\big)^2 - \sum_j \big(U_{ij}^{(p)}\big)^2 \Big) \Big/ \Big( \sqrt{\sum_j \big(U_{ij}^{(p+1)}\big)^2} + \sqrt{\sum_j \big(U_{ij}^{(p)}\big)^2} \Big)\\
< & (2\eta \kappa + N\eta^2) \Big/ \Big( \sqrt{\sum_j \big(U_{ij}^{(p+1)}\big)^2} + \sqrt{\sum_j \big(U_{ij}^{(p)}\big)^2} \Big). \label{eq:ine}
\end{align}
By Cauchy-Schwarz inequality, we have 
\begin{align}
\sqrt{\sum_j (U_{ij}^{(p+1)})^2} \sqrt{\sum_j (U_{ij}^{*})^2}
\geq \sum_j U_{ij}^{(p+1)}U_{ij}^{*} \geq \eta\kappa p 
\end{align}
and therefore
\begin{align}
\sqrt{\sum_j (U_{ij}^{(p+1)})^2} 
\geq \frac{\eta\kappa p}{\sqrt{\sum_j (U_{ij}^{*})^2}}.
\end{align}
Also,
\begin{align}
\sqrt{\sum_j (U_{ij}^{(p+1)})^2} &= \sqrt{\sum_j (U_{ij}^{(p+1)})^2} - \sqrt{\sum_j (U_{ij}^{(p)})^2} + ... + \sqrt{\sum_j (U_{ij}^{(2)})^2} - \sqrt{\sum_j (U_{ij}^{(1)})^2} \\
&+ \sqrt{\sum_j (U_{ij}^{(1)})^2} \\
&< \sum_{q=1}^p (2\eta \kappa + N\eta^2) \Big/ \Big( \sqrt{\sum_j \big(U_{ij}^{(q+1)}\big)^2} + \sqrt{\sum_j \big(U_{ij}^{(q)}\big)^2} \Big)  \\
&< \sum_{q=1}^p(2\eta\kappa + N\eta^2)\sqrt{\sum_j (U_{ij}^{*})^2} \frac{1}{\eta\kappa(2q-1) }\\
&=\frac{\eta\kappa + N\eta^2/2}{\eta\kappa}\sqrt{\sum_j (U_{ij}^{*})^2} \sum_{q=1}^p \frac{1}{ q-1/2 }
\end{align}
due to (\ref{eq:ine}) and $U_{ij}^{(1)} = 0$.
Note that for $q > 1$
\begin{align}
\frac{1}{q - 1/2} \leq  \int_{q-3/2}^{q-1/2}  \frac{1}{x}dx = \log(q-1/2) - \log(q-3/2) 
\end{align}
and
\begin{align}
\sum_{q=1}^p\frac{1}{q - 1/2} = \frac{1}{2} + \sum_{q=2}^p\frac{1}{q - 1/2} \leq  \frac{1}{2} + \int_{1/2}^{p-1/2} \frac{1}{x}dx = 2 + \log(p-1/2) - \log(1/2).
\end{align}
Therefore, 
\begin{align}
    \sqrt{\sum_j (U_{ij}^{(p+1)})^2} = O(\log(p))
\end{align}
and
\begin{align}
     \sum_j U_{ij}^{(p+1)}U_{ij}^{*} = \Omega(p)
\end{align}
as $p\to\infty$. We have,
\begin{align}
    X_i^{(p+1)} = \frac{\sum_j U_{ij}^{(p+1)}U_{ij}^{*}}{\sqrt{\sum_j (U_{ij}^{(p+1)})^2} \sqrt{\sum_j (U_{ij}^{*})^2}} > 1
\end{align}
for some $p$. This contradicts that $X_i^{(p+1)} \leq 1$. Thus, the updating $\mathbf{U}$ converges.

Upon the convergence of updating $\mathbf{U}$, we can prove the convergence of $\mathbf{V}$ if there exists $\mathbf{V}^*$ such that 
 for all $t$ and $i$,
\begin{align}
x_i(t+1)\sum_k V^*_{ik} y_k(t) \geq \kappa
\end{align}
by a similar proof.

\section{Proof of Theorem 3}

If $\mu_i(t)=0$, then $U_{ik}'=U_{ik}$ and $\mu_i'(t) = \mu_i(t) = 0$. If $\mu_i(t)=1$, then 
\begin{align}
\mu_i'(t) &= H\Big(\kappa - z_i(t+1)\sum_{k=1}^N \big(U_{ik} + \eta z_i(t+1)x_k(t)\big) x_k(t)\Big) \\
&= H\Big(\kappa - z_i(t+1)\sum_{k=1}^N U_{ik}x_k(t) - \eta \big(z_i(t+1)\big)^2\sum_{k=1}^N \big(x_k(t)\big)^2\Big) \\
&= H\Big(\kappa - z_i(t+1)\sum_{k=1}^N U_{ik}x_k(t) - \eta N \Big) = 0
\end{align}
for sufficiently large $\eta > 0$ given $x_k(t)=\pm 1$, $z_i(t+1)=\pm 1$ and the property of Heaviside function.

\section{Proof of Theorem 4}

If $\nu_j(t)=0$, then we have 
\begin{align}
     x_j(t+1)\sum_{k=1}^M V_{jk} y_k(t) \geq \kappa.
\end{align}
Next,
\begin{align}
     x_j(t+1)\sum_{k=1}^M V_{jk} \hat{y}_k(t) &= x_j(t+1)\sum_{k=1}^M V_{jk} \big({y}_k(t) + \hat{y}_k(t) - {y}_k(t)\big) \\
     &= x_j(t+1)\sum_{k=1}^M V_{jk} {y}_k(t) + x_j(t+1)\sum_{k=1}^M V_{jk} \big(\hat{y}_k(t) - {y}_k(t)\big)  \\
     &\geq \kappa + x_j(t+1)\sum_{k=1}^M V_{jk}\big(\hat{y}_k(t) - {y}_k(t)\big) \\
     &\geq \kappa - \Big| \sum_{k=1}^M V_{jk}\big(\hat{y}_k(t) - {y}_k(t)\big) \Big| \\
     &\geq \kappa -\max_k |V_{jk}|\sum_{k=1}^M  |\hat{y}_k(t) - {y}_k(t)| \\
     &> \kappa -\max_k |V_{jk}|  \cdot \epsilon
     > 0
\end{align}
since $x_j(t+1)=\pm 1$, which implies
\begin{align}
     x_j(t+1) = \text{sign}\Big(\sum_{k=1}^M V_{jk} \hat{y}_k(t)\Big).
\end{align}

\clearpage

\section{Numerical Results for Figure 5 (a) and (b)}

In the main paper, we only showed bar charts (Figure 5 (a) and (b)) of the results in Section 6.2. Here, for more information, we provide the numerical results for Figure 5 (a) in Table 1 and Figure 5 (b) in Table 2.

\begin{table}[h!]

\centering
\begin{tabular}{|c|c|c|c|c|c|c|c|c|c|c|c|c|c|c|c|}
\hline
$T$ & 10 & 20 & 30 & 40 & 50 & 60 & 70 & 80 & 90 & 100 & 110 & 120 & 130 & 140 & 150 \\
\hline
Learning only $\mathbf{V}$ &  100  & 100 &  100  &  91  &  66  &  19  &   8  &   2  &   0     & 0 &    0  &   0  &   0 &    0   &  0\\
\hline
Learning $\mathbf{U}$ and $\mathbf{V}$ & 100  & 100  & 100 &  100 &  100  &  99  &  88 &   52  &  20  &   1  &   0   &  0  &   0  &   0  &   0 \\
\hline
\end{tabular}
\caption{Successful retrievals out of 100 trials with different sequence period lengths $T$.}
\end{table}

\begin{table}[h!]

\centering
\begin{tabular}{|c|c|c|c|c|c|c|c|c|c|c|}
\hline
$M$ & 100 & 200 & 300 & 400 & 500 & 600 & 700 & 800 & 900 & 1000 \\
\hline
Learning only $\mathbf{V}$ &  0   &  0  &   1  &   3  &   6  &  16  &  14 &   27  &  30 &   37  \\
\hline
Learning $\mathbf{U}$ and $\mathbf{V}$ & 10  &  52  &  85  &  90  &  94  &  88  &  95 &   96  &  96  &  97  \\
\hline
\end{tabular}
\caption{Successful retrievals out of 100 trials with different numbers of hidden neurons $M$.}
\end{table}

\section{Ablation Experiments: Joint Learning of $\mathbf{U}$ and $\mathbf{V}$}

\noindent To verify the effective of the proposed learning algorithm in Section 5, we show additional experimental results in which three methods for the networks of hidden units in learning the sequences in Section 5.3 are compared.

\begin{enumerate}

    \item Fixing $\mathbf{U}$ and learning $\mathbf{V}$ by the temporal asymmetric Hebbian algorithm
    $$V_{ji} = \sum_t x_j(t+1) y_i(t)$$ where $$y_i(t) = \text{sign}\big(\sum_{k=1}^N U_{ik} x_k(t)\big).$$

    \item Fixing $\mathbf{U}$ and learning $\mathbf{V}$ with the three-factor rule (7)(8)(9) in Section 5.

    \item Learning both $\mathbf{U}$ and $\mathbf{V}$ with the three-factor rule  (4)(5)(6)(7)(8)(9) in Section 5.    
    
\end{enumerate}
The experimental settings are the same as in Section 6.3.  The results are shown in Figure 1-5, from which we can see the algorithm proposed in Section 5 is indeed effective.

\section{Ablation Experiments: Sparsity}

We provide some further ablation study of our algorithm on the effect of sparsity under the experimental settings of Section 6.2 in the main paper. 

\paragraph{Figure \ref{fig:sparse_U}: Sparse Random Projected Inputs}

We compare our method (learning both $\mathbf{U}$
 and $\mathbf{V}$
 with the three-factor rule) with using fixed random $\mathbf{U}$
 whose elements are sampled i.i.d. from the standard Gaussian distribution and learning only 
 with the three-factor rule. The sparse random projected inputs are defined as 
 $$y_i(t) = \text{sign}\Big(\sum_{k=1}^N U_{ik}x_k(t) - \theta \Big)$$
 where $\theta > 0$ controls the sparsity level.

\paragraph{Figure \ref{fig:sparse}: Sparse Random Projected Targets}

We test different levels of sparsity in the random projected targets defined as 
$$z_i(t=1) = \text{sign}\Big(\sum_{k=1}^N P_{ik}x_k(t+1) - \theta \Big)$$
where $\theta > 0$ controls the sparsity level.

From both sets of experiments, we do not find sparsity enlarges significantly the capacity of the networks in learning sequences as attractors.

\begin{figure}[b!]
    \centering
    \subfloat[][Ground truth]{
    \includegraphics[width=0.09\textwidth]{img/silhouette_1_gt.pdf}
    \includegraphics[width=0.09\textwidth]{img/silhouette_2_gt.pdf}
    \includegraphics[width=0.09\textwidth]{img/silhouette_3_gt.pdf}
    \includegraphics[width=0.09\textwidth]{img/silhouette_4_gt.pdf}
    \includegraphics[width=0.09\textwidth]{img/silhouette_5_gt.pdf}
    \includegraphics[width=0.09\textwidth]{img/silhouette_6_gt.pdf}
    \includegraphics[width=0.09\textwidth]{img/silhouette_7_gt.pdf}
    \includegraphics[width=0.09\textwidth]{img/silhouette_8_gt.pdf}
    \includegraphics[width=0.09\textwidth]{img/silhouette_9_gt.pdf}
    \includegraphics[width=0.09\textwidth]{img/silhouette_10_gt.pdf} 
    }
    
    \subfloat[][Fixing $\mathbf{U}$ and learning only $\mathbf{V}$ with temporal asymmetric Hebbian algorithm]{
    \includegraphics[width=0.09\textwidth]{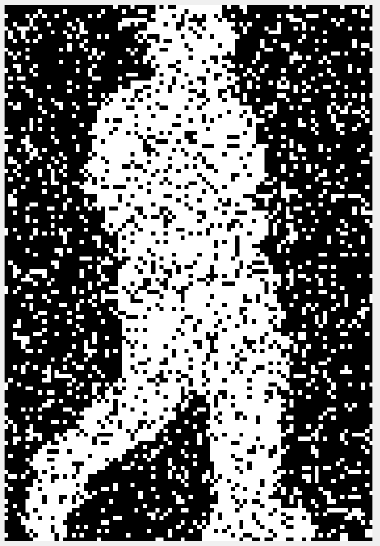}
    \includegraphics[width=0.09\textwidth]{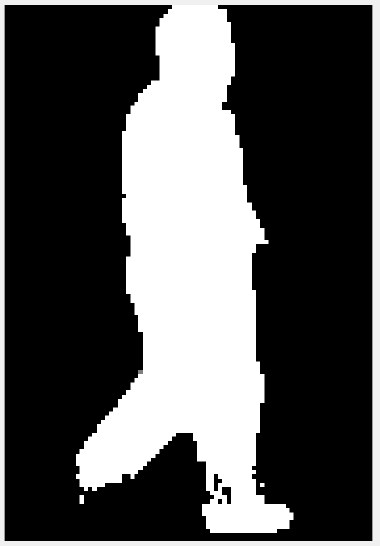}
    \includegraphics[width=0.09\textwidth]{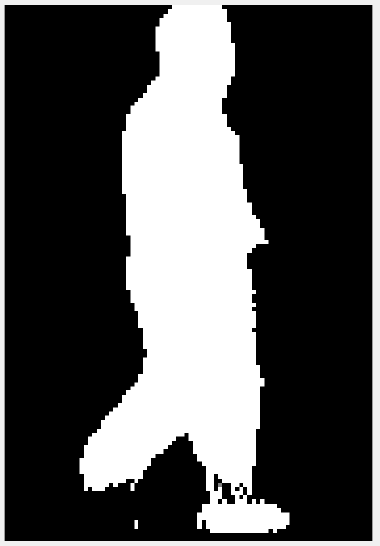}
    \includegraphics[width=0.09\textwidth]{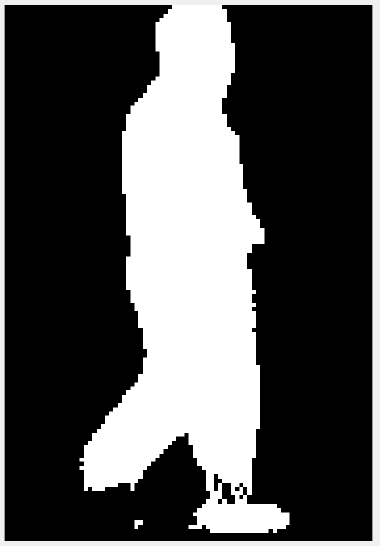}
    \includegraphics[width=0.09\textwidth]{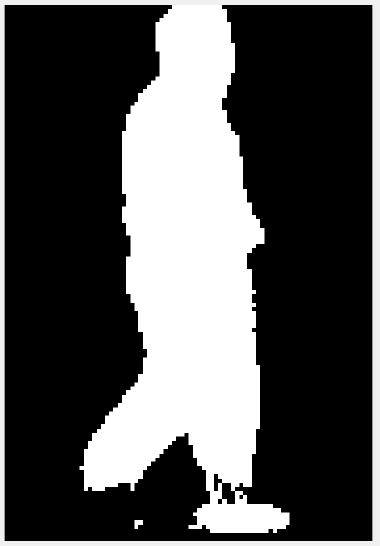}
    \includegraphics[width=0.09\textwidth]{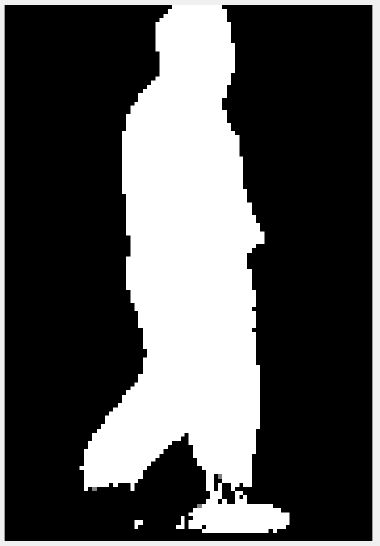}
    \includegraphics[width=0.09\textwidth]{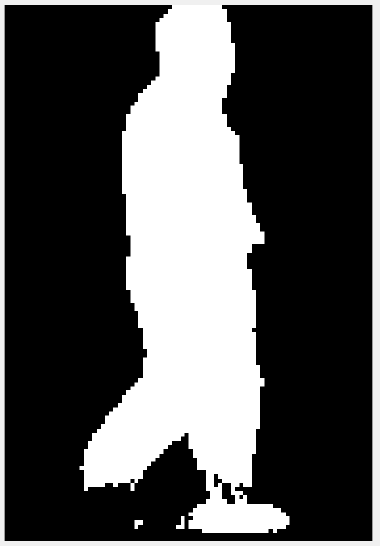}
    \includegraphics[width=0.09\textwidth]{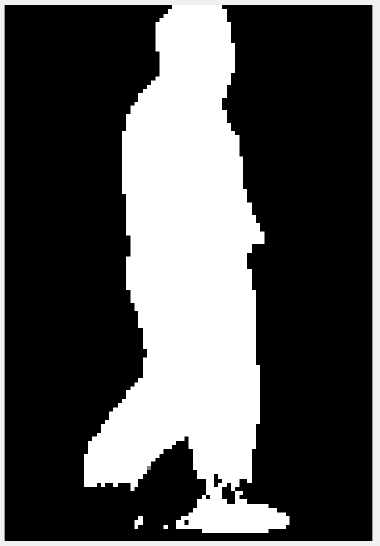}
    \includegraphics[width=0.09\textwidth]{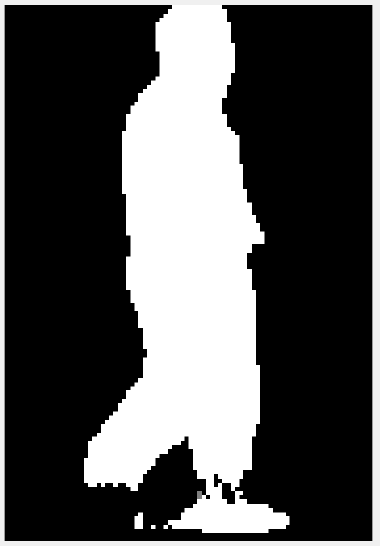}
    \includegraphics[width=0.09\textwidth]{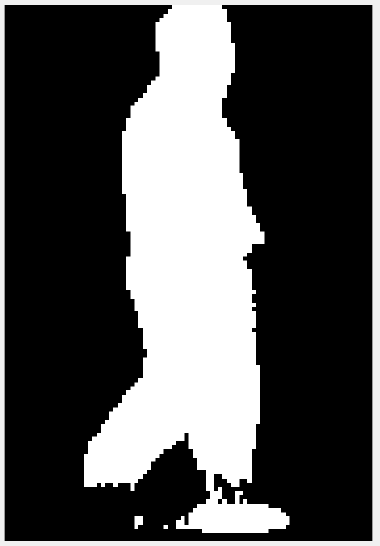}}

    \subfloat[][Fixing $\mathbf{U}$ and learning only $\mathbf{V}$ with three-factor rule]{
    \includegraphics[width=0.09\textwidth]{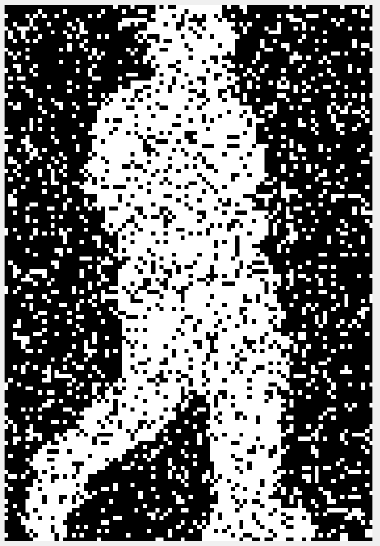}
    \includegraphics[width=0.09\textwidth]{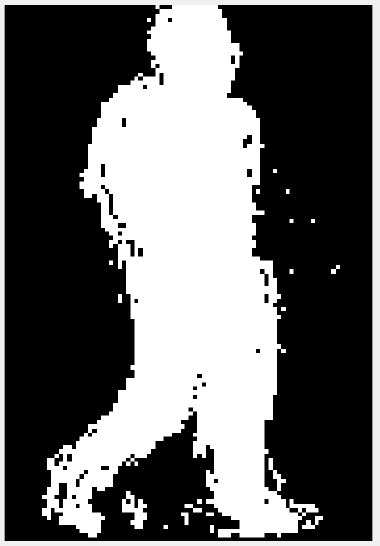}
    \includegraphics[width=0.09\textwidth]{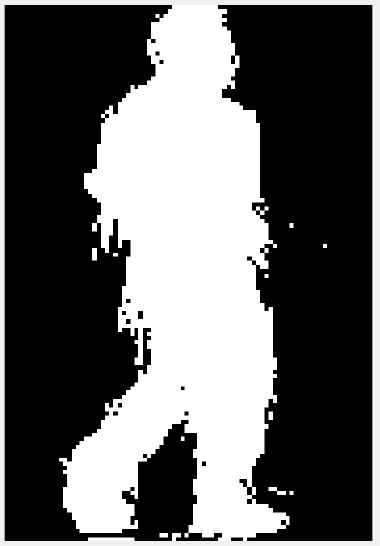}
    \includegraphics[width=0.09\textwidth]{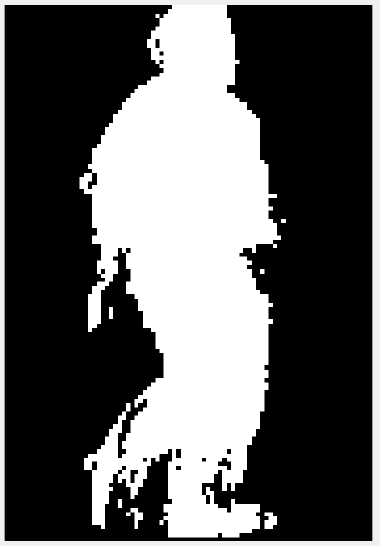}
    \includegraphics[width=0.09\textwidth]{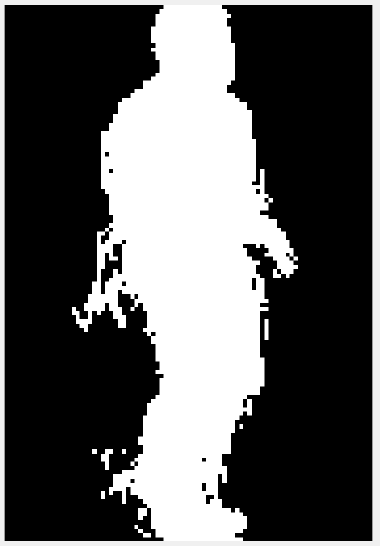}
    \includegraphics[width=0.09\textwidth]{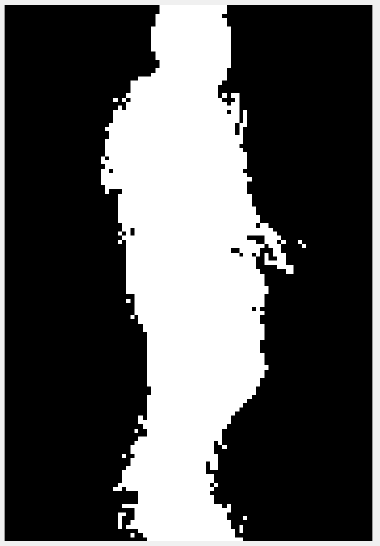}
    \includegraphics[width=0.09\textwidth]{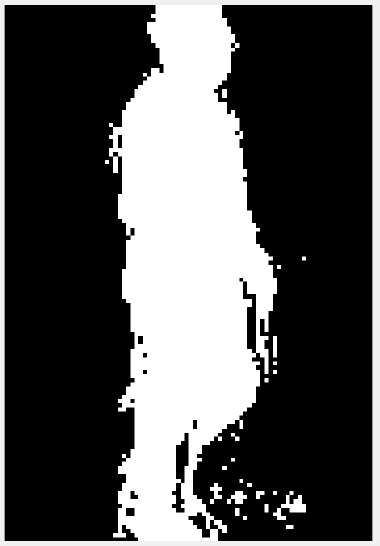}
    \includegraphics[width=0.09\textwidth]{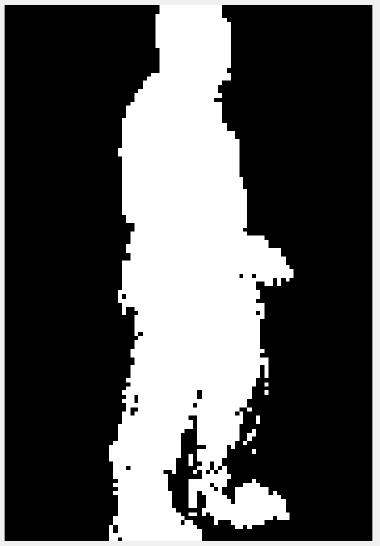}
    \includegraphics[width=0.09\textwidth]{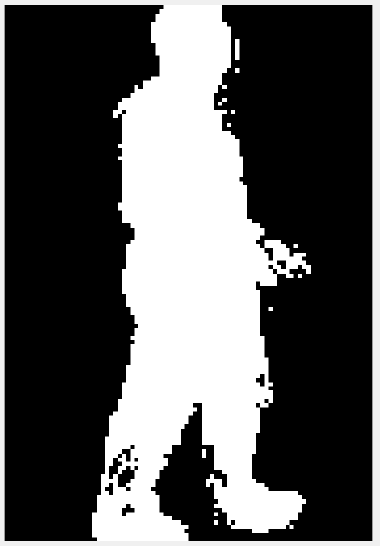}
    \includegraphics[width=0.09\textwidth]{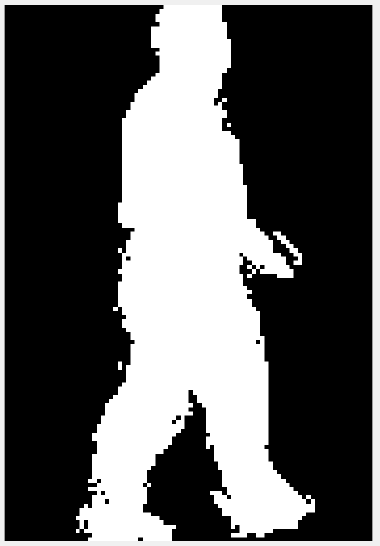}}    

    \subfloat[][Learning $\mathbf{U}$ and $\mathbf{V}$ with three-factor rule]{
    \includegraphics[width=0.09\textwidth]{img/silhouette_1_two.pdf}
    \includegraphics[width=0.09\textwidth]{img/silhouette_2_two.pdf}
    \includegraphics[width=0.09\textwidth]{img/silhouette_3_two.pdf}
    \includegraphics[width=0.09\textwidth]{img/silhouette_4_two.pdf}
    \includegraphics[width=0.09\textwidth]{img/silhouette_5_two.pdf}
    \includegraphics[width=0.09\textwidth]{img/silhouette_6_two.pdf}
    \includegraphics[width=0.09\textwidth]{img/silhouette_7_two.pdf}
    \includegraphics[width=0.09\textwidth]{img/silhouette_8_two.pdf}
    \includegraphics[width=0.09\textwidth]{img/silhouette_9_two.pdf}
    \includegraphics[width=0.09\textwidth]{img/silhouette_10_two.pdf}}

    \caption{OU-ISIR gait sequence for $t = 1,...,10$.}

\end{figure}

\begin{figure}[t!]
    \centering
    \subfloat[][Ground truth]{
    \includegraphics[width=0.1\textwidth]{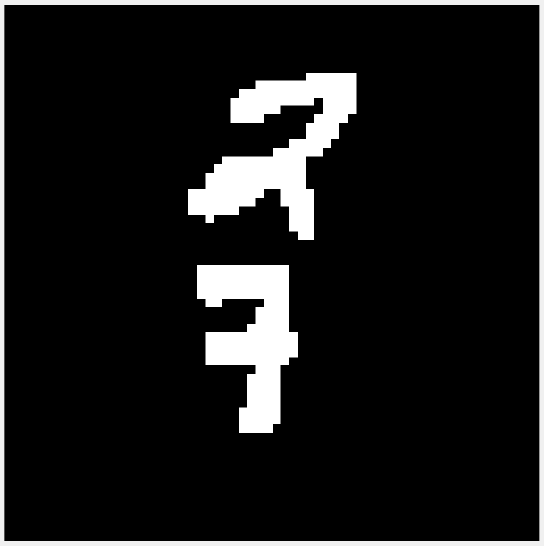}
    \includegraphics[width=0.1\textwidth]{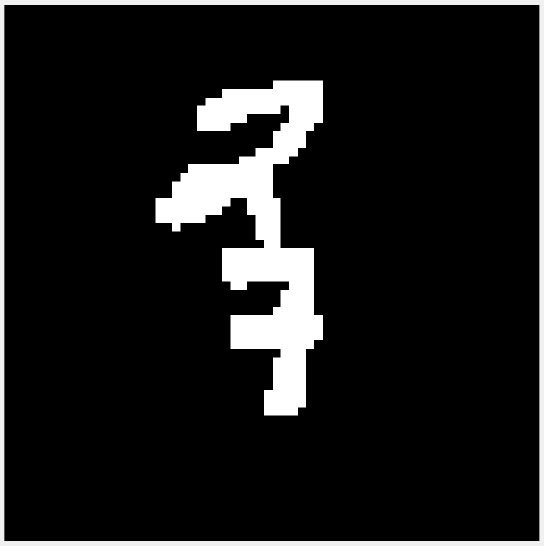}
    \includegraphics[width=0.1\textwidth]{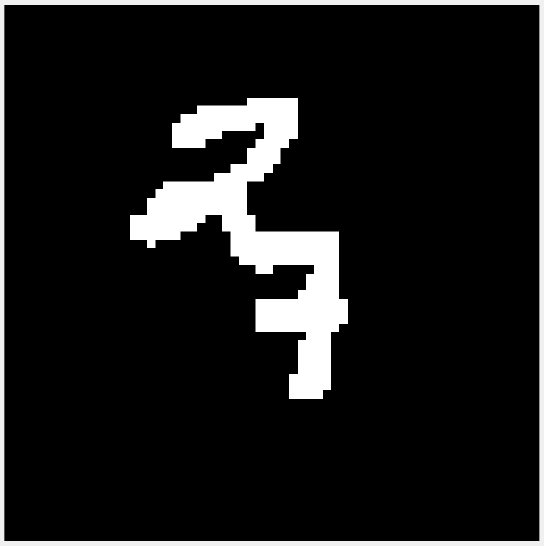}
    \includegraphics[width=0.1\textwidth]{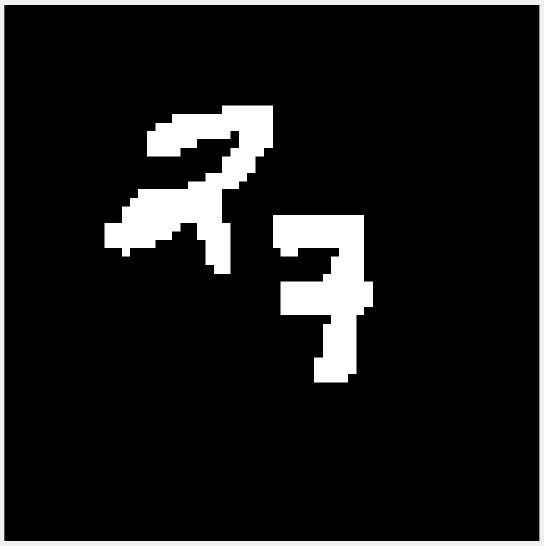}
    \includegraphics[width=0.1\textwidth]{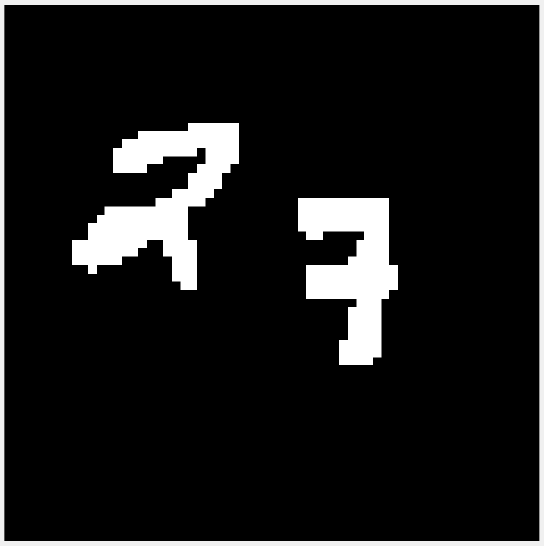}
    \includegraphics[width=0.1\textwidth]{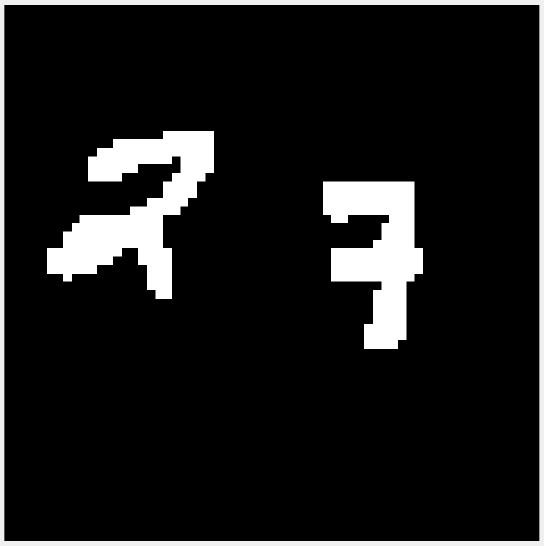}
    \includegraphics[width=0.1\textwidth]{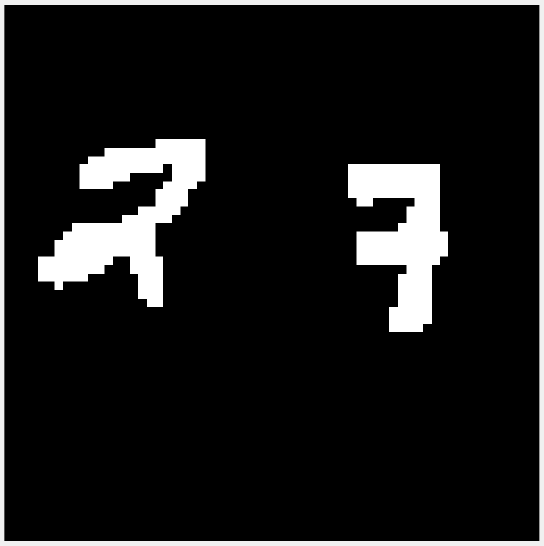}
    \includegraphics[width=0.1\textwidth]{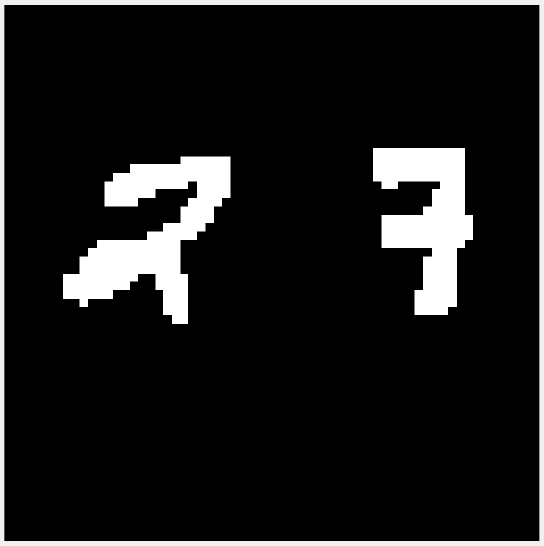}
    }

    \subfloat[][Fixing $\mathbf{U}$ and learning only $\mathbf{V}$ with temporal asymmetric Hebbian algorithm]{
    \includegraphics[width=0.1\textwidth]{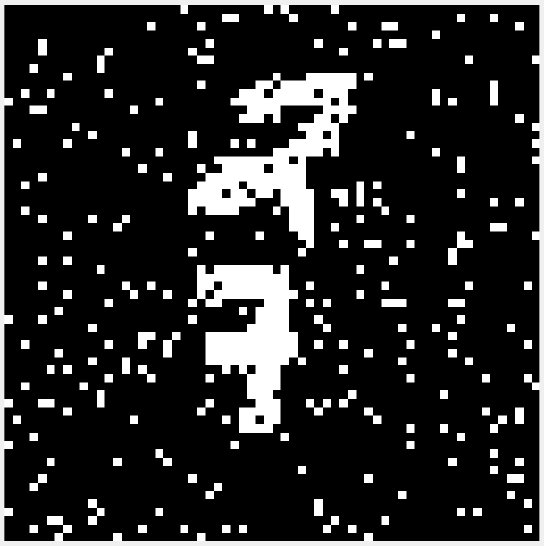}
    \includegraphics[width=0.1\textwidth]{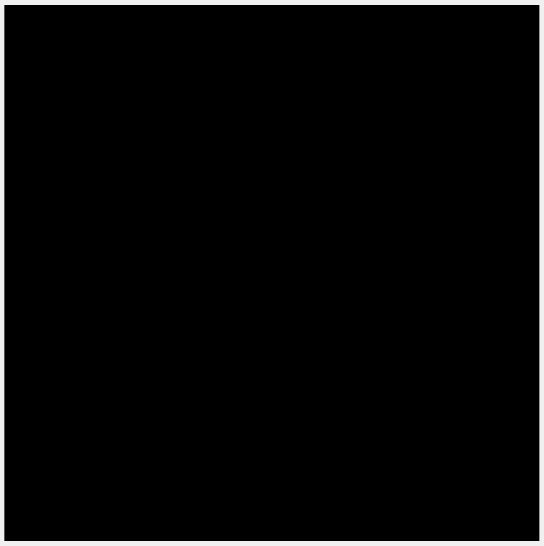}
    \includegraphics[width=0.1\textwidth]{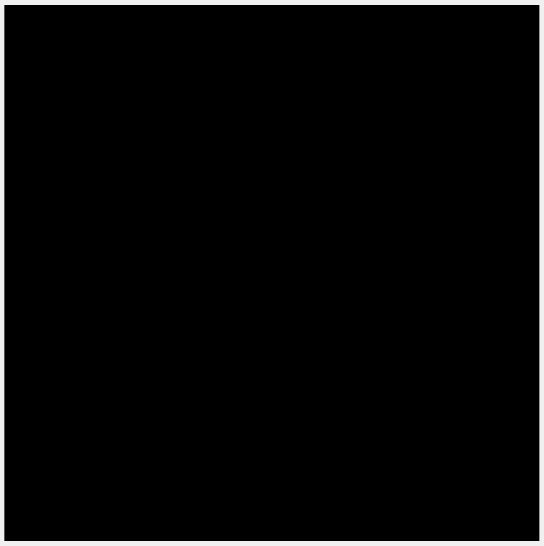}
    \includegraphics[width=0.1\textwidth]{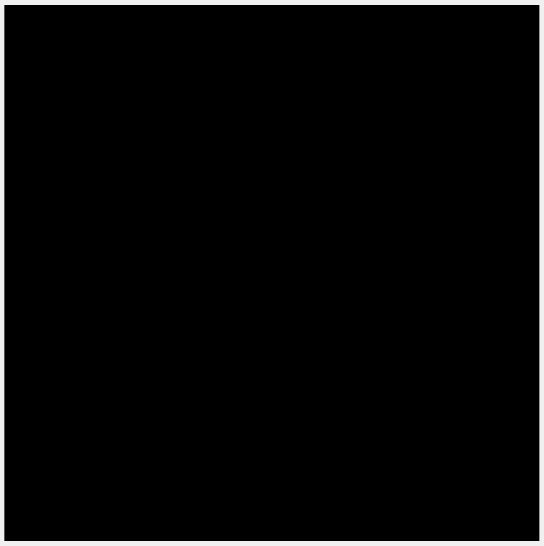}
    \includegraphics[width=0.1\textwidth]{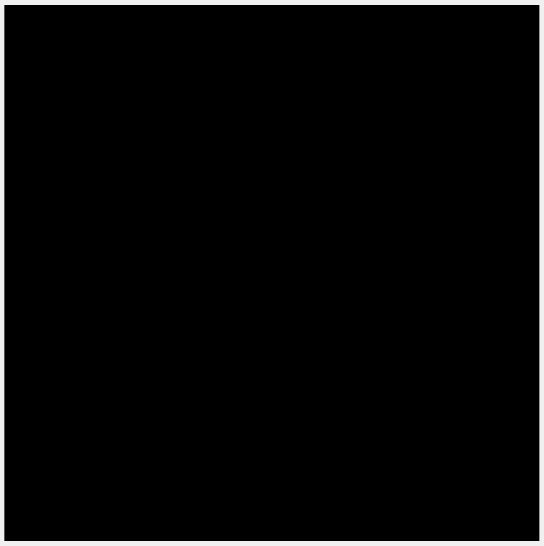}
    \includegraphics[width=0.1\textwidth]{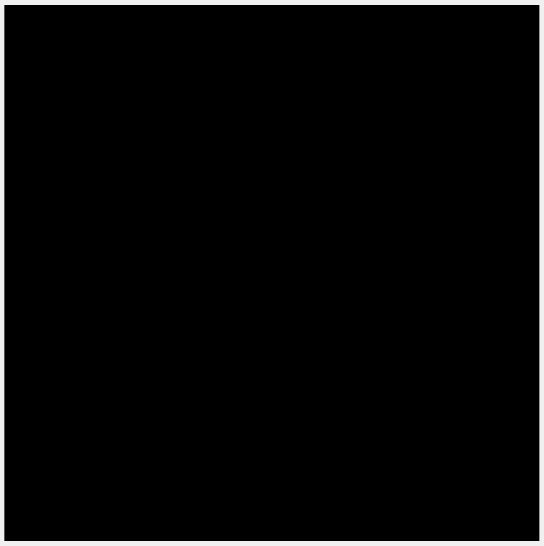}
    \includegraphics[width=0.1\textwidth]{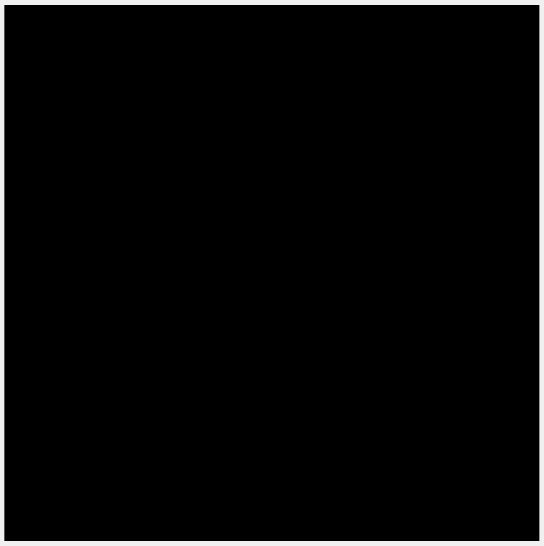}
    \includegraphics[width=0.1\textwidth]{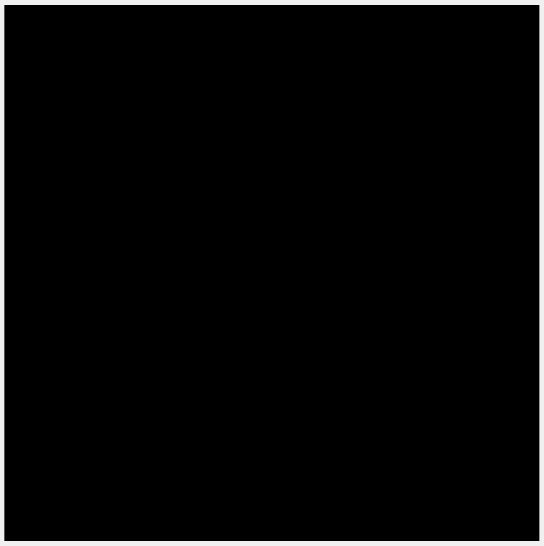}
    }

    \subfloat[][Fixing $\mathbf{U}$ and learning only $\mathbf{V}$ with three-factor rule]{
    \includegraphics[width=0.1\textwidth]{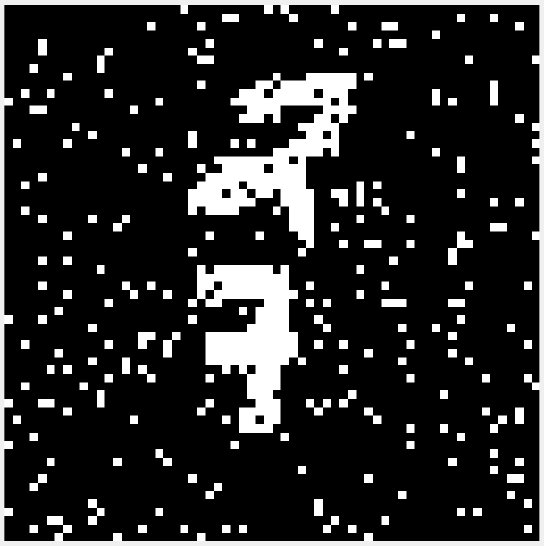}
    \includegraphics[width=0.1\textwidth]{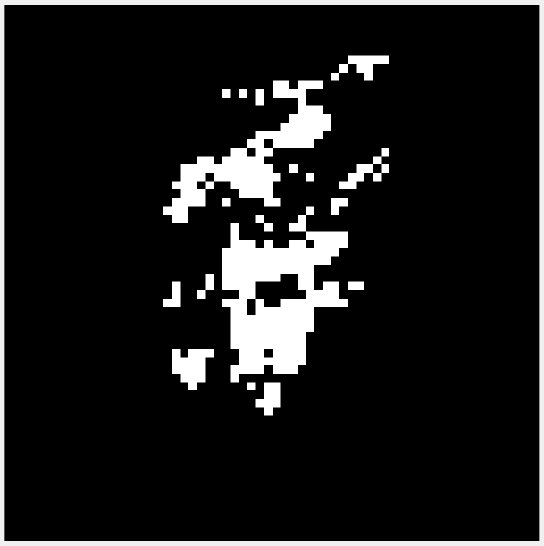}
    \includegraphics[width=0.1\textwidth]{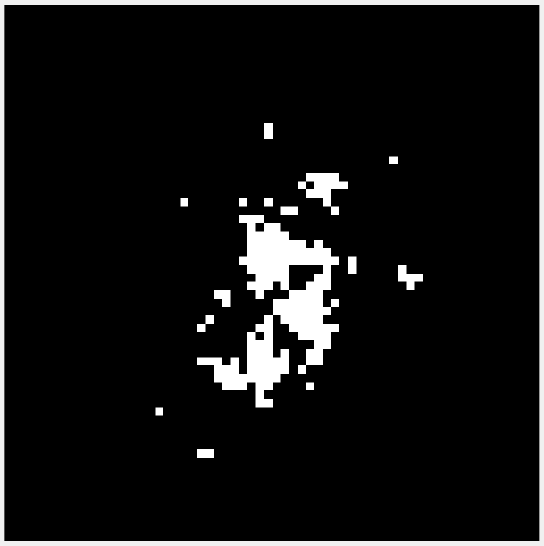}
    \includegraphics[width=0.1\textwidth]{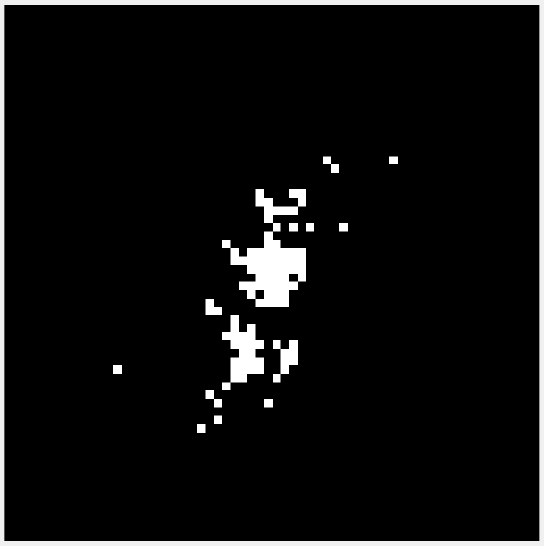}
    \includegraphics[width=0.1\textwidth]{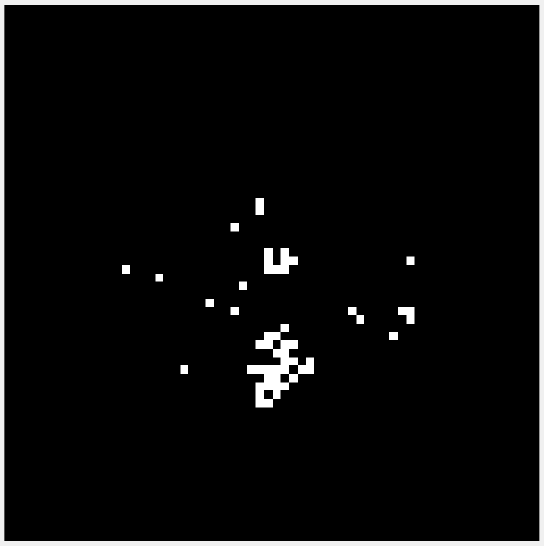}
    \includegraphics[width=0.1\textwidth]{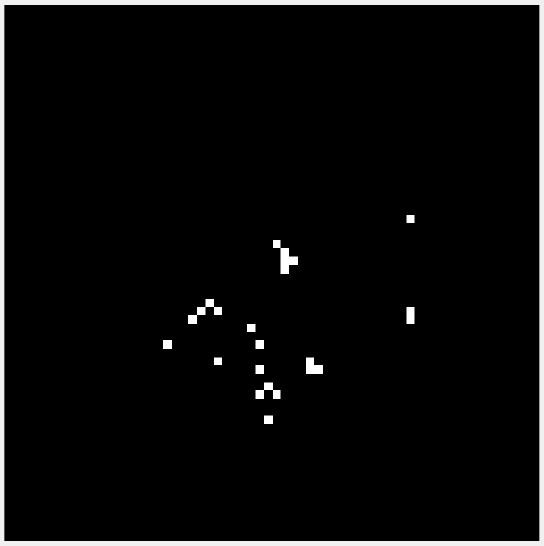}
    \includegraphics[width=0.1\textwidth]{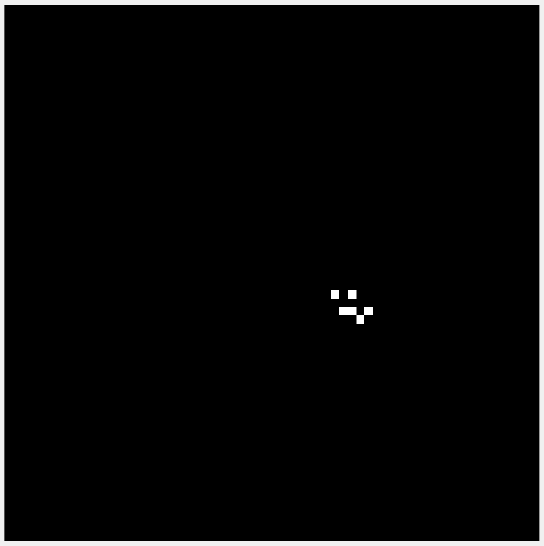}
    \includegraphics[width=0.1\textwidth]{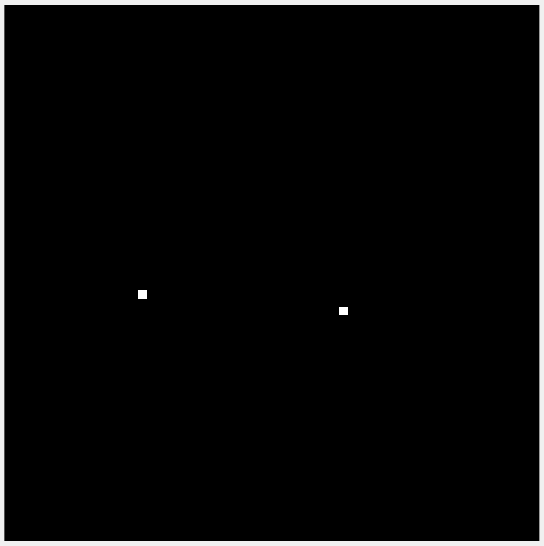}
    }    

    \subfloat[][Learning $\mathbf{U}$ and $\mathbf{V}$ with three-factor rule]{
    \includegraphics[width=0.1\textwidth]{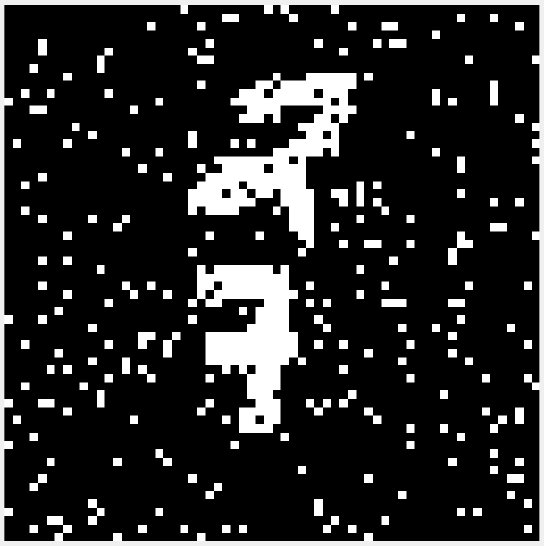}
    \includegraphics[width=0.1\textwidth]{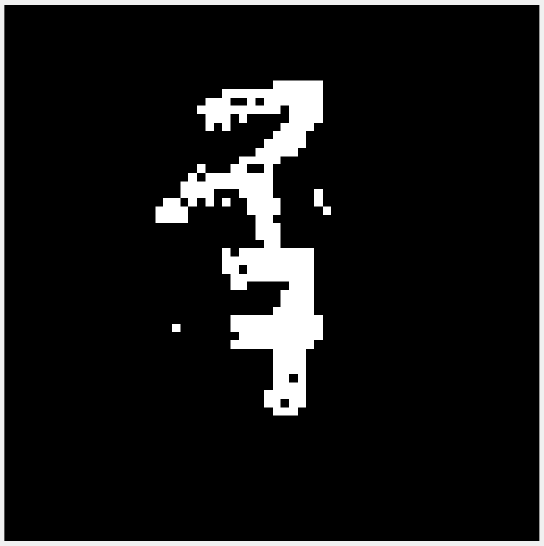}
    \includegraphics[width=0.1\textwidth]{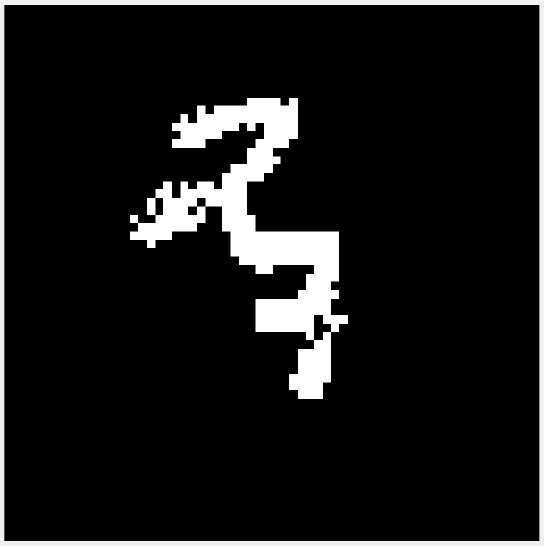}
    \includegraphics[width=0.1\textwidth]{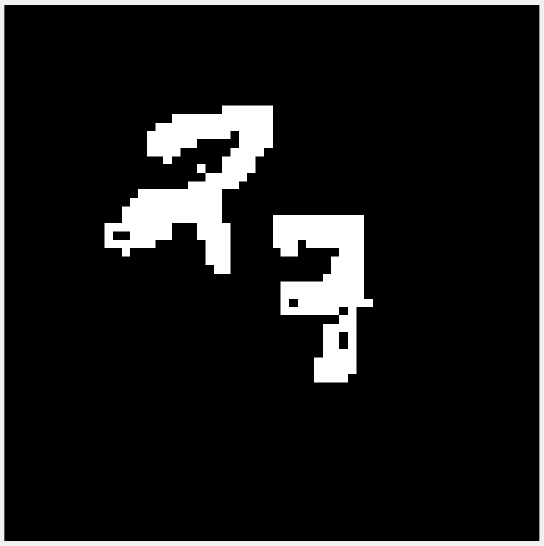}
    \includegraphics[width=0.1\textwidth]{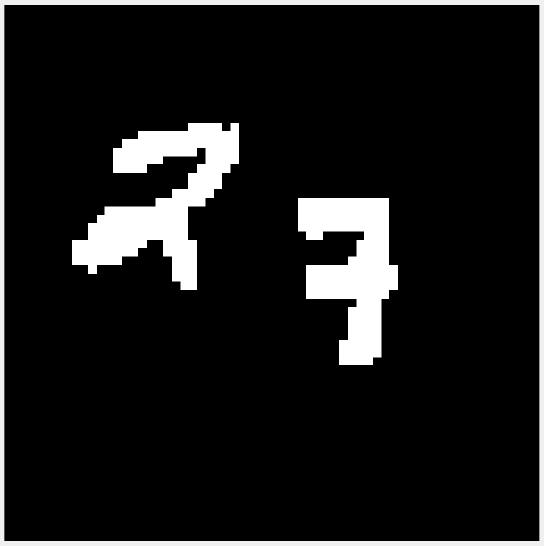}
    \includegraphics[width=0.1\textwidth]{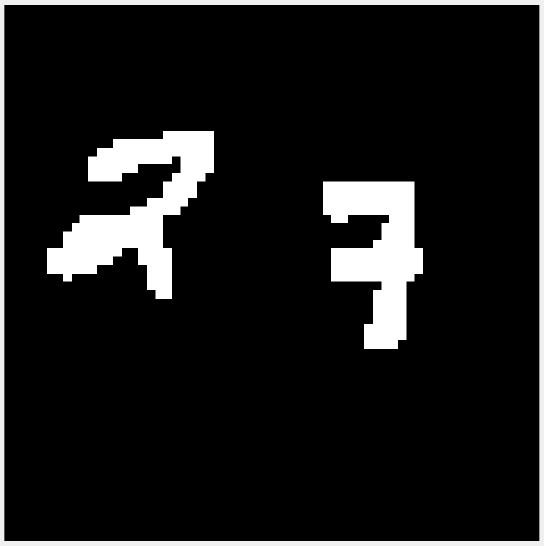}
    \includegraphics[width=0.1\textwidth]{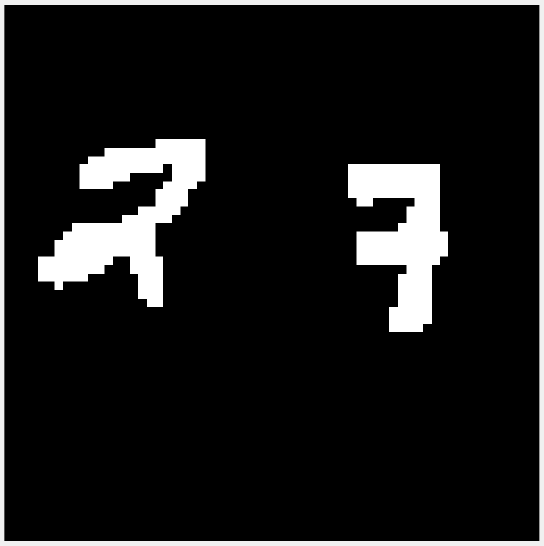}
    \includegraphics[width=0.1\textwidth]{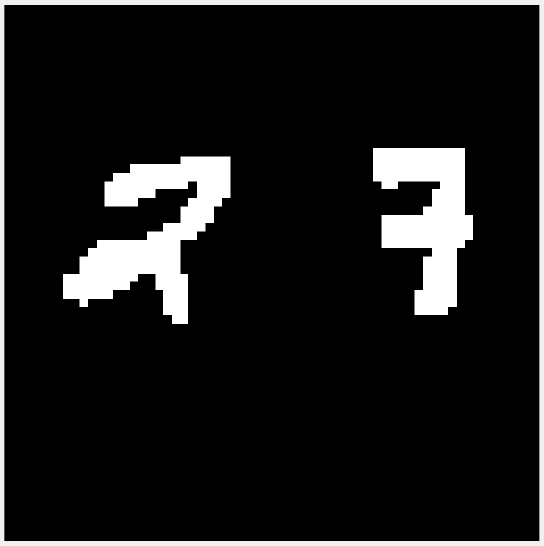}
    }    
    \caption{Moving MNIST sequence 1 for $t = 1,...,8$.}

\end{figure}

\begin{figure}[b!]
    \centering
    \subfloat[][Ground truth]{
    \includegraphics[width=0.1\textwidth]{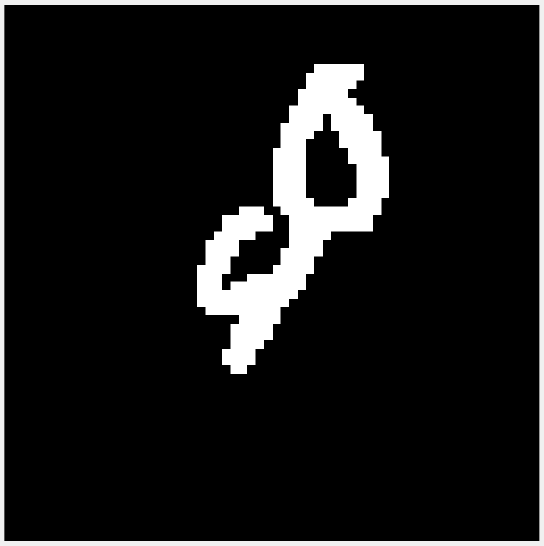}
    \includegraphics[width=0.1\textwidth]{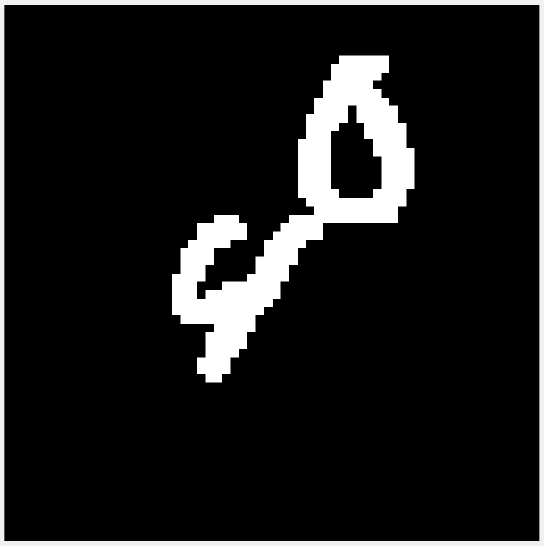}
    \includegraphics[width=0.1\textwidth]{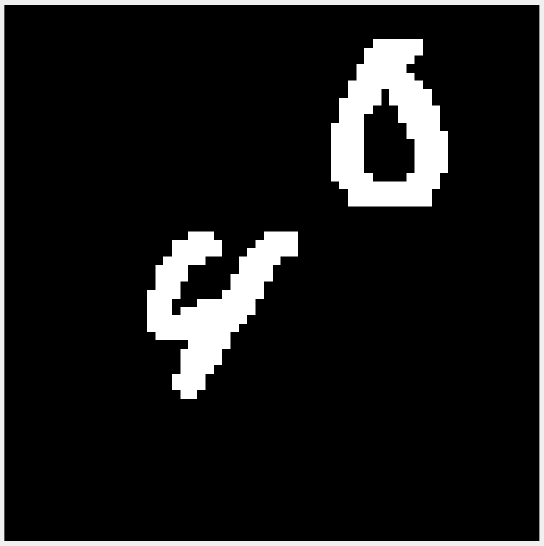}
    \includegraphics[width=0.1\textwidth]{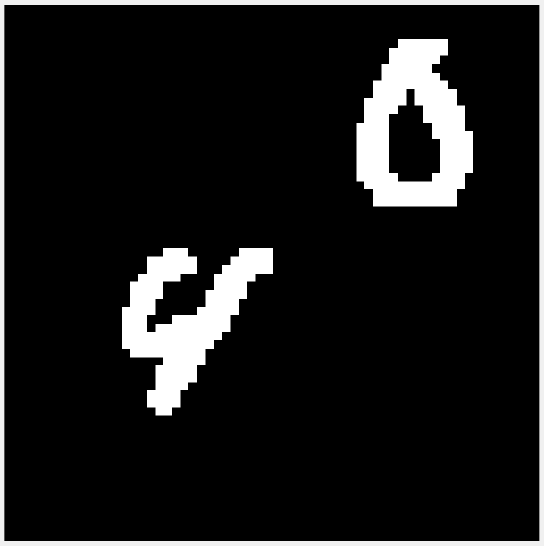}
    \includegraphics[width=0.1\textwidth]{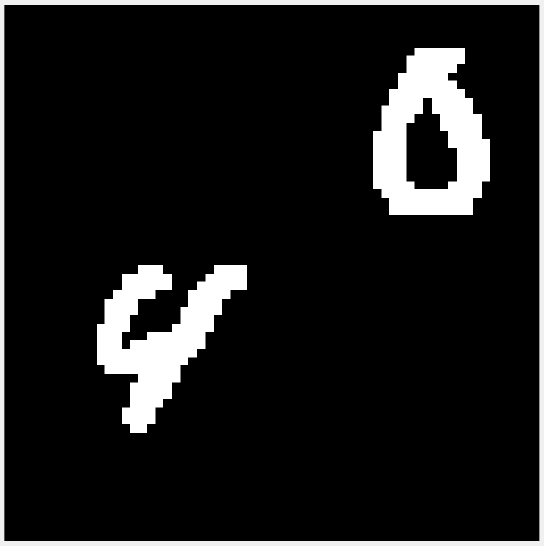}
    \includegraphics[width=0.1\textwidth]{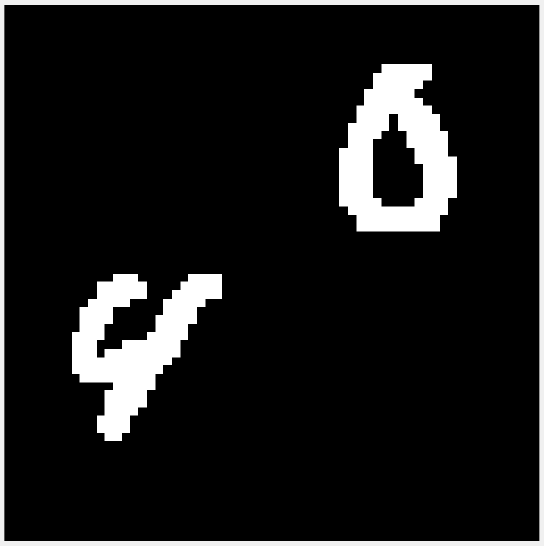}
    \includegraphics[width=0.1\textwidth]{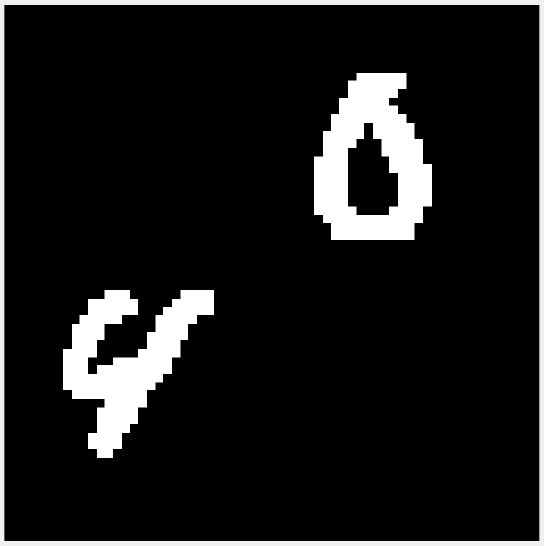}
    \includegraphics[width=0.1\textwidth]{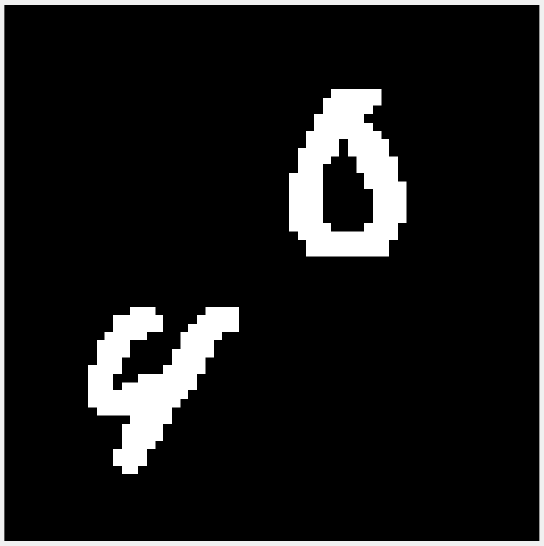}
    }

    \subfloat[][Fixing $\mathbf{U}$ and learning only $\mathbf{V}$ with temporal asymmetric Hebbian algorithm]{
    \includegraphics[width=0.1\textwidth]{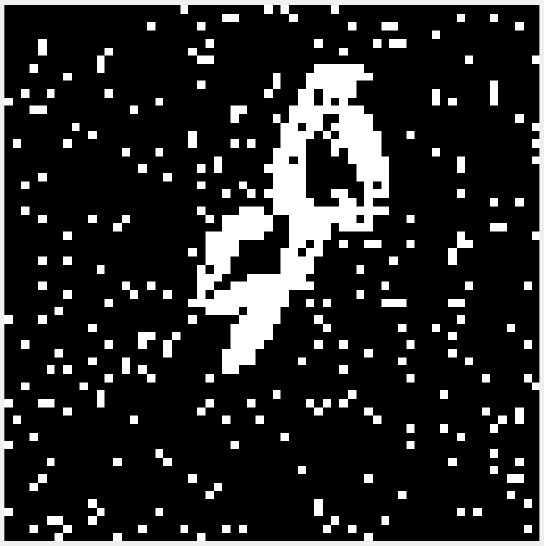}
    \includegraphics[width=0.1\textwidth]{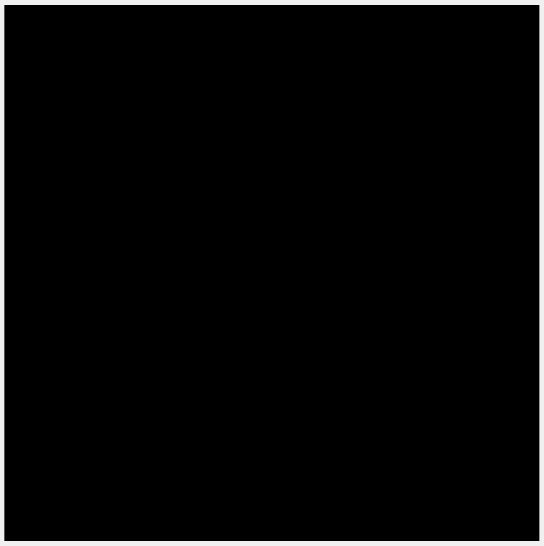}
    \includegraphics[width=0.1\textwidth]{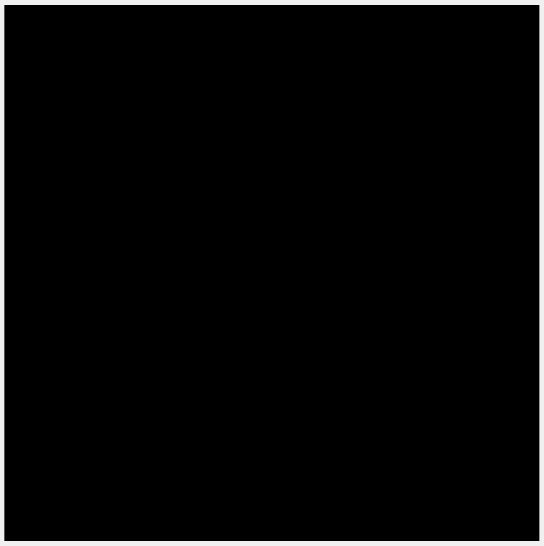}
    \includegraphics[width=0.1\textwidth]{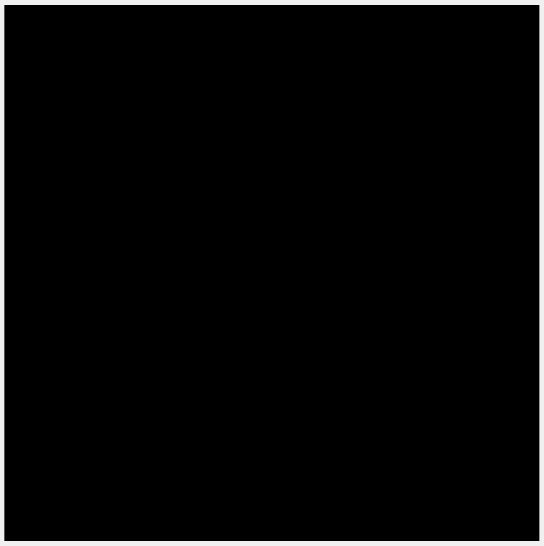}
    \includegraphics[width=0.1\textwidth]{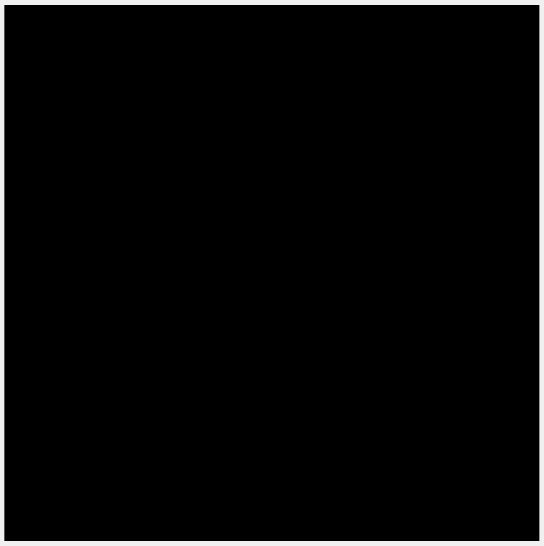}
    \includegraphics[width=0.1\textwidth]{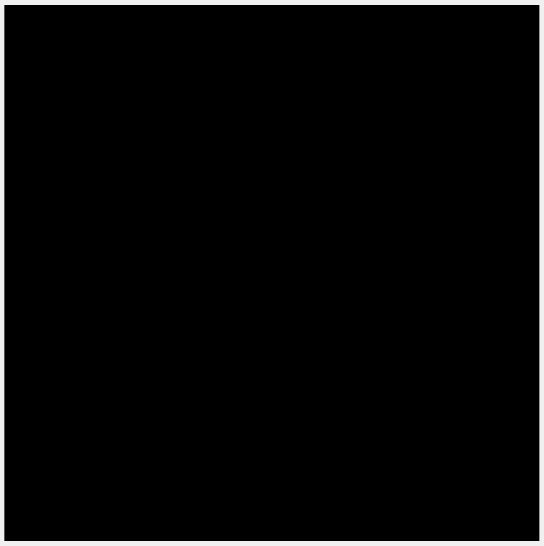}
    \includegraphics[width=0.1\textwidth]{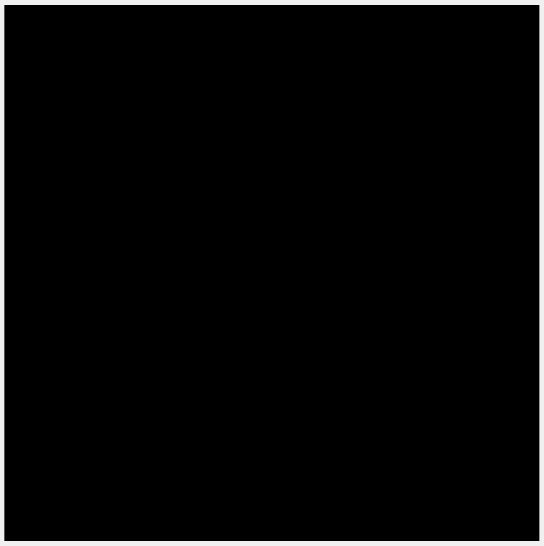}
    \includegraphics[width=0.1\textwidth]{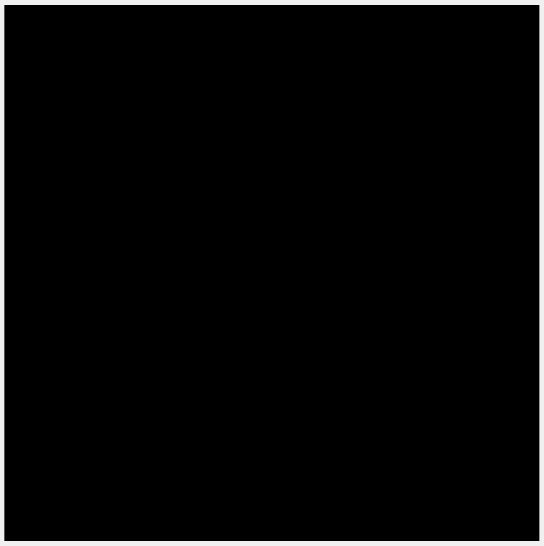}
    }

    \subfloat[][Fixing $\mathbf{U}$ and learning only $\mathbf{V}$ with three-factor rule]{
    \includegraphics[width=0.1\textwidth]{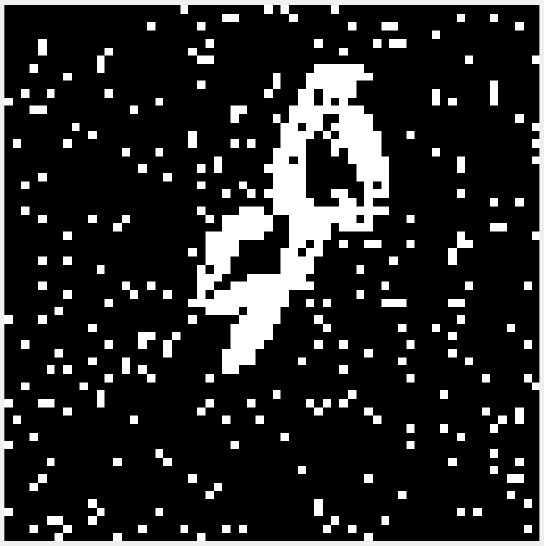}
    \includegraphics[width=0.1\textwidth]{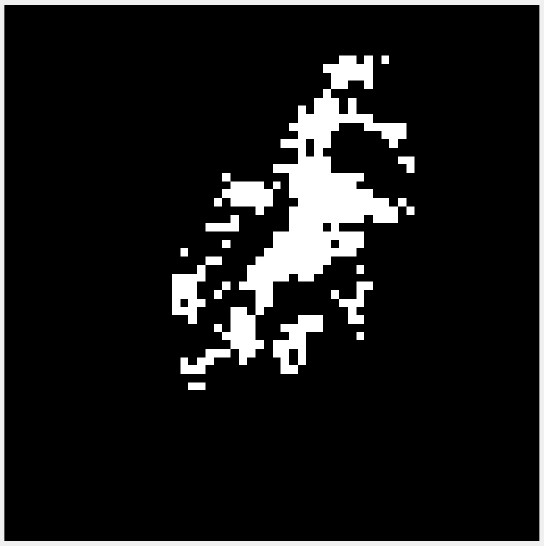}
    \includegraphics[width=0.1\textwidth]{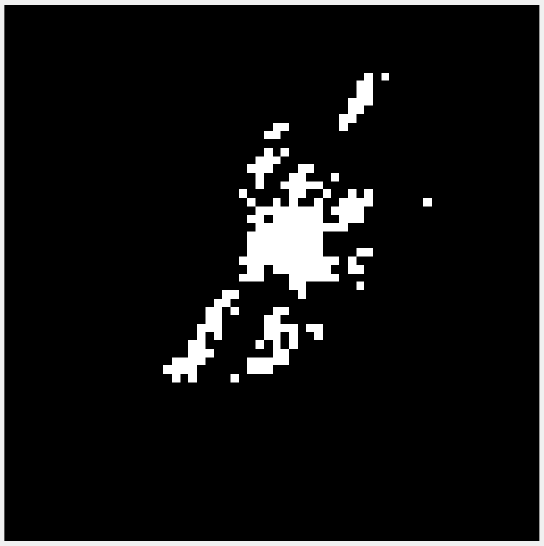}
    \includegraphics[width=0.1\textwidth]{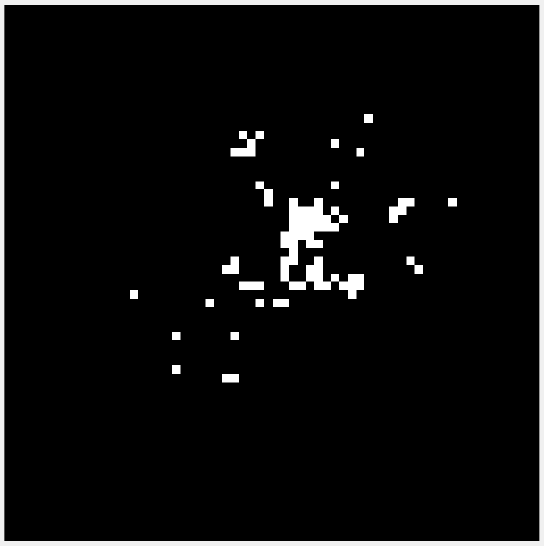}
    \includegraphics[width=0.1\textwidth]{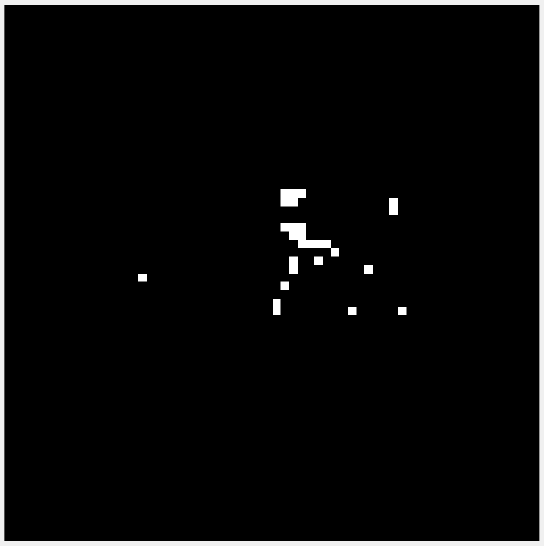}
    \includegraphics[width=0.1\textwidth]{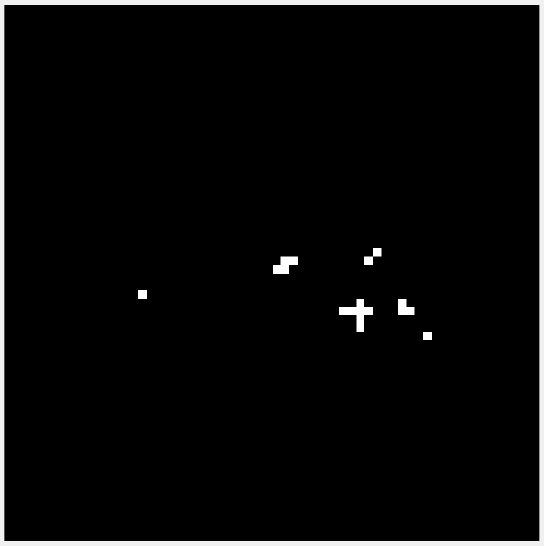}
    \includegraphics[width=0.1\textwidth]{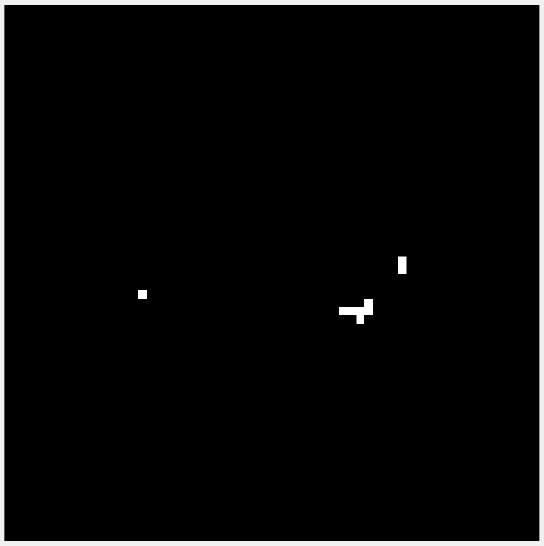}
    \includegraphics[width=0.1\textwidth]{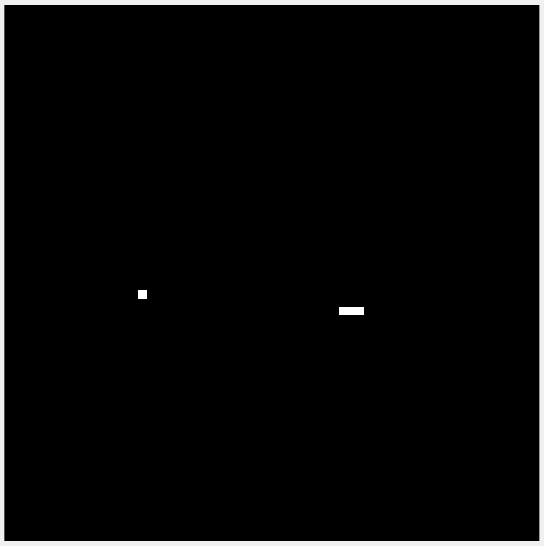}
    }    

    \subfloat[][Learning $\mathbf{U}$ and $\mathbf{V}$ with three-factor rule]{
    \includegraphics[width=0.1\textwidth]{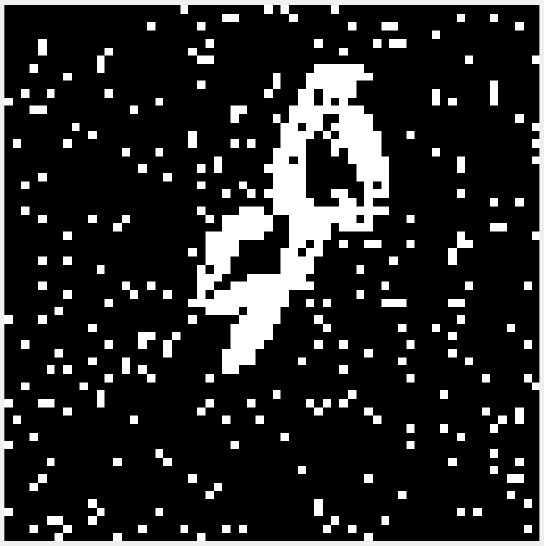}
    \includegraphics[width=0.1\textwidth]{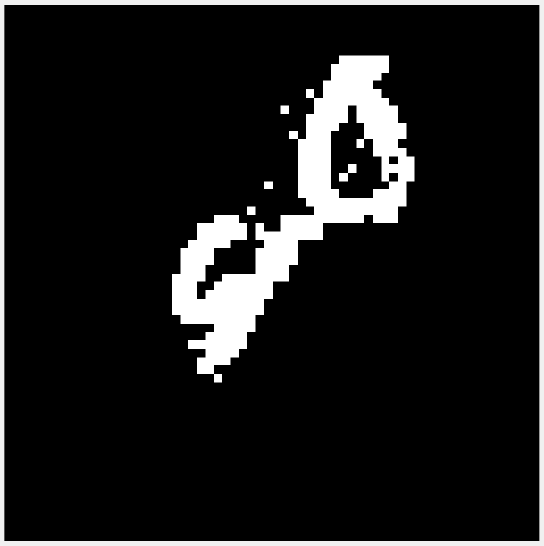}
    \includegraphics[width=0.1\textwidth]{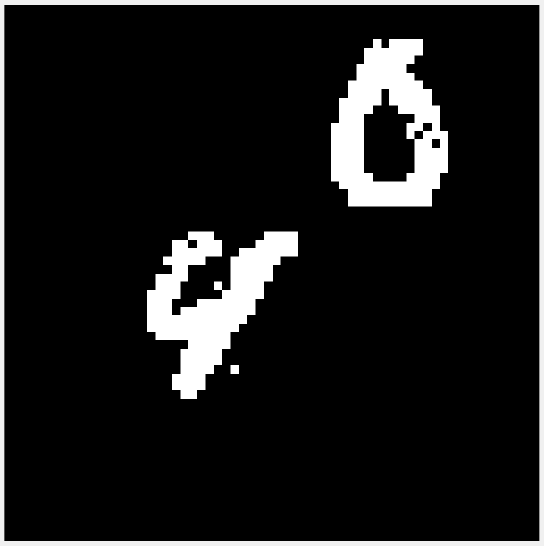}
    \includegraphics[width=0.1\textwidth]{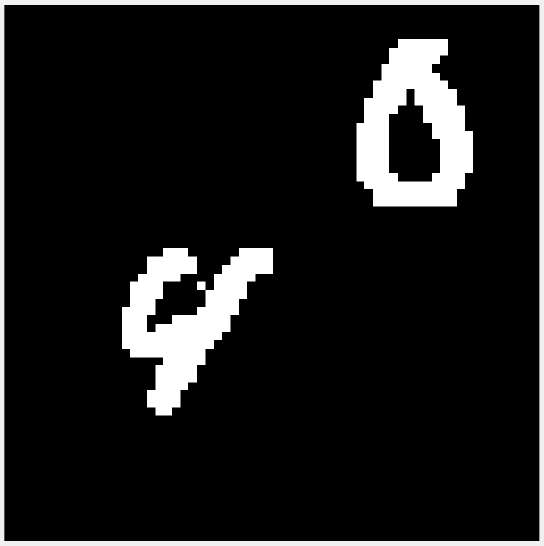}
    \includegraphics[width=0.1\textwidth]{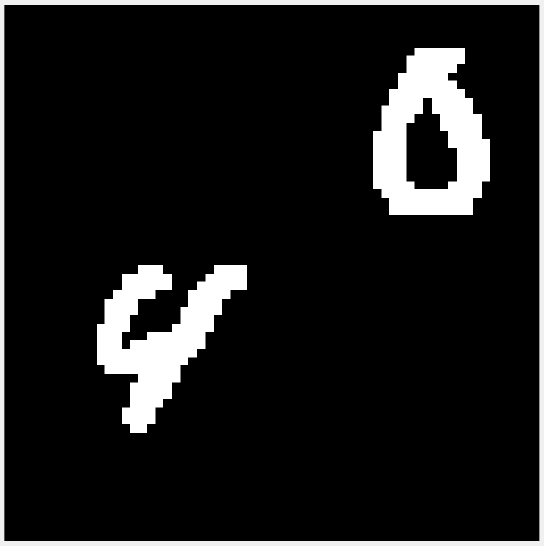}
    \includegraphics[width=0.1\textwidth]{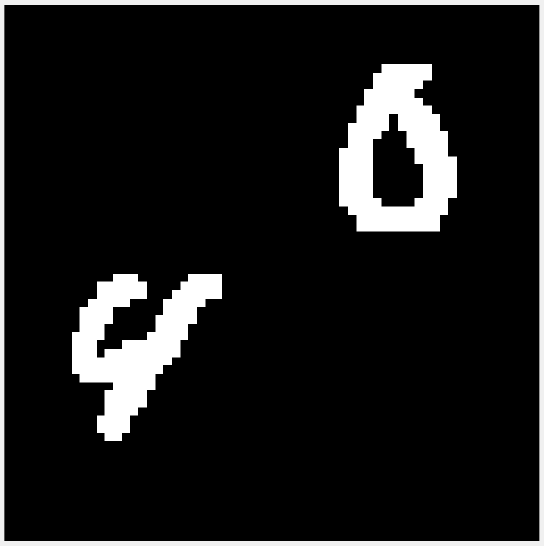}
    \includegraphics[width=0.1\textwidth]{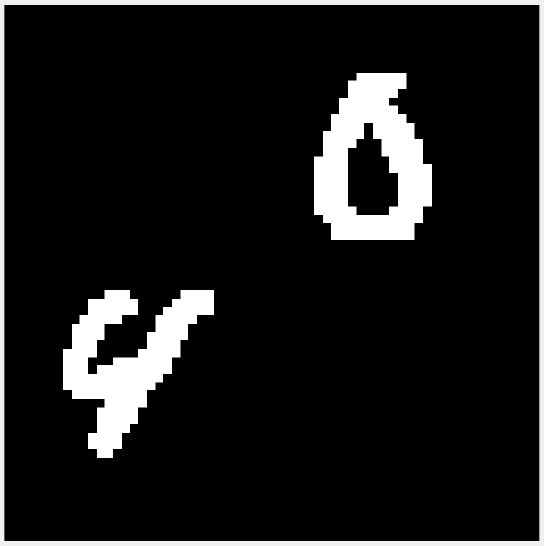}
    \includegraphics[width=0.1\textwidth]{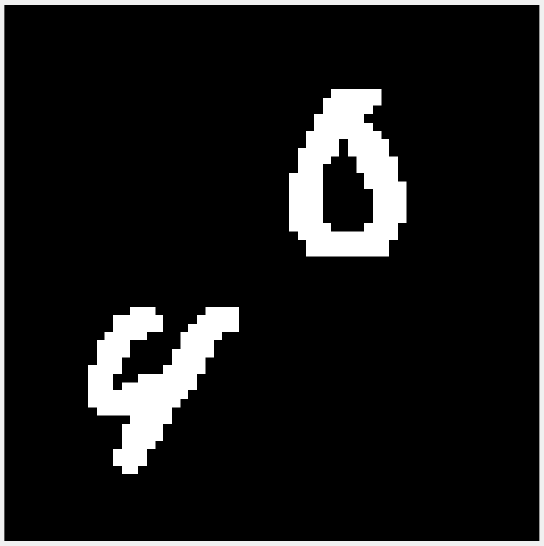}
    }    
    \caption{Moving MNIST sequence 6 for $t = 1,...,8$.}

\end{figure}

\begin{figure}[h!]
    \centering
    \subfloat[][Ground truth]{
    \includegraphics[width=0.1\textwidth]{img/seq_11_1_gt.pdf}
    \includegraphics[width=0.1\textwidth]{img/seq_11_2_gt.pdf}
    \includegraphics[width=0.1\textwidth]{img/seq_11_3_gt.pdf}
    \includegraphics[width=0.1\textwidth]{img/seq_11_4_gt.pdf}
    \includegraphics[width=0.1\textwidth]{img/seq_11_5_gt.pdf}
    \includegraphics[width=0.1\textwidth]{img/seq_11_6_gt.pdf}
    \includegraphics[width=0.1\textwidth]{img/seq_11_7_gt.pdf}
    \includegraphics[width=0.1\textwidth]{img/seq_11_8_gt.pdf}
    }

    \subfloat[][Fixing $\mathbf{U}$ and learning only $\mathbf{V}$ with temporal asymmetric Hebbian algorithm]{
    \includegraphics[width=0.1\textwidth]{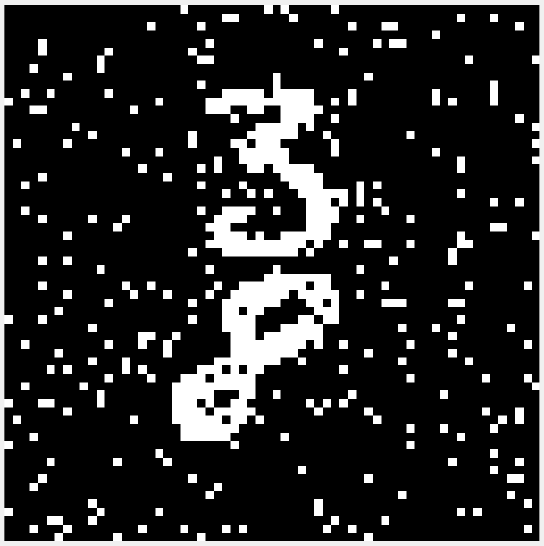}
    \includegraphics[width=0.1\textwidth]{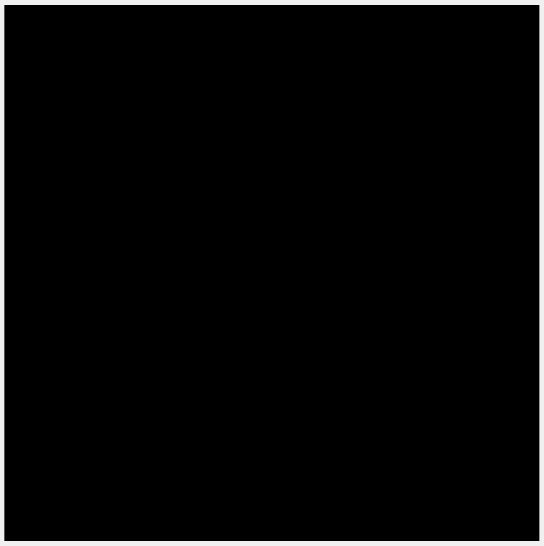}
    \includegraphics[width=0.1\textwidth]{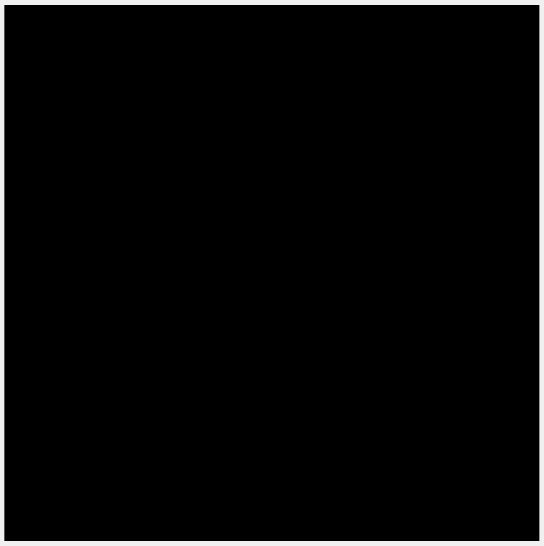}
    \includegraphics[width=0.1\textwidth]{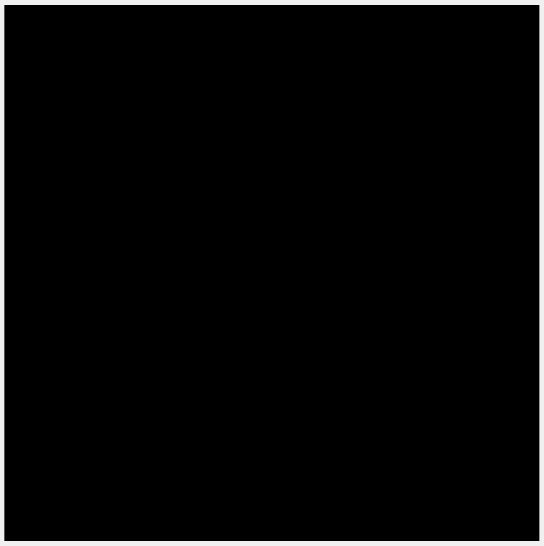}
    \includegraphics[width=0.1\textwidth]{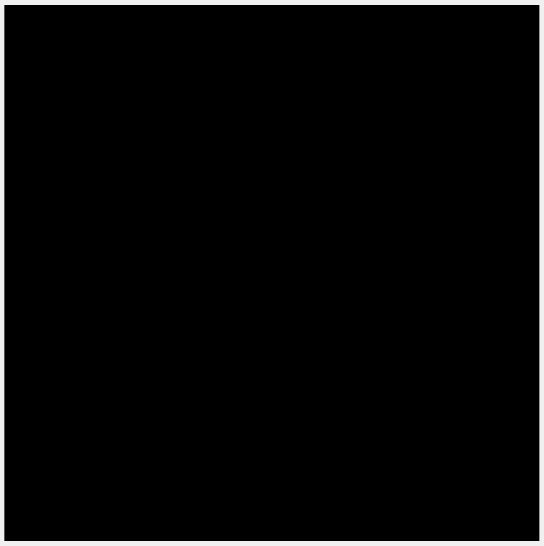}
    \includegraphics[width=0.1\textwidth]{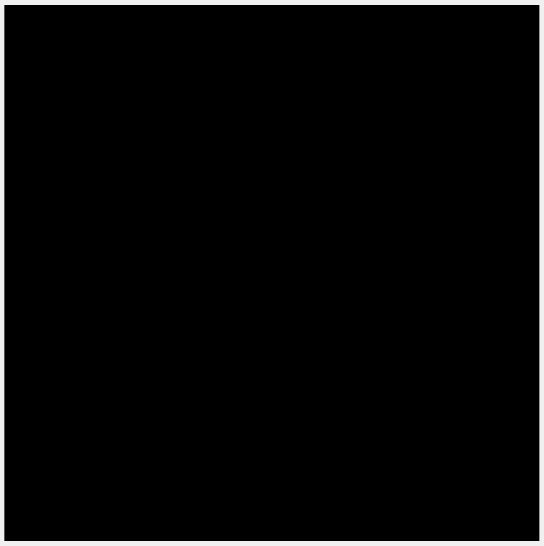}
    \includegraphics[width=0.1\textwidth]{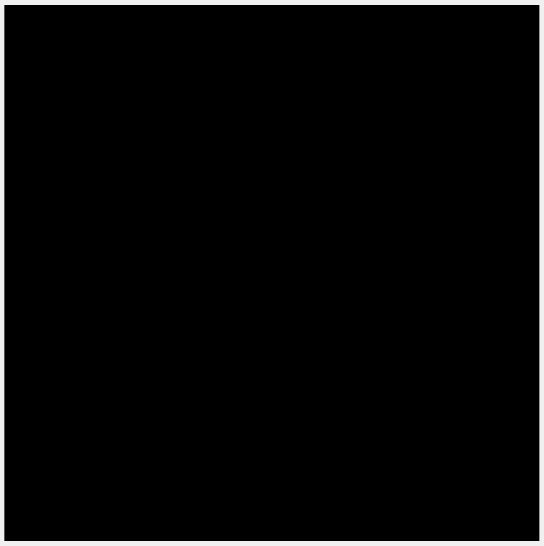}
    \includegraphics[width=0.1\textwidth]{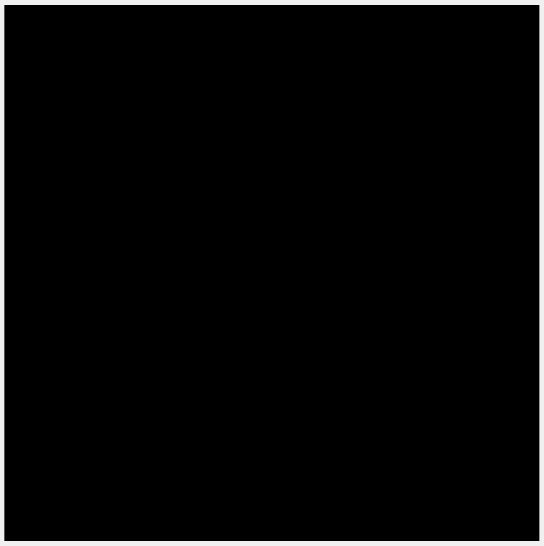}
    }

    \subfloat[][Fixing $\mathbf{U}$ and learning only $\mathbf{V}$ with three-factor rule]{
    \includegraphics[width=0.1\textwidth]{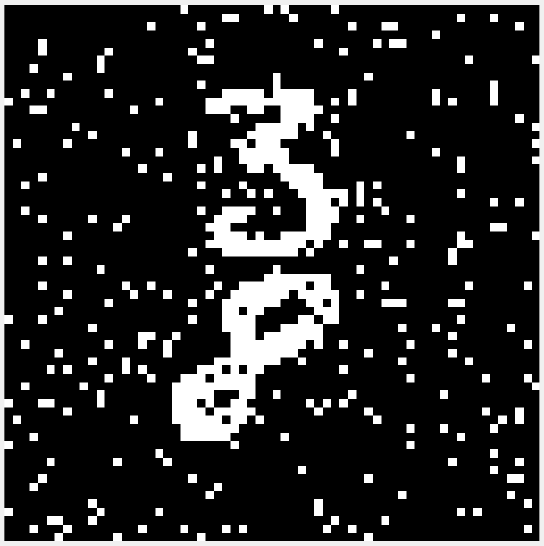}
    \includegraphics[width=0.1\textwidth]{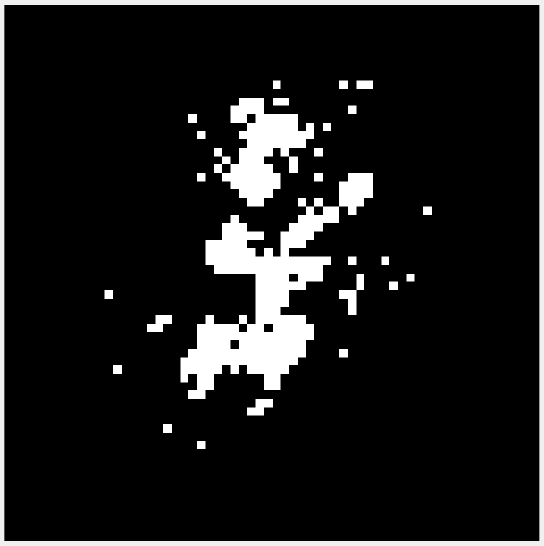}
    \includegraphics[width=0.1\textwidth]{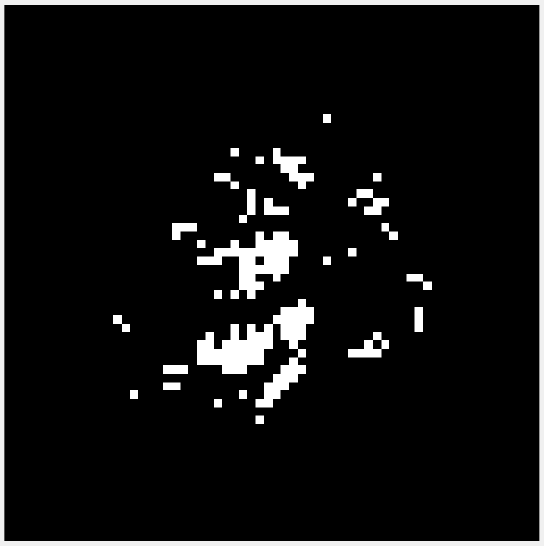}
    \includegraphics[width=0.1\textwidth]{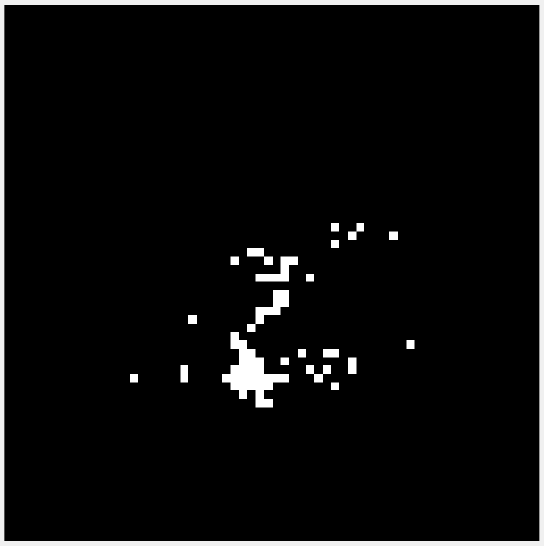}
    \includegraphics[width=0.1\textwidth]{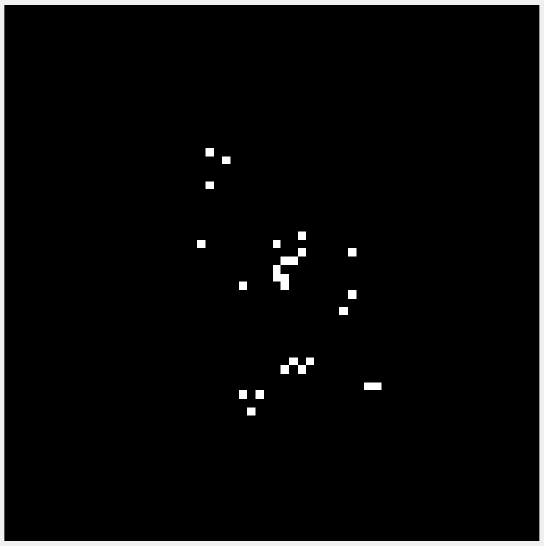}
    \includegraphics[width=0.1\textwidth]{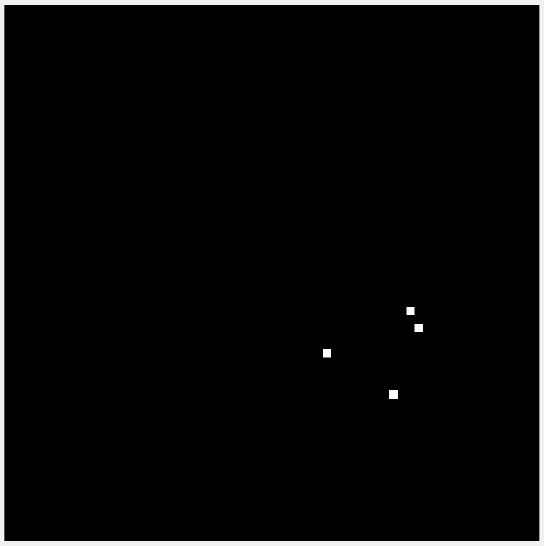}
    \includegraphics[width=0.1\textwidth]{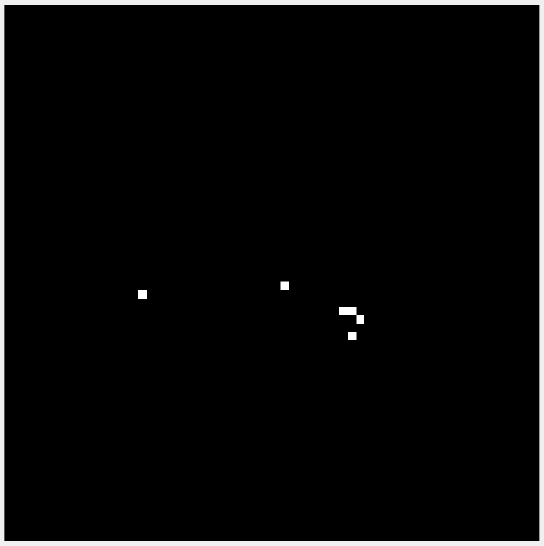}
    \includegraphics[width=0.1\textwidth]{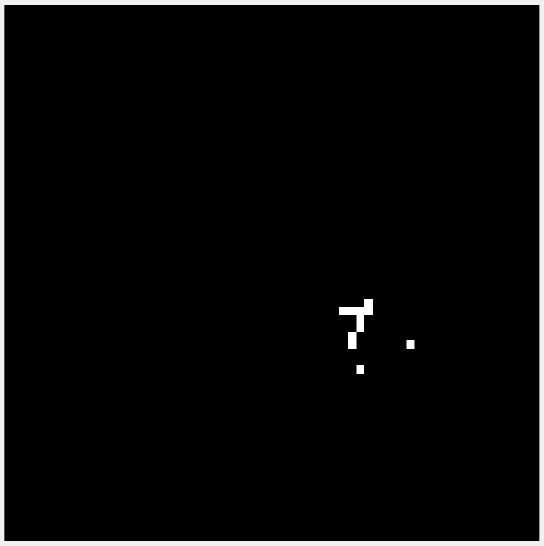}
    }  

    \subfloat[][Learning $\mathbf{U}$ and $\mathbf{V}$ with three-factor rule]{
    \includegraphics[width=0.1\textwidth]{img/seq_11_1_two.pdf}
    \includegraphics[width=0.1\textwidth]{img/seq_11_2_two.pdf}
    \includegraphics[width=0.1\textwidth]{img/seq_11_3_two.pdf}
    \includegraphics[width=0.1\textwidth]{img/seq_11_4_two.pdf}
    \includegraphics[width=0.1\textwidth]{img/seq_11_5_two.pdf}
    \includegraphics[width=0.1\textwidth]{img/seq_11_6_two.pdf}
    \includegraphics[width=0.1\textwidth]{img/seq_11_7_two.pdf}
    \includegraphics[width=0.1\textwidth]{img/seq_11_8_two.pdf}
    }
   
    \caption{Moving MNIST sequence 11 for $t = 1,...,8$.}

\end{figure}

\begin{figure}[h!]
    \centering
    \subfloat[][Ground truth]{
    \includegraphics[width=0.1\textwidth]{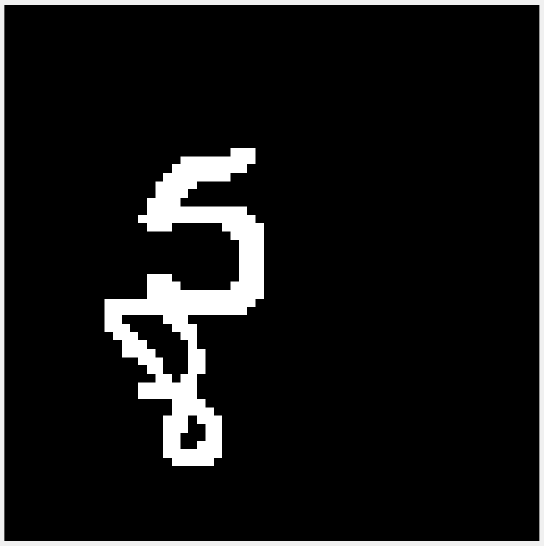}
    \includegraphics[width=0.1\textwidth]{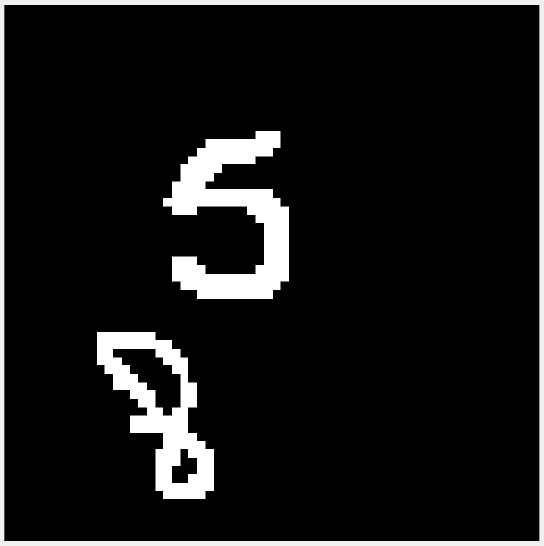}
    \includegraphics[width=0.1\textwidth]{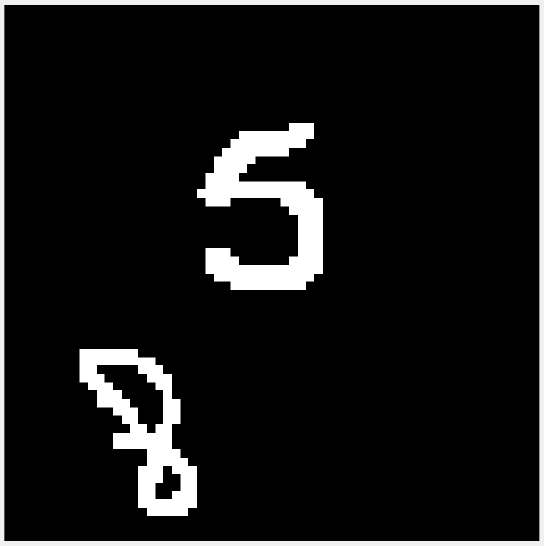}
    \includegraphics[width=0.1\textwidth]{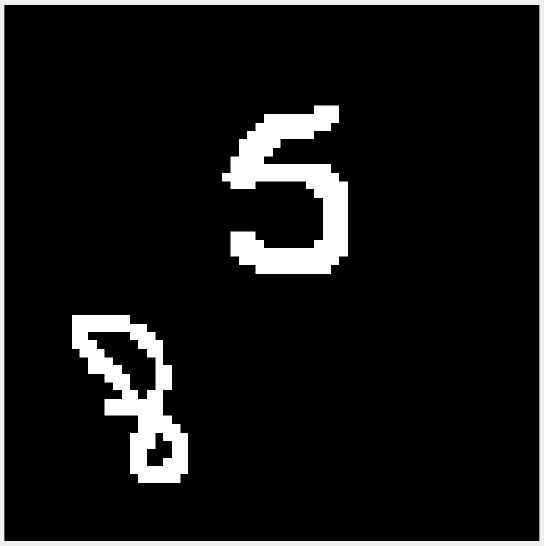}
    \includegraphics[width=0.1\textwidth]{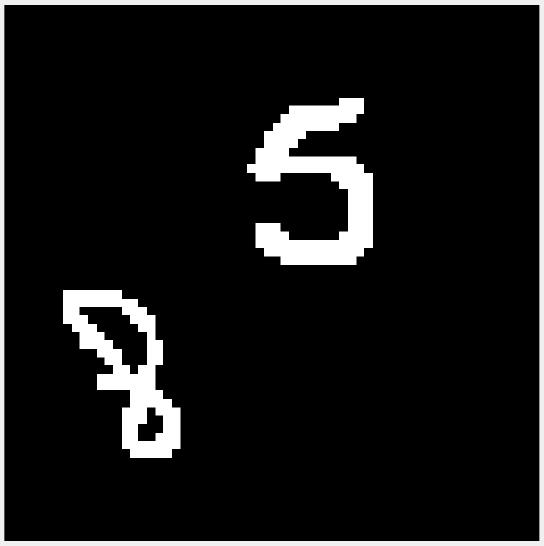}
    \includegraphics[width=0.1\textwidth]{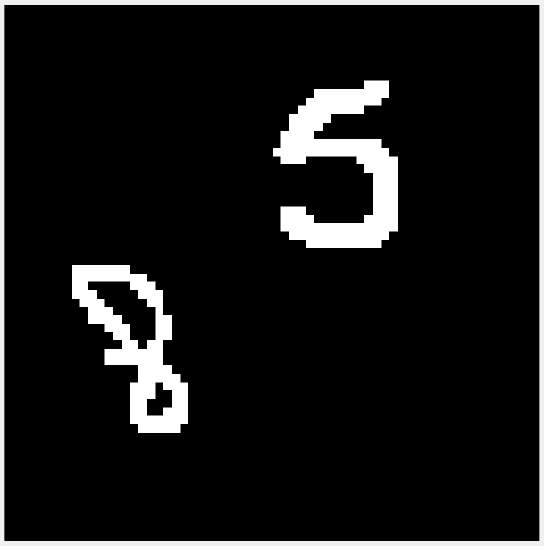}
    \includegraphics[width=0.1\textwidth]{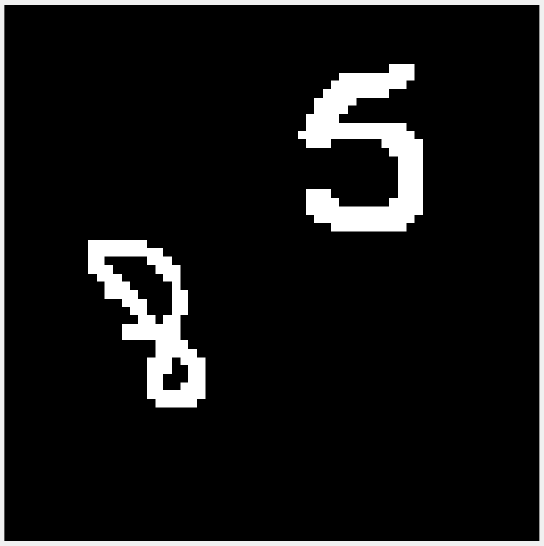}
    \includegraphics[width=0.1\textwidth]{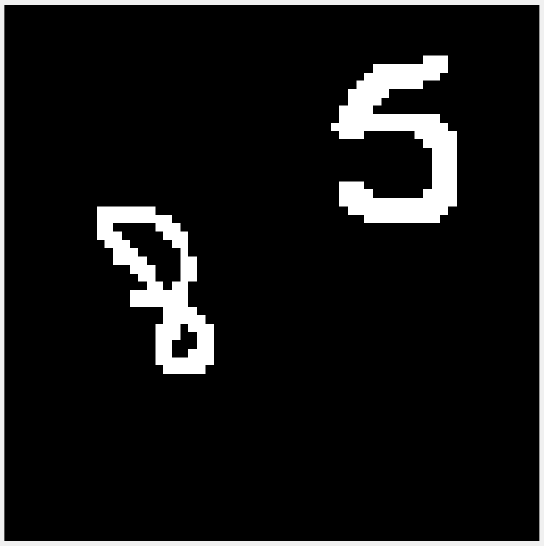}
    }

    \subfloat[][Fixing $\mathbf{U}$ and learning only $\mathbf{V}$ with temporal asymmetric Hebbian algorithm]{
    \includegraphics[width=0.1\textwidth]{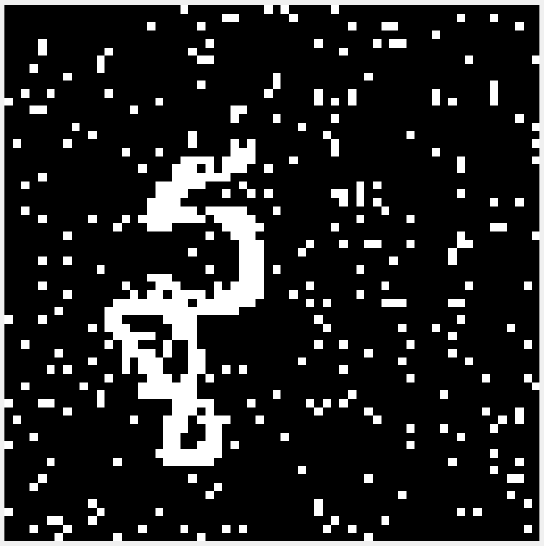}
    \includegraphics[width=0.1\textwidth]{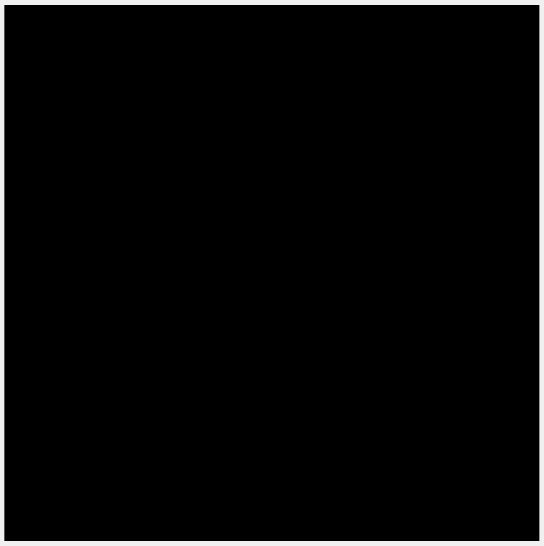}
    \includegraphics[width=0.1\textwidth]{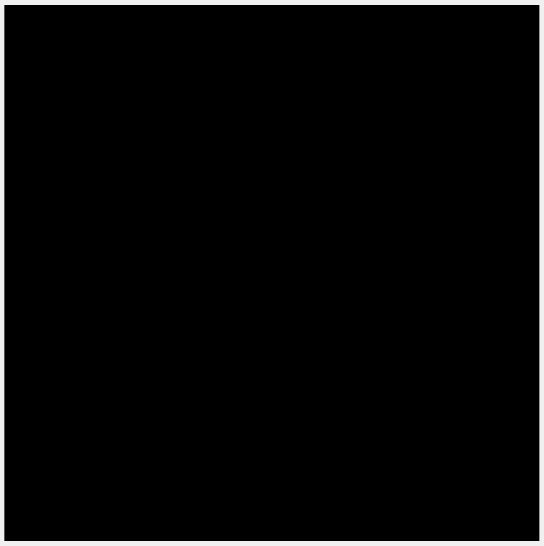}
    \includegraphics[width=0.1\textwidth]{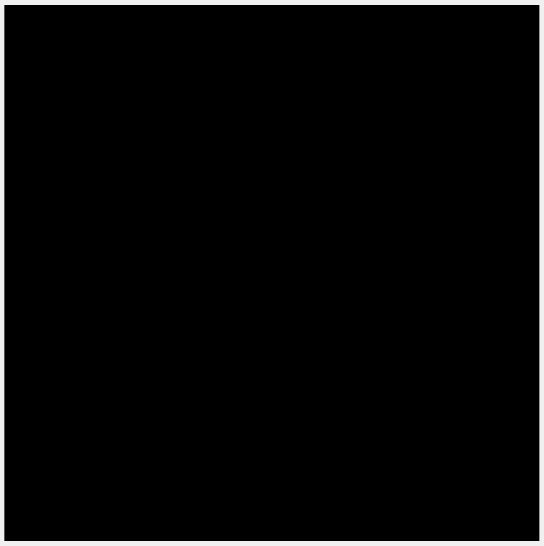}
    \includegraphics[width=0.1\textwidth]{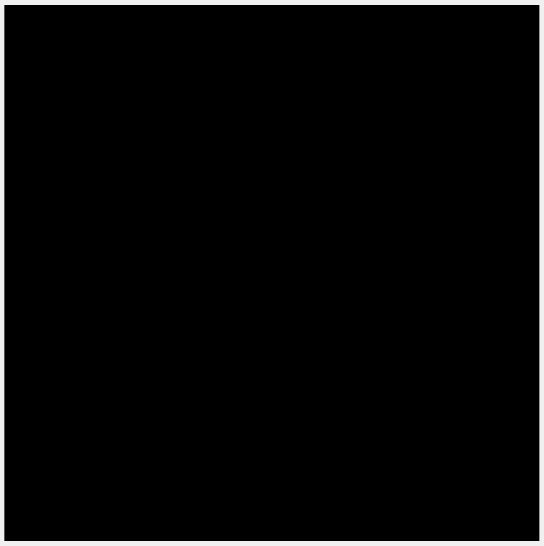}
    \includegraphics[width=0.1\textwidth]{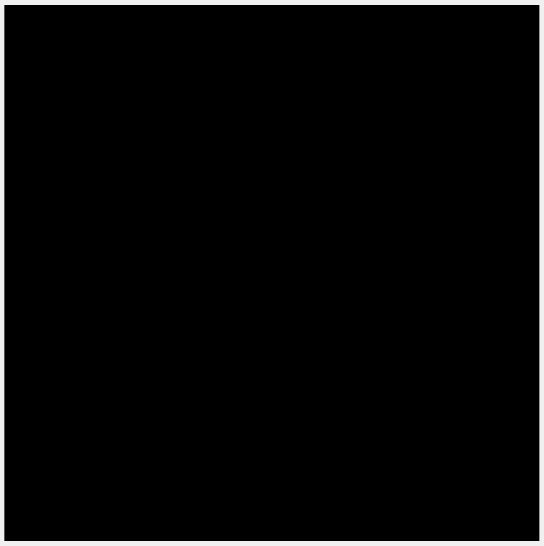}
    \includegraphics[width=0.1\textwidth]{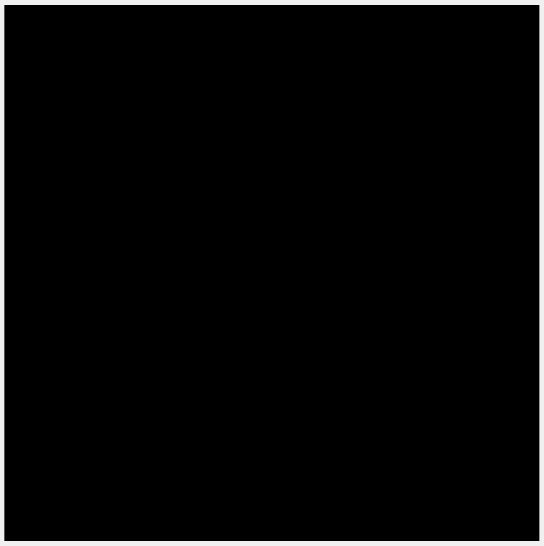}
    \includegraphics[width=0.1\textwidth]{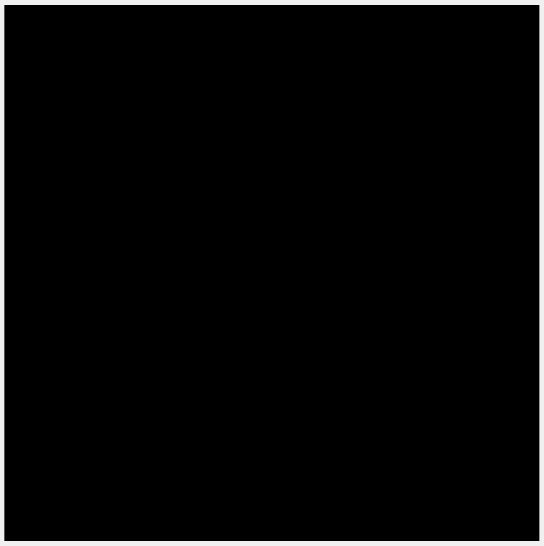}
    }

    \subfloat[][Fixing $\mathbf{U}$ and learning only $\mathbf{V}$ with three-factor rule]{
    \includegraphics[width=0.1\textwidth]{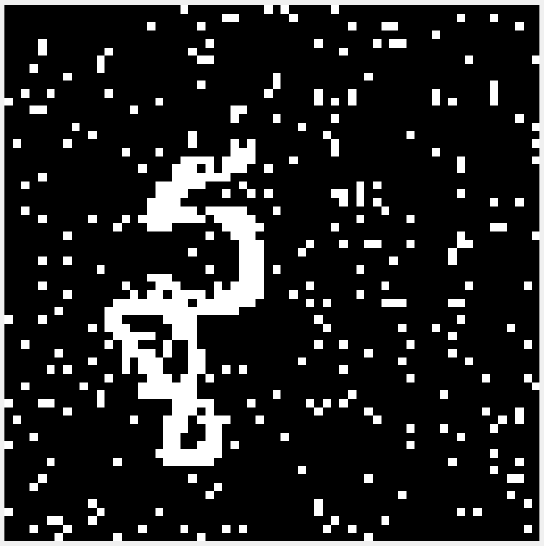}
    \includegraphics[width=0.1\textwidth]{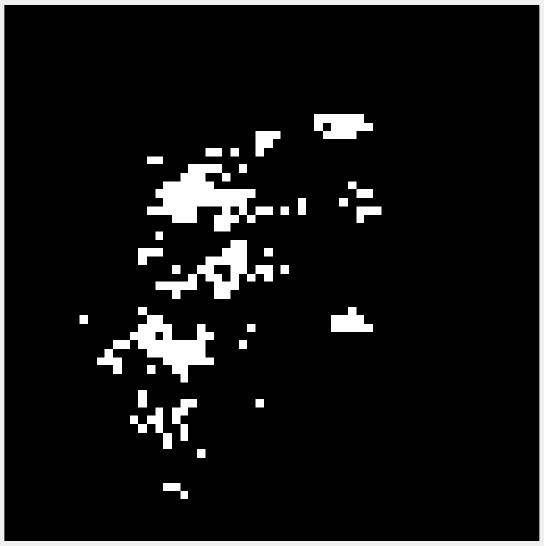}
    \includegraphics[width=0.1\textwidth]{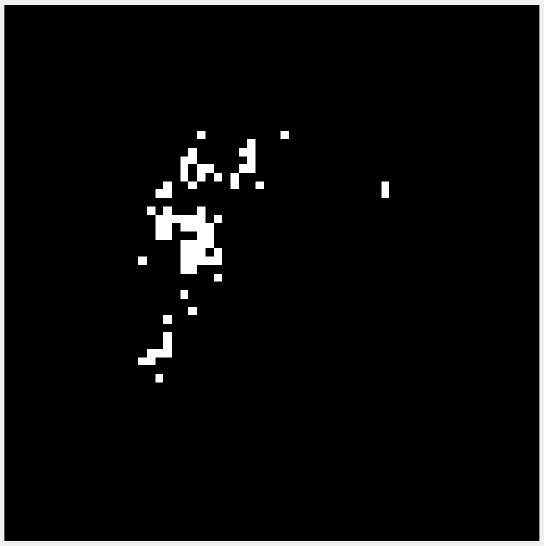}
    \includegraphics[width=0.1\textwidth]{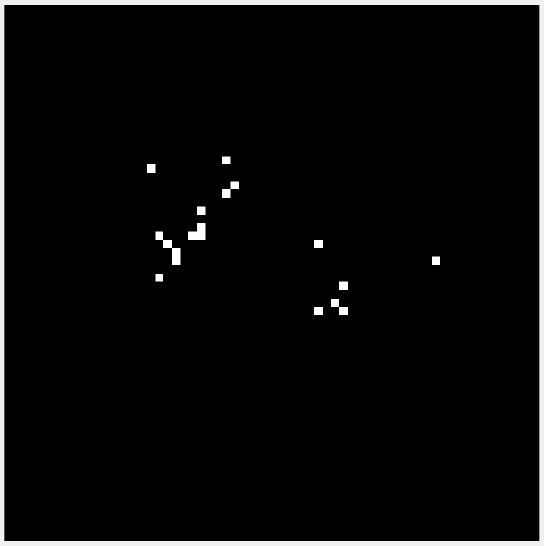}
    \includegraphics[width=0.1\textwidth]{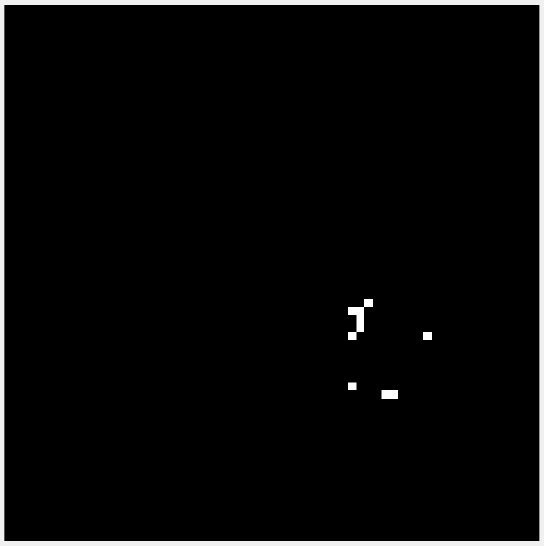}
    \includegraphics[width=0.1\textwidth]{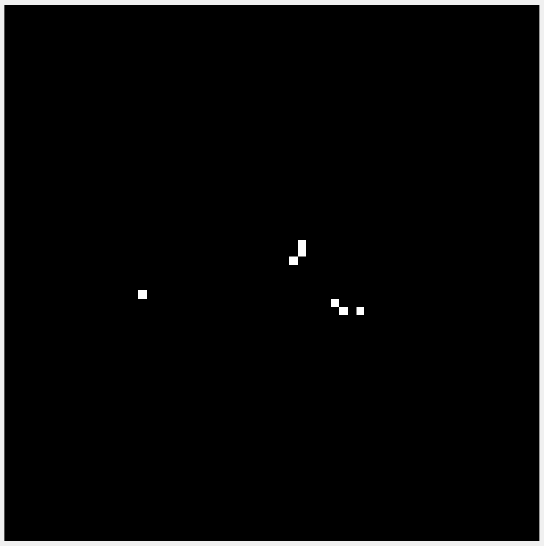}
    \includegraphics[width=0.1\textwidth]{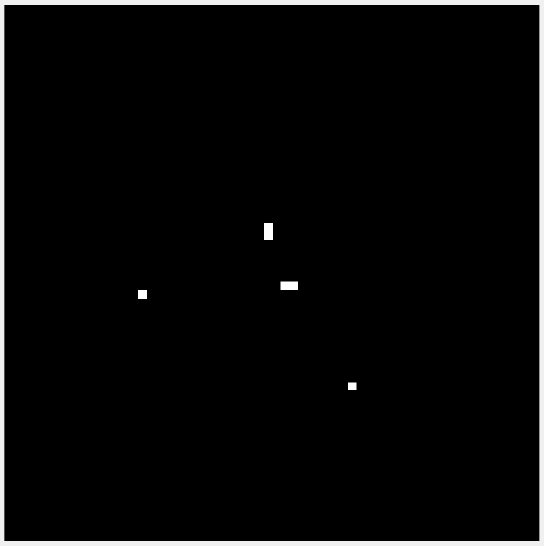}
    \includegraphics[width=0.1\textwidth]{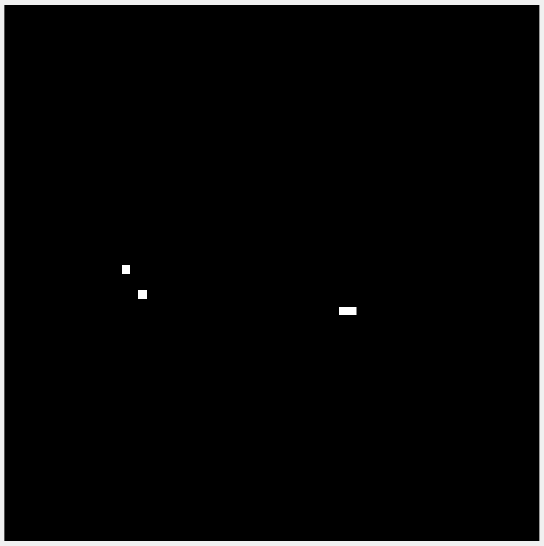}
    }    

    \subfloat[][Learning $\mathbf{U}$ and $\mathbf{V}$ with three-factor rule]{
    \includegraphics[width=0.1\textwidth]{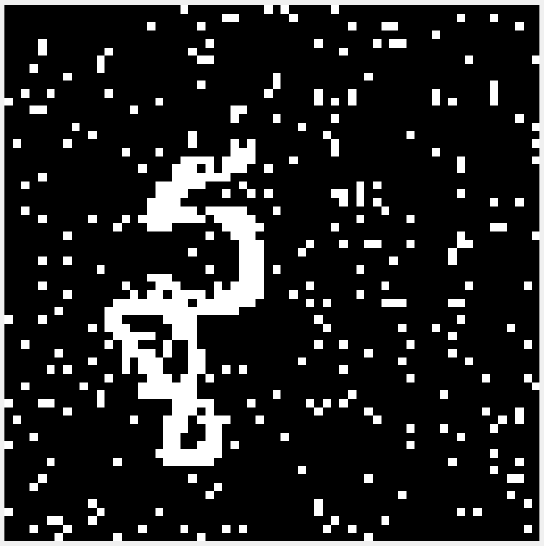}
    \includegraphics[width=0.1\textwidth]{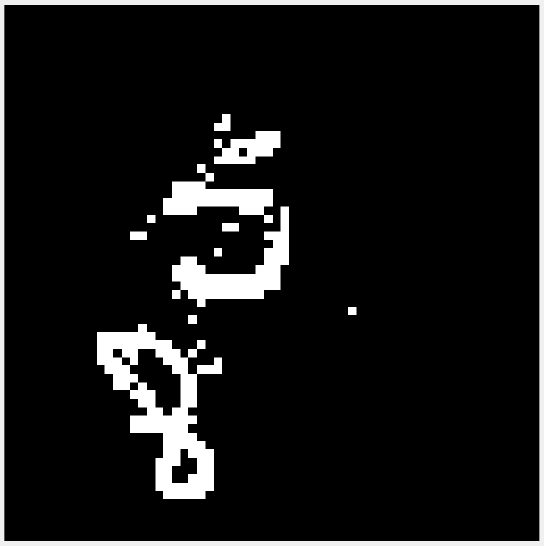}
    \includegraphics[width=0.1\textwidth]{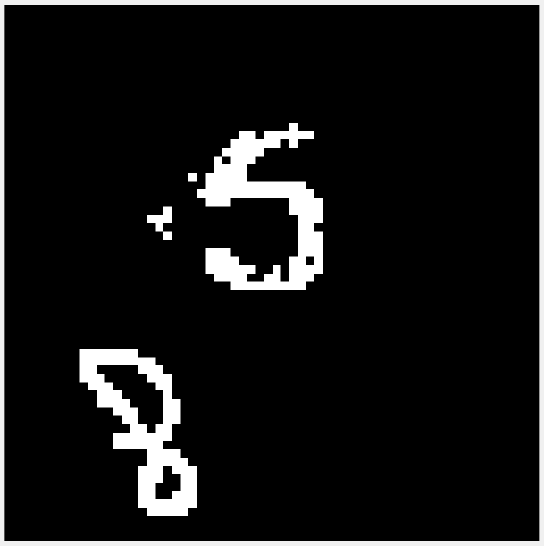}
    \includegraphics[width=0.1\textwidth]{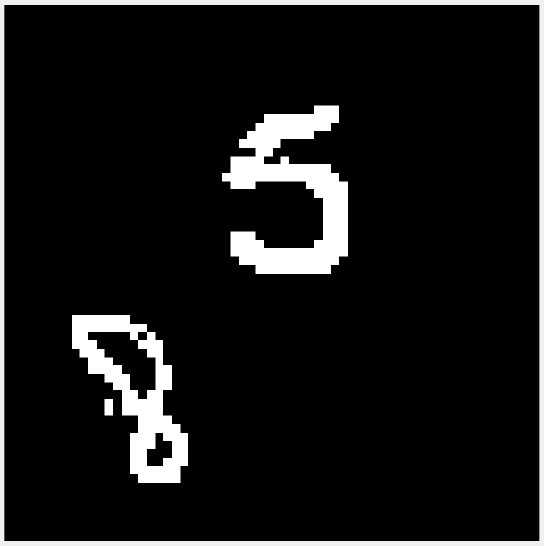}
    \includegraphics[width=0.1\textwidth]{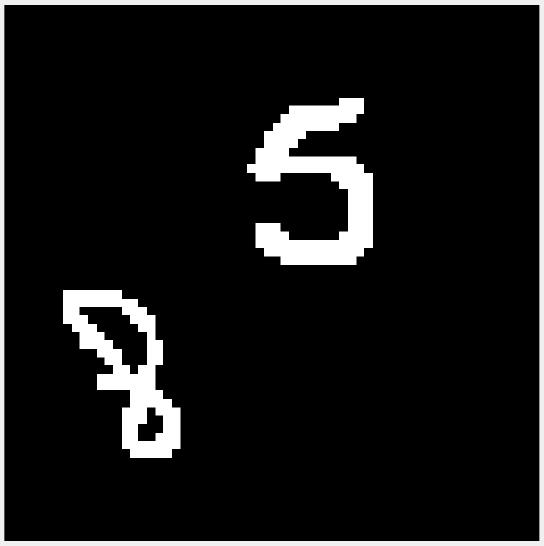}
    \includegraphics[width=0.1\textwidth]{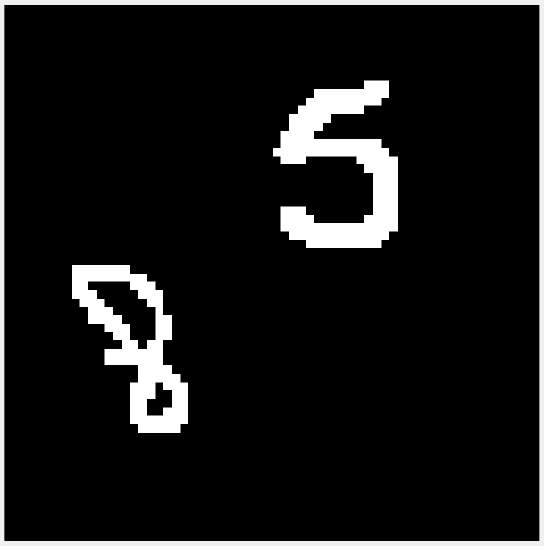}
    \includegraphics[width=0.1\textwidth]{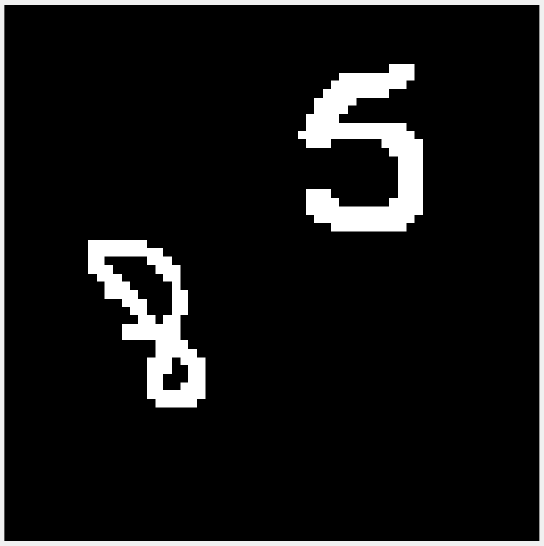}
    \includegraphics[width=0.1\textwidth]{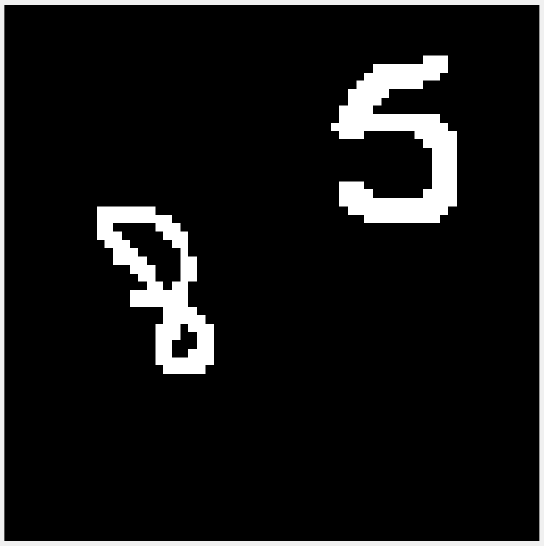}
    }    
    \caption{Moving MNIST sequence 16 for $t = 1,...,8$.}

\end{figure}

\begin{figure}[h!]
\centering
\includegraphics[width=0.9\textwidth]{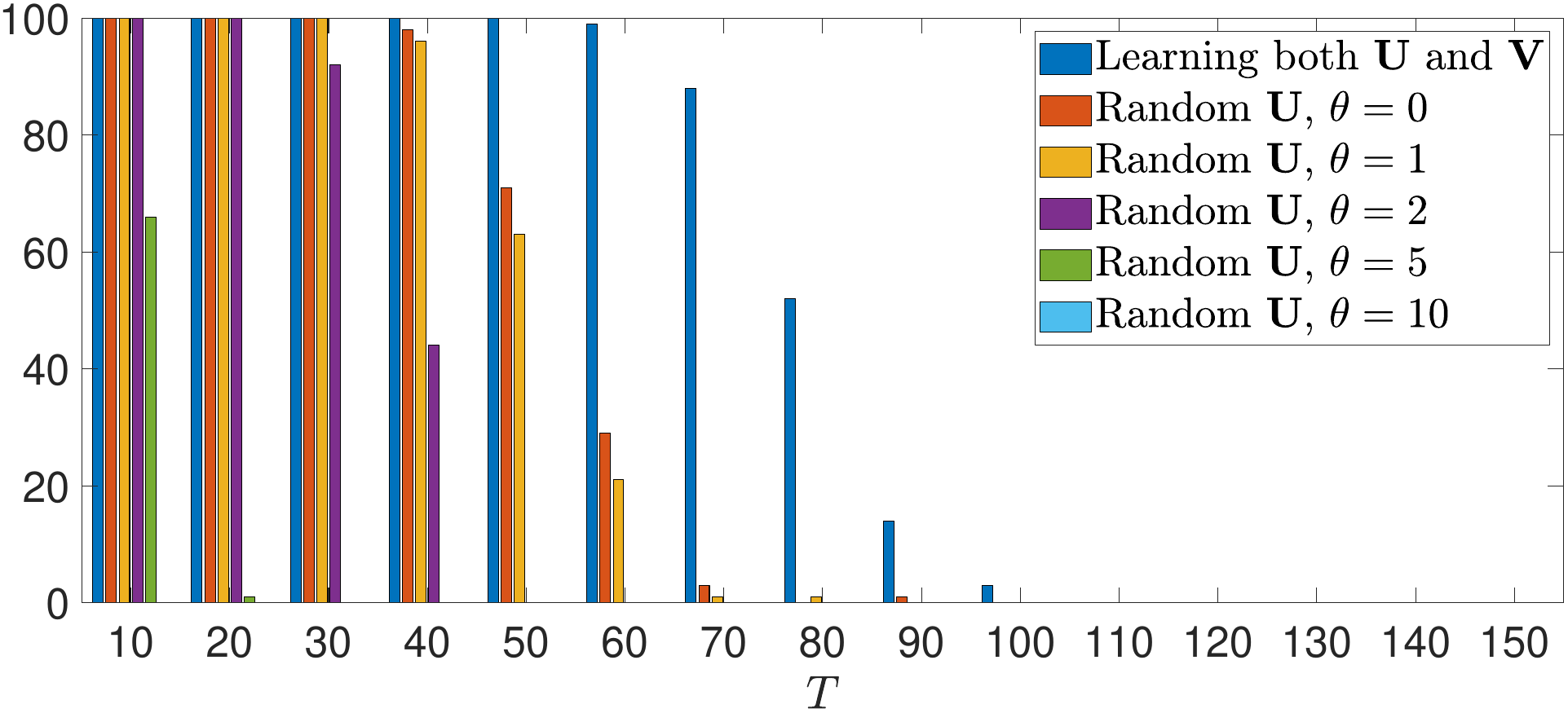}
\caption{Successful retrievals out of 100 trials with different sequence period lengths (sparse random projected inputs).}
\label{fig:sparse_U}
\end{figure}

\vspace{0.5cm}

\begin{figure}[h!]
\centering
\includegraphics[width=0.9\textwidth]{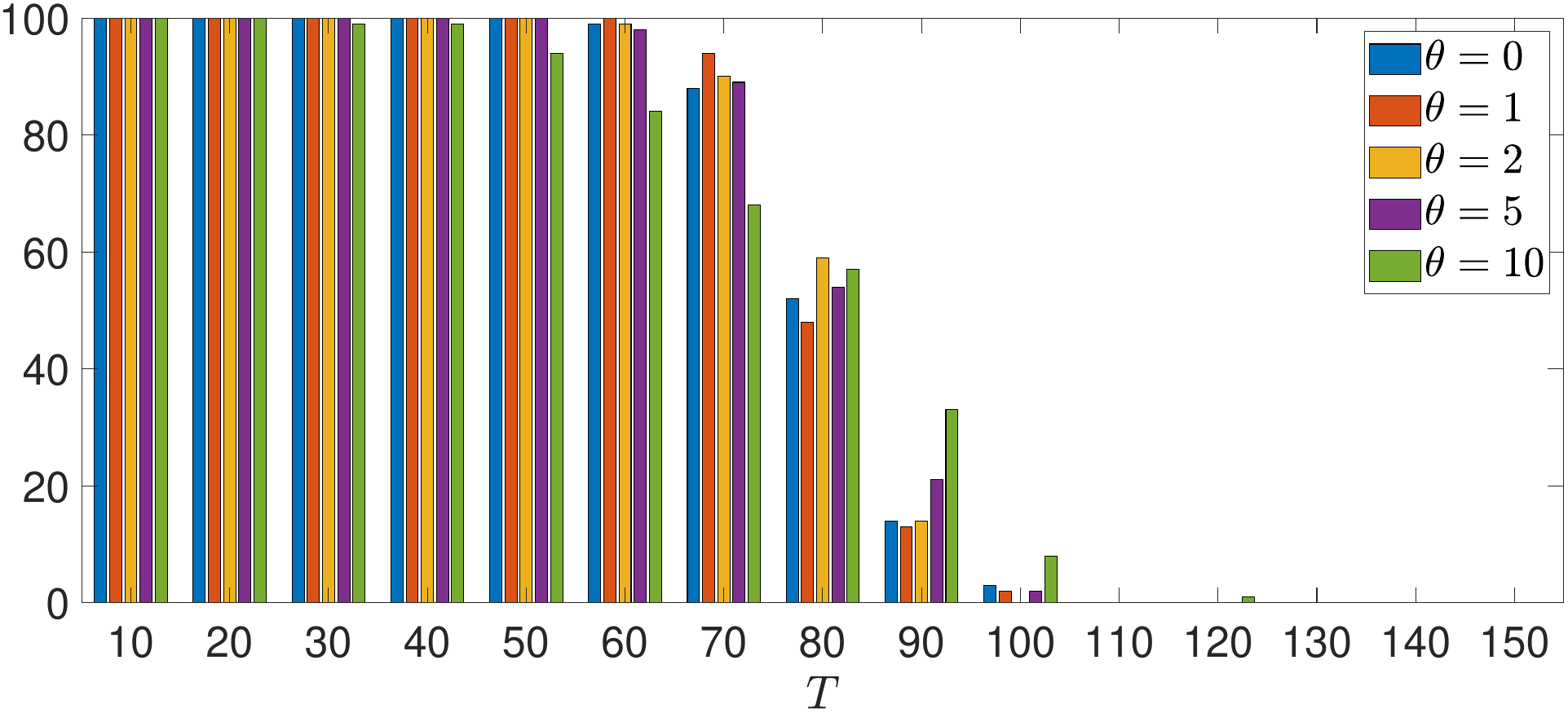}
\caption{Successful retrievals out of 100 trials with different sequence period lengths (sparse random projected targets).}
\label{fig:sparse}
\end{figure}

\end{document}